\documentclass{article}
\usepackage[a4paper, total={2in, 4in}]{geometry}
\usepackage[utf8]{inputenc}
\usepackage[english]{babel}

\usepackage{fullpage}
\usepackage{graphicx} %
\usepackage{hyperref}

\usepackage[dvipsnames]{xcolor}
\usepackage{bm}
\usepackage{amssymb}
\usepackage{amsthm}
\usepackage[cmex10]{amsmath}
\usepackage{booktabs}
\usepackage{caption}
\usepackage{subcaption}
\usepackage{comment}
\usepackage{mathtools}
\mathtoolsset{centercolon} 
\usepackage{gensymb}
\usepackage{mathrsfs} 
\usepackage{multirow}
\usepackage{bbm}
\usepackage{tabularx}
\usepackage{pdflscape}
\usepackage{float}
\usepackage{authblk}

\graphicspath{{./figure/}}

\usepackage{tikz} 
\usetikzlibrary{arrows.meta,arrows,shapes.geometric,positioning,fit}
\tikzset{>={Latex[width=3mm,length=2mm]}}

\usepackage{todonotes}
\usepackage{footnote}
\makesavenoteenv{tabular}
\makesavenoteenv{table}

\title{A Structural FHMM for Interpretable Disease Trajectories in T2DM}

\author[1]{Alessandro Mari}
\author[2]{Ekaterina Krymova}
\author[2]{Guillaume Obozinski}
\author[3]{Maria Luisa Marques de Sa Faquetti}
\author[3]{Adrian Martinez de la Torre}
\author[3]{Andrea Burden}
\affil[1]{\small Swiss Data Science Center (SDSC), Ecole Polytechnique Fédérale de Lausanne (EPFL), Switzerland}
\affil[2]{\small Swiss Data Science Center (SDSC), ETH Zürich, Switzerland}
\affil[3]{\small Pharmacoepidemiology Group, Institute of Pharmaceutical Sciences, Zurich, Switzerland}
\date{August 24, 2026}

\newcommand{\R}{\mathbb{R}}

\newcommand{\E}{\mathbb{E}}

\DeclareMathOperator*{\cov}{Cov}
\DeclareMathOperator*{\diag}{Diag}

\newcommand{\df}{\stackrel{\rm def}{=} }
\DeclareMathOperator{\Tr}{Tr}

\newcommand{\indicator}[1]{\mathbbm{1}_{\{#1\}}}

\DeclareMathOperator*{\argmax}{arg\,max}

\begin{document}

\maketitle

\begin{abstract} 
 In this work, we propose a structural variant of the Factorial Hidden Markov Model (FHMM) for the analysis of disease trajectories in patients with Type 2 diabetes mellitus (T2DM). The model represents a patient’s latent health state as a combination of multiple independent, simultaneously evolving components, associated with comorbidities and lab results. This structured latent representation facilitates the identification of clinically meaningful patient states and clustering of common disease trajectories. We evaluate the proposed approach using The IQVIA Medical Research Data incorporating data from THIN, a Cegedim database of anonymized electronic health records (EHR), identifying patients with a first-ever prescription for a non-insulin antidiabetic drug (NIAD) between January 2006 and December 2019. The model identifies multiple clinically coherent latent components corresponding to known patterns of diabetes-related complications and reveals heterogeneous progression pathways, including distinct microvascular-dominant and multi-organ trajectories associated with elevated comorbidity burden and mortality. These results demonstrate that the proposed framework captures meaningful longitudinal structure in EHR data and provides interpretable insights into the evolution of T2DM and its comorbidities.

\end{abstract}

\section{Background}

Type 2 diabetes mellitus (T2DM) is a chronic metabolic disorder characterized by elevated blood glucose levels resulting from insulin resistance and inadequate insulin secretion. It represents a major global health concern and is frequently accompanied by a range of comorbidities, which substantially worsen quality of life and increase the risk of morbidity and mortality \cite{Tomic2022}. The presence and progression of such comorbidities contribute to the heterogeneity of T2DM trajectories, making disease management and risk stratification particularly challenging.
The inference of a patient’s underlying health state from longitudinal clinical observations, together with information on T2DM and its associated comorbidities, can support the identification of disease progression patterns and emerging health risks. Leveraging retrospective electronic health records (EHRs) to characterize these evolving health states offers the potential to improve disease understanding, stratify patients into clinically meaningful subgroups, and inform timely interventions.

In this study, we analyze longitudinal observations from EHRs associated with T2DM and multiple comorbidities in order to infer patients’ latent health states and characterize common patterns of disease evolution. To this end, we propose a statistical model designed to capture the multi-dimensional and dynamic nature of patient health, with a particular focus on interpretability of the latent representation and its clinical relevance.
 
Specifically, we consider a structural variant of the Factorial Hidden Markov Model (FHMM) \cite{ghahramani1995factorial}, which enables modeling the dynamics of a multi-dimensional patient health state through several independent latent components evolving simultaneously over time. Given longitudinal clinical observations, the proposed model infers the latent configuration of these components, providing a structured representation of disease progression and associated health risks. In contrast to a standard Hidden Markov Model (HMM), which represents patient health using a single latent state, the FHMM framework allows the patient’s overall condition to be approximated as a combination of a small number of evolving latent processes. This structure increases modeling flexibility and facilitates interpretation, as individual latent chains can be associated with distinct aspects of disease evolution or comorbidity burden.
  
 By using retrospective information on patients’ health indicators and disease history, the proposed model enables the identification of clinically meaningful latent states and the analysis of transitions between them. The resulting latent trajectories can be used to cluster patients according to common progression patterns, characterize typical disease pathways, and assess the evolution of comorbidity-related risks over time. 

\textbf{Related work.}
Tracking and modeling comorbidities from longitudinal health observations in patients with T2DM is crucial for understanding disease heterogeneity and supporting effective disease management. A wide range of statistical and machine learning approaches have been proposed to analyze large-scale health data and uncover temporal patterns associated with disease progression and risk.
HMM and their generalizations are among the most widely used probabilistic approaches for modeling disease dynamics in healthcare applications \cite{alaa2017learning,wang2014unsupervised,ben2021profiling,liu2015efficient}. Extensions such as mixtures of Input–Output Hidden Markov Models have been proposed to capture more complex temporal dependencies \cite{ceritli2022mixture}. Nonparametric latent feature models, including approaches based on the Indian Buffet Process, have also been explored for comorbidity analysis, particularly in the context of psychiatric disorders \cite{ruiz2014bayesian}.

More recently, deep generative models have been applied to longitudinal clinical data. In \cite{trottet2023generative}, the authors propose a variational autoencoder-based framework to learn latent representations that jointly generate clinical measurements and medical concepts. While effective, their approach requires manual assignment of latent dimensions to specific medical concepts. In contrast, our model automatically learns a structured factorization of the latent space by construction. Moreover, the latent representation in our approach is discrete and subject to monotonicity constraints, which enhances interpretability and facilitates the analysis of disease trajectories.

Methods addressing the joint evolution of multiple diseases have also been proposed. For example, \cite{oflaz2023modeling} introduces a bivariate copula-based hidden process to model the interaction between two diseases using population-level time series. In the context of T2DM, machine learning approaches have been applied to model disease progression \cite{fregoso2021machine}, while clustering-based analyses have been used to study patterns of comorbidities \cite{martinez2022comorbidity}. Our work complements these approaches by providing a structured, interpretable latent-state model adapted to the analysis of complex, multi-comorbidity disease trajectories.

\section{Methods}\label{sec:methods}

\subsection{Data}

The IQVIA Medical Research Database UK (IMRD-UK) incorporates data supplied by The Health Improvement Network (THIN), a Cegedim database of anonymized electronic health records generated from the daily records of General Practitioners (GPs). The study protocol and use of the data were reviewed and approved by the Scientific Review Committee (SRC, reference number 20SR062).  

The dataset includes patients from participating primary care practices who have provided general consent for their data to be used for research purposes. Consent may be withdrawn at any time, in which case the individual’s data are removed from the dataset. Secondary care providers, such as ambulatory care or specialists, provide additional feedback on diagnoses to the GPs.  

Generally, the database comprises routinely collected data from clinical practice, including patient diagnoses, symptoms, prescriptions, referrals, immunizations, laboratory test results (including glycated hemoglobin [HbA1c]), selected behavioral factors (body mass index [BMI], alcohol use, and smoking status), and demographic characteristics.
As with most electronic healthcare databases, data are generated during routine care rather than for research purposes, resulting in unequally spaced observations and incomplete capture of certain variables. Lifestyle factors such as BMI, smoking status, and alcohol consumption are subject to missingness. In addition, diagnoses are recorded only when they are among the main reasons for a visit, laboratory tests are captured only when ordered; and while medication prescriptions are identified, prescription fills and adherence are not known.


 \subsection{Cohort description}
 We identified all adult patients (aged 18+) with a first-ever prescription of a non-insulin antidiabetic drug (NIAD) between January 1, 2006, and December 31, 2019. The index date was defined as the date of the first NIAD prescription. To ensure identification of new users, patients were required to have at least one year of valid data prior to the index date.
We excluded patients without any laboratory measurement for variables of interest (see Table \ref{tab:list_of_var}) within the 4-years prior to the index date after data aggregation (see Section \ref{sec:data_processing}).

\subsubsection{Data preprocessing}\label{sec:data_processing}

\begin{figure}[ht]
    \centering
    \includegraphics[width=\textwidth]{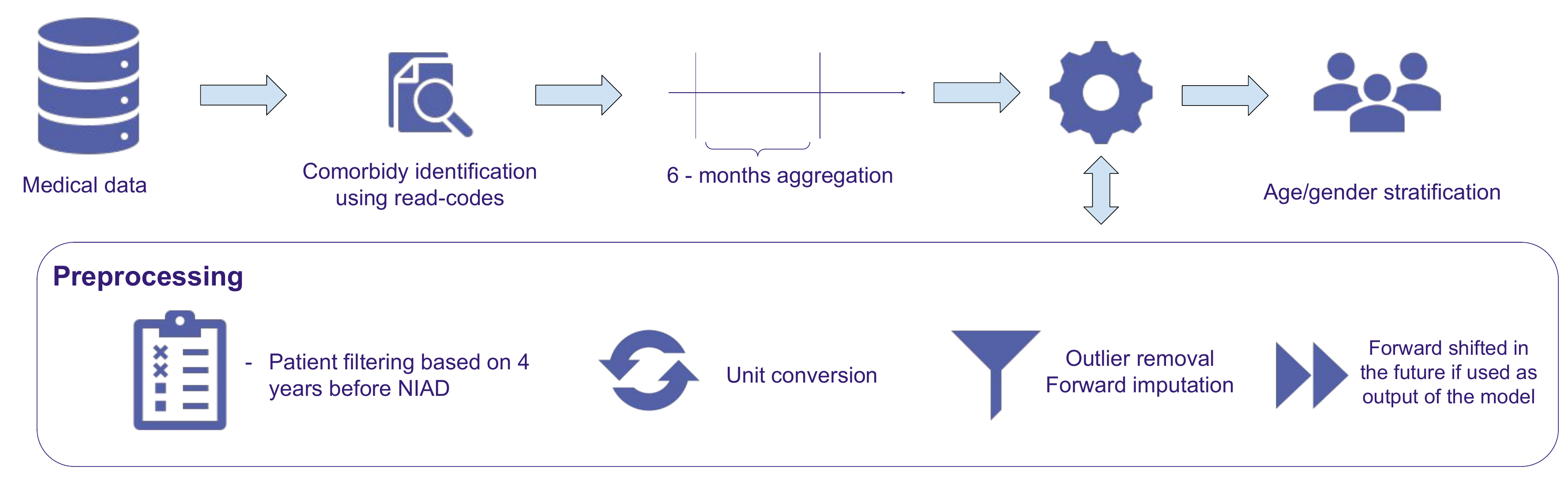}
    \caption{Graphical representation of the preprocessing steps.}
    \label{fig:preprocessing_step}
\end{figure} 

To harmonize the dataset and prepare it for modeling, we applied the following data preparation steps. A summary of these steps is provided in Figure \ref{fig:preprocessing_step}.

\paragraph{Comorbidities and Laboratory Measurements:} 
The comorbidities and laboratory variables included in the analysis are listed in Table \ref{tab:list_of_var}. Comorbidities were identified at or prior to the index date and during follow-up using read codes recorded at each patient’s clinical encounters. Only chronic conditions were considered. Accordingly, a comorbidity 
$c$ was defined as present at time 
$t$ if any corresponding read code appeared during the relevant six-month interval. Once identified, comorbidities were assumed to persist over time, reflecting their chronic or recurrent nature; thus, the indicator for comorbidity 
$c$ remained set to 1 following its first occurrence. In the base model, drugs were only used in modeling the initial state. A detailed discussion of the effects of drugs used in the transition matrix is provided in Appendix \ref{sec:drug}.

The laboratory values are standardized to consistent units (see Table \ref{tab:list_of_var}), and outliers are removed using predefined thresholds (see Appendix \ref{sec:data_preparation}).
To address irregular measurement timing and partial missingness inherent to routine clinical data, laboratory measurements were aggregated into six-month intervals, with the median value used when multiple measurements were available within a given interval. Patients were excluded if any required laboratory variable had no recorded measurements during the four years preceding NIAD initiation. This step was necessary to enable forward imputation of missing values during the post-NIAD period.

Additionally, to separate the timing of doctors' decisions regarding diagnoses and medication prescriptions, we used the comorbidity status (presence or absence) and lab tests shifted forward in time relative to the prescriptions. This approach helps avoid issues related to six-month aggregation, where diagnoses might be influenced by medication prescriptions rather than the patient's actual health state.

To further preserve the intended temporal ordering between patient health status and clinical decision-making, comorbidity indicators and laboratory measurements were shifted forward in time relative to medication prescriptions. This approach mitigates potential reverse-causation artifacts introduced by six-month aggregation, whereby diagnoses or laboratory results recorded shortly before any treatment initiation could otherwise appear contemporaneous with (or influenced by) drug exposure.

\paragraph{Clinical Stratification:}

Patient observations were stratified by age and sex to account for clinically meaningful heterogeneity. Age was determined at the index date and categorized into two groups: patients aged 64 years or younger and those aged 65 years or older. This stratification reflects differences in diabetes progression and the burden of age-related comorbidities across the life course \cite{Kalyani2017}. Additionally, analyses were stratified by sex to capture known differences in disease prevalence and comorbidity profiles, such as osteoporosis, between male and female patients \cite{Russo2016}.

\begin{table}[ht]
    \centering

\resizebox{\textwidth}{!}{\begin{tabular}{  p{4.5cm} | p{5.cm}| p{5.cm}  }
  \toprule
  Comorbidities & Lab Values & Drugs (with ATC code) \\ 
  \midrule
  \begin{itemize}
      \item 
  Stroke
  \item 
Myocardial infarction (MI)
\item 
Congestive heart failure (CHF)
\item 
Peripheral vascular disease (PVD)
\item 
Diabetic retinopathy (DR)
\item 
Hypertension (HT)
\item 
Chronic kidney disease (CKD)
\item 
Osteoporosis
\item 
Chronic obstructive pulmonary disease (COPD)
\item Chronic liver disease (CLD)
  \end{itemize}
& 
\begin{itemize}
    \item 
Glycated hemoglobin (Hb A1c) in \%
\item 
Low-density Lipoprotein (LDL) cholesterol in mmol/L
\item 
High-density Lipoprotein (HDL) cholesterol in mmol/L
\item 
Triglycerides (TG) in mmol/L
\item 
Blood pressure (Systolic/Diastolic) in mmHg
\item 
Body Mass Index (BMI) in kg/m\textsuperscript{2}
\item
Glomerular filtration rate (GFR) in mL/min/1.73m\textsuperscript{2}
\end{itemize}
&
\begin{itemize}
\item Diabetes Drugs \footnote{Including Gliclazide,
 Glimepiride,
 Glipizide,
 Metformin,
 Pioglitazone,
 Rosiglitazone,
 Sitagliptin, Dapagliflozin.}
\item Antithrombotic agents (B01)
\item Cardiac therapy (C01)
\item Antihypertensives (C02)
\item Diuretics (C03)
\item Beta blocking agents (C07)
\item Calcium channels blockers (C08)
\item ACE inhibitors (C09)
\item Statins (C10)
\item Corticosteroids (H02)
\item Drugs for treatment of bone diseases (M05)
\end{itemize}\\
  \bottomrule
\end{tabular}}
\caption{List of variables that are included in the model.\label{tab:list_of_var}}
\end{table}

\subsection{Structural FHMM}\label{sec:fhmm}

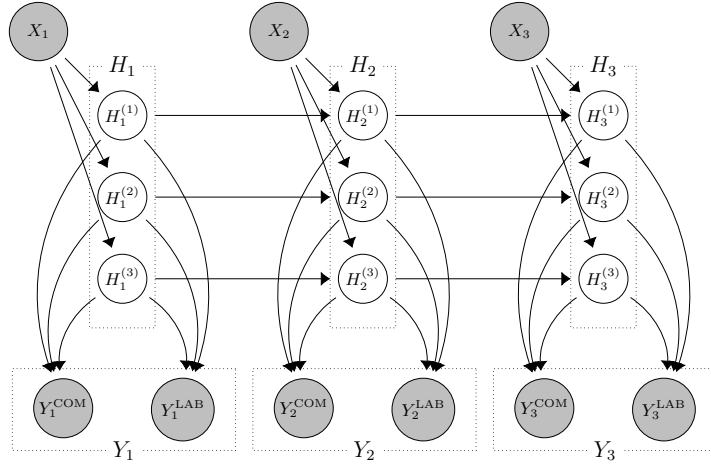
\begin{figure}[ht]
\centering\scalebox{0.6}{
\begin{tikzpicture}[node distance={15mm},main/.style = {draw, circle},scale=1., every node/.style={scale=1.2},titlerect/.style={draw,label={[anchor=center,fill=white,font=\large\bfseries\sffamily]above: #1}},titlerectbelow/.style={draw,label={[anchor=center,fill=white,font=\large\bfseries\sffamily]below: #1}}]
\node[main] (x1) [fill=lightgray, inner sep=6pt] {\small$X_1$};
\node[main] (x2) [right=4cm of x1,fill=lightgray, inner sep=6pt] {\small$X_2$}; 
\node[main] (x3) [right=4cm of x2,fill=lightgray, inner sep=6pt] {\small$X_3$};   
    
\node[main] (h11) [below right=1cm and 1cm of x1, inner sep=1.5pt] {\small $H_1^{(1)}$};
\node[main] (h21) [below right=1cm and 1cm of x2, inner sep=1.5pt] {\small $H_2^{(1)}$};
\node[main] (h31) [below right=1cm and 1cm of x3, inner sep=1.5pt] {\small $H_3^{(1)}$}; 
\node[main] (h12) [below of=h11,inner sep=1.5pt]{\small $H_1^{(2)}$};
\node[main] (h22) [below of=h21,inner sep=1.5pt] {\small $H_2^{(2)}$};
\node[main] (h32) [below of=h31,inner sep=1.5pt] {\small $H_3^{(2)}$}; 
\node[main] (h13) [below of=h12,inner sep=1.5pt]{\small $H_1^{(3)}$};
\node[main] (h23) [below of=h22,inner sep=1.5pt] {\small $H_2^{(3)}$};
\node[main] (h33) [below of=h32,inner sep=1.5pt] {\small $H_3^{(3)}$}; 

\node[main] (y1) [below right=2.cm and 0.45cm of h13,fill=lightgray,inner sep=3pt] {\small$Y_1^\text{LAB}$};
\node[main] (y2) [below right=2.cm and 0.45cm of h23,fill=lightgray,inner sep=3pt] {\small$Y_2^\text{LAB}$}; 
\node[main] (y3) [below right=2.cm and 0.45cm of h33,fill=lightgray,inner sep=3pt] {\small$Y_3^\text{LAB}$};

\node[main] (y1_) [below left=2.cm and 0.45cm of h13,fill=lightgray,inner sep=1.5pt] {\small$Y_1^\text{COM}$};
\node[main] (y2_) [below left=2.cm and 0.45cm of h23,fill=lightgray,inner sep=1.5pt] {\small$Y_2^\text{COM}$}; 
\node[main] (y3_) [below left=2.cm and 0.45cm of h33,fill=lightgray,inner sep=1.5pt] {\small$Y_3^\text{COM}$};

\node[draw,dotted,fit=(h11) (h12) (h13),inner sep=1.5pt,titlerect={$H_1$}] {};
\node[draw,dotted,fit=(h21) (h22) (h23),inner sep=1.5pt,titlerect={$H_2$}] {};
\node[draw,dotted,fit=(h31) (h32) (h33),inner sep=1.5pt,titlerect={$H_3$}] {};

\node[draw,dotted,fit=(y1) (y1_),inner sep=1.5pt,titlerectbelow={$Y_1$}] {};
\node[draw,dotted,fit=(y2) (y2_),inner sep=1.5pt,titlerectbelow={$Y_2$}] {};
\node[draw,dotted,fit=(y3) (y3_),inner sep=1.5pt,titlerectbelow={$Y_3$}] {};

\draw[->,shorten >= 5pt, shorten <= 5pt] (h11) -- (h21);
\draw[->,shorten >= 5pt, shorten <= 5pt] (h12) -- (h22);
\draw[->,shorten >= 5pt, shorten <= 5pt] (h13) -- (h23);
\draw[->,shorten >= 5pt, shorten <= 5pt] (h21) -- (h31);
\draw[->,shorten >= 5pt, shorten <= 5pt] (h22) -- (h32);
\draw[->,shorten >= 5pt, shorten <= 5pt] (h23) -- (h33);

\draw[->,shorten >= 5pt, shorten <= 5pt] (h11) to [bend left=30] (y1);
\draw[->,shorten >= 5pt, shorten <= 5pt] (h12) to [bend left=30] (y1);
\draw[->,shorten >= 5pt, shorten <= 5pt] (h13) to [bend left=30] (y1);
\draw[->,shorten >= 5pt, shorten <= 5pt] (h21) to [bend left=30] (y2);
\draw[->,shorten >= 5pt, shorten <= 5pt] (h22) to [bend left=30] (y2);
\draw[->,shorten >= 5pt, shorten <= 5pt] (h23) to [bend left=30] (y2);
\draw[->,shorten >= 5pt, shorten <= 5pt] (h31) to [bend left=30] (y3);
\draw[->,shorten >= 5pt, shorten <= 5pt] (h32) to [bend left=30] (y3);
\draw[->,shorten >= 5pt, shorten <= 5pt] (h33) to [bend left=30] (y3);

\draw[->,shorten >= 5pt, shorten <= 5pt] (h11) to [bend right=30] (y1_);
\draw[->,shorten >= 5pt, shorten <= 5pt] (h12) to [bend right=30] (y1_);
\draw[->,shorten >= 5pt, shorten <= 5pt] (h13) to [bend right=30] (y1_);
\draw[->,shorten >=5pt, shorten <= 5pt] (h21) to [bend right=30] (y2_);
\draw[->,shorten >= 5pt, shorten <= 5pt] (h22) to [bend right=30] (y2_);
\draw[->,shorten >= 5pt, shorten <= 5pt] (h23) to [bend right=30] (y2_);
\draw[->,shorten >= 5pt, shorten <= 5pt] (h31) to [bend right=30] (y3_);
\draw[->,shorten >= 5pt, shorten <= 5pt] (h32) to [bend right=30] (y3_);
\draw[->,shorten >= 5pt, shorten <= 5pt] (h33) to [bend right=30] (y3_);

\draw[->,shorten >= 5pt, shorten <= 5pt] (x1) to [right=30] (h11);
\draw[->,shorten >= 5pt, shorten <= 5pt] (x1) to [right=30] (h12);
\draw[->,shorten >= 5pt, shorten <= 5pt] (x1) to [right=30] (h13);
\draw[->,shorten >= 5pt, shorten <= 5pt] (x2) to [right=30] (h21);
\draw[->,shorten >= 5pt, shorten <= 5pt] (x2) to [right=30] (h22);
\draw[->,shorten >= 5pt, shorten <= 5pt] (x2) to [right=30] (h23);
\draw[->,shorten >= 5pt, shorten <= 5pt] (x3) to [right=30] (h31);
\draw[->,shorten >= 5pt, shorten <= 5pt] (x3) to [right=30] (h32);
\draw[->,shorten >= 5pt, shorten <= 5pt] (x3) to [right=30] (h33);
\end{tikzpicture} }
\caption{Directed graphical model of a FHMM with $M=3$ hidden chains. Arrows represent causal relationships and gray nodes are observed variables. White nodes are not observed. The variables $X_t$ are the drugs taken at time $t$ that influence the patient health state $H_t\df(H_t^{(1)},\dots,H_t^{(M)})$. The results of the lab tests $Y_t^{\text{LAB}}$ and the set of comorbidities $Y_t^{\text{COM}}$ are direct consequences of the health state $H_t$ at time $t$.}
\label{fig:fhmm}
\end{figure}

Our modeling approach is based on Factorial Hidden Markov Model (FHMM) \cite{ghahramani1995factorial}, which is an extension of the hidden Markov chain, where the output is conditioned not on one hidden Markov chain with $M$ states, but on $K$ independent hidden Markov chains. This allows to model the time series depending on several independent latent processes (see Figure \ref{fig:fhmm}). In terms of comorbidities and lab-values progression modeling, the FHMM approach translates into the assumption that there exist $K$ hidden health state components, which are independent and that are jointly influencing the output. 
Although the independence assumption may appear unrealistic, since body systems rarely function in isolation, the FHMM can be understood as an independent basis decomposition of the dynamics, where several hidden processes evolve separately and jointly generate the observed comorbidities and lab values. This view highlights FHMM as a structured approximation: the latent chains act as abstract basis processes that capture key aspects of disease progression.

We assume that the evolution of comorbidity states and laboratory test measurements, denoted by $Y_t$, follows the generative process defined by
\begin{align}
    H_1^{(m)}\mid X_1&\sim \pi^{(m)}(\cdot\mid X_1;{\theta^\pi})\label{eqn:initial_gen},& m=1,\dots,M,\\
    H_t^{(m)}\mid H_{t-1}^{(m)},X_t&\sim P^{(m)}(\cdot\mid H_{t-1}^{(m)},X_t;{\theta^P})\label{eqn:trans_gen},& m=1,\dots,M,\\
    Y_t\mid H_t&\sim p(\cdot\mid H_t;\eta),\label{eqn:emission_gen}
\end{align}
where $H_t\df(H_t^{(1)},\dots,H_t^{(M)})$ denotes the joint latent configuration of $M$ parallel hidden Markov chains with $K$ possible states each. At each time point, the combination of these $M$ latent variables, each taking one of K-states, determines the distribution of the observable $Y_t$. 

In the presence of the auxillary input variables, such as in our case medications, one can model the output assuming the dependence of the hidden states on the input, which leads to Input-output FHMM formulation.

Exact inference for \eqref{eqn:emission_gen} can, in principle, be carried out by representing the FHMM as a single HMM with $K^M$ states, where the transition matrix can be written as
\begin{equation*}
    P = P^{(1)} \otimes P^{(2)} \otimes \cdots \otimes P^{(M)}\in \R^{K^M}\times \R^{K^M},
\end{equation*}
where $\otimes$ is the Kronecker product. Nevertheless, the introduction of additional chains makes the estimation intractable, as a naive implementation of the Forward-Backward algorithm has complexity $O(K^{2M}T)$. The complexity can be reduced to $O(MK^{M+1}T)$ by carefully computing the matrix-vector dot product \cite{FactorialHMM_python}, but still grows exponentially in the number of chains.
One of the ways to make the estimation tractable is to use variational approximation, which we describe in Section \ref{sec:model_estimation}.

We impose the following problem-specific structure on the FHMM to improve identifiability and interpretability.

\subsubsection{Problem-specific structure}\label{sec:structure}

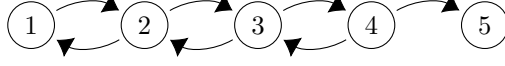
\begin{figure}[ht!]
    \centering
    \begin{tikzpicture}[node distance={15mm},main/.style = {draw, circle}]
\node[main] (x1)  {1};
\node[main] (x2) [right of=x1] {2}; 
\node[main] (x3) [right of=x2] {3};  
\node[main] (x4) [right of=x3] {4};  
\node[main] (x5) [right of=x4] {5};  
    
\draw[->,shorten >= 2pt, shorten <= 2pt] (x1) to [bend left=30] (x2);
\draw[->,shorten >= 2pt, shorten <= 2pt] (x2) to [bend left=30] (x1);
\draw[->,shorten >= 2pt, shorten <= 2pt] (x2) to [bend left=30] (x3);
\draw[->,shorten >= 2pt, shorten <= 2pt] (x3) to [bend left=30] (x2);
\draw[->,shorten >= 2pt, shorten <= 2pt] (x3) to [bend left=30] (x4);
\draw[->,shorten >= 2pt, shorten <= 2pt] (x4) to [bend left=30] (x3);
\draw[->,shorten >= 2pt, shorten <= 2pt] (x4) to [bend left=30] (x5);
\end{tikzpicture} 
    \caption{Possible transitions between states of a hidden chain with $5$ states and jumps of size $d=1$.
    \label{fig:markov_chain}}
\end{figure}

The transition matrix specifies the probabilities of moving between latent health states from one time step to the next. The general FHMM defined in Section \ref{sec:fhmm} is not easily interpretable as the states of the chaines are not ordered and any permutation of the states labels $\{1,...,K\}$ gives the same model. 
In order to make the identification of the states possible, we assume that states are ordered by worsening of the patient's condition, and that lower state values represent a healthy patient. Additionally, we assume that the health of a patient can degrade only by $d$ units (a jump of $d$ states) and that the last state is an absorbing state, representing the impossibility of improving the  patient's health condition. These assumptions mean that the transition matrix $P_{\theta^P}$ is a band matrix with bandwidth $d$, where the last row of the matrix $P_{\theta^P}$ contains all zeros except in the last column, which is equal to one. Figure \ref{fig:markov_chain} shows all possible transitions when $d=1$. The transition matrix is fully parameterized without the inclusion of input variables, such as medications. A discussion on incorporating medications into the parameterization is provided in Appendix \ref{sec:drug}.

    The Emission probability $p(\cdot|H_t;\eta)$ should capture that the higher the patients' state value is,  the worse their health conditions are. In particular, the conditional mean 

    $$\E[Y_t|H_t]\df g\left(\eta_0+\sum_m {\ \tilde{H}_t^{(m)}}^\top \eta^{(m)}\right),$$
    should be a monotone function of each state component $H_t^{(m)}$ for all $m$, where $\tilde{H}_t^{(m)}\in \{0,1\}^K$ is a one-hot encoded version of $H_t^{(m)}\in \{1,\dots,K\}$ and $g$ is some link function. Note that $\eta_1^{(m)}=0$ for identifiability issues and that the intercept is captured by $\eta_0$.
    
    As in \cite{Chiquet2012},  to guarantee monotonicity we  parametrize the variable $\eta^{(m)}$ as the cumulative sum of $\eta_+^{(m)}\geq0 $ and $\eta_-^{(m)}\geq0$, i.e.,
    $$\eta^{(m)} \df D \eta_+^{(m)}- D \eta_-^{(m)},$$ 
    where $D$ is a lower triangular matrix with all ones. This means that each difference between successive coefficients is either non-negative or non-positive: if $\eta_-^{(m)}=0$, then 
    $$\eta_i^{(m)}-\eta_{i-1}^{(m)}=\eta_{+,i}^{(m)}\geq 0 \iff \eta_i^{(m)}\geq \eta_{i-1}^{(m)},$$ hence $\eta^{(m)}$ is non-decreasing.
    Similarly, if $\eta_+^{(m)}=0$, then 
    $$\eta_i^{(m)}-\eta_{i-1}^{(m)}=-\eta_{-,i}^{(m)}\leq 0 \iff \eta_i^{(m)}\leq \eta_{i-1}^{(m)},$$
     hence $\eta^{(m)}$ is non-increasing.
To decide whether the slope  should be positive or negative, we add a group-lasso penalty \cite{grouplasso}: 
    \begin{equation}\label{eqn:group_lasso}
        p(\eta)=\lambda\times \sum_m  \left(\|\eta_-^{(m)}\|_2+\|\eta_-^{(m)}\|_2\right).
    \end{equation}
This penalty enforces sparsity at the group level, i.e. it encourages either $\eta_+^{(m)}=0$
(purely decreasing slope) or $\eta_-^{(m)}=0$
(purely increasing slope).

The initial state is modeled as a function $\pi_{\theta^\pi}$ of the list of drugs taken over a period of 4 years prior to NIAD, age, smoking states and alcohol status jointly denoted by $X_1$. The distribution is assumed to be unimodal and is parametrized by a gamma distribution, i.e.,
\begin{equation}\label{eqn:initial_dist}
    \left[\pi_{\theta^\pi}(X_1)\right]_i=\text{Softmax}\left\{(\alpha(X_1)-1)\log i-\beta^{\text{init}} i\right\}\propto i^{\alpha(X_1)-1} \exp(-\beta^{\text{init}} i),
\end{equation}
with $\alpha(X_1) =  W^{\text{init}}X_1+\alpha^{\text{init}}$ and ${\theta^\pi}=\{W^{\text{init}},\alpha^{\text{init}},\beta^{\text{init}}\}$. All parameters are assumed to be non-negative. If $\alpha$ is close to zero, the probability mass is concentrated mostly around the lowest state value, whereas larger values of $\alpha$ increase the probability of starting from a higher state value.

Assuming that a drug prescription is equivalent to a degradation of the patient's health, we impose $ W^{\text{init}}\geq 0$ and $\alpha^{\text{init}}\geq 0$. In addition, $W^{\text{init}}$, ${\alpha^{\text{init}}}$ and ${\beta^{\text{init}}}$ follow a priori a (truncated) Gaussian distribution centered at 0, 1 and 1 respectively, which is equivalent to add the penalizations $\|W^{\text{init}}\|_2^2$, $\|{\alpha^{\text{init}}}-1\|_2^2$ and $\|{\beta^{\text{init}}}-1\|_2^2$ to the objective function. The underlying assumptions are 
\begin{enumerate}
    \item the drug effect is a priori small (i.e., $W^{\text{init}}$ is likely to be small),
    \item in the absence of drugs, the likelihood of starting from a higher state is small, that is, the prior distribution is 
   $ \left[\pi_{\theta^\pi}(X_1)\right]_i\propto \exp(-i).$
\end{enumerate}

\subsection{Model estimation}\label{sec:model_estimation}
Parameters of the models are estimated via Expectation-Maximization (EM) following the work of Ghahramani \cite{ghahramani1995factorial}, with the aim to maximize the Evidence Lower Bound (ELBO)
\begin{equation}\label{eqn:elbo}
   \max_{\theta,q} F(q,\theta) = \max_{\theta,q} \E_{h_{1:T}\sim q}\left[\log\frac{ p_\theta(y_{1:T},h_{1:T}\mid x_{1:T})}{q(h_{1:T}\mid x_{1:T},y_{1:T})}\right],
\end{equation} 
over a class of approximative distributions $q\in \mathcal{D}$ (E step) and the parameters $\theta=(\theta^\pi,\theta^P,\eta)$ (M step), where we used the notation $s_{a:b}$ to denote the sequence $s$ from $a$ to $b$. In \cite{ghahramani1995factorial}, the authors proposed to use a structured mean field variational approximation \cite{Wainwright08} 
$$q(h_{1:T}\mid x_{1:T})\df\prod_m q(h_{1:T}^{(m)}\mid x_{1:T},y_{1:T}),$$ 
with 
\begin{align}
\begin{split}\label{eqn:variational_eq}
  q\left(H_1^{(m)}=i\mid x_1,y_{1:T}\right)&\propto e_{1,i}^{(m)}\times \pi_i^{(m)},\\
q\left(H_t^{(m)}=j\mid H_{t-1}^{(m)}=i, x_t,y_{1:T}\right)&\propto e_{t,j}^{(m)} \times P_{i,j}^{(m)},
\end{split}
\end{align}
as an approximation of the optimal distribution $q(h_{1:T}\mid x_{1:T},y_{1:T})=p(h_{1:T}\mid y_{1:T}, x_{1:T})$, which reduces the inference complexity to $O(K^2MT)$ in the case of Gaussian emissions. The parameters $\phi =\{e_{t,j}^{(m)}\}_{t,j,m}$ are learned by optimizing the ELBO \eqref{eqn:elbo}.

In the case of binary outputs, such as comorbidities, the complexity still grows exponentially in the number of chains $O(MK^{M+1}T)$ \cite{FactorialHMM_python}. However, inspired by the Gaussian case, a second order approximation of the emission log-likelihood $\log p_\eta(y_t \mid {h_t^{(1:M)}})$ is derived using the inequality \cite{tanh}
\begin{equation}\label{eqn:lower_bound_elbo}
    -\log(1+\exp y)\geq -\log\left\{1+\exp(-z)\right\} - (y+z)/2 - \frac{\lambda(z)}{2}(y^2-z^2),
\end{equation}
with $\lambda(z)=\frac{\text{tanh}(z/2)}{2z}$ and some additional variable $z\in\R$. Exploiting this inequality reduces the complexity of the model training to $O(K^2MT)$ (see Appendix \ref{sec:variational} for more details). After training, parameters $\theta$ are kept fixed and the parameters $\phi$ are learned when presented with new data. We refer this step as "fine-tuning".

\subsection{Trajectory analysis}\label{sec:trajectory_analysis}

The emission parameters $\eta^{(m)}$ model the risk of developing a comorbidity for a given state, but do not capture the progression of comorbidities over time. On the other hand, the sequence of hidden states $\{H_t^{(m)}\}_{t>0}$ encodes the temporal dynamics of disease evolution, allowing us to infer how patients transition between latent health states and how these transitions influence the development and accumulation of comorbidities across time.

In particular, the variational posterior distribution $q_\phi(H_t^{(m)}\mid x_{1:T},y_{1:T})$ is given by \eqref{eqn:variational_eq}, after fitting the parameters $\phi$ for each patient. The latter is computed using the Baum–Welch algorithm for HMMs (see \cite{ghahramani1995factorial}).
From the posterior distribution, we calculate the expected state of each latent chain at each time step as $$\E_q[h_t^{(m)}] = \sum_{h_t} h_t  q_\phi^{(m)}(h_t\mid x_{1:T},y_{1:T}),$$ 
yielding an $M$-dimensional representation per time point. We then apply K-means clustering to all time steps, selecting the number of clusters based on the Davies–Bouldin index and prediction strength \cite{Davies1979Cluster,Tib2005Cluster}.

\subsection{Experimental setup}

Patient data were partitioned into three distinct subsets: training (80\%), validation (10\%), and test (10\%). Stratification was applied to ensure a consistent distribution of comorbidities across these subsets (see Table \ref{tab:proportion_split} in the Appendix). 

All models were trained for a maximum of 300 EM iterations, with early stopping applied if the objective function $F$ showed no improvement. Model parameters were initialized to zero, except for $\alpha^{\text{init}}$, $\beta^{\text{init}}$, which were initialized to one.
Multiple FHMMs with varying numbers of chains and states were trained, and model selection was performed using the Akaike Information Criterion (AIC) \cite{akaike1974} and the Bayesian Information Criterion (BIC) \cite{schwarz1978}.

\section{Results}

\subsection{Patient cohort and characteristics}

\begin{figure}[ht]
    \centering
    \includegraphics[width=0.5\linewidth]{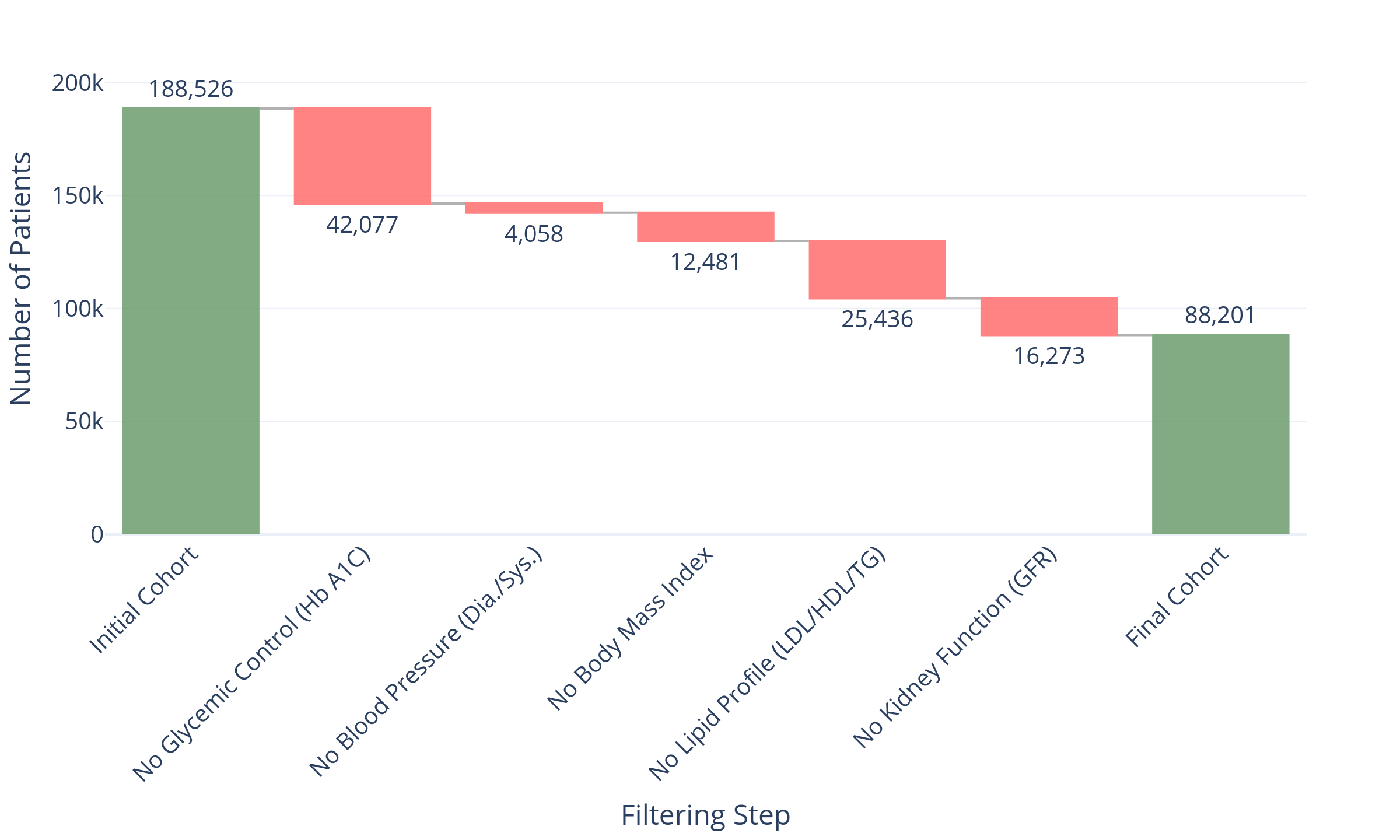}
    \caption{Filtering of patients based on missing laboratory data during the four years preceding NIAD initiation. A total of 100325 patients were excluded. Abbreviations are defined in Table~\ref{tab:list_of_var}.}
    \label{fig:removed_pat}
\end{figure}

After applying the preprocessing workflow described in Section \ref{sec:data_processing}, the final cohort comprised 88201 patients, including 47983 men and 40218 women. Figure \ref{fig:removed_pat} illustrates the filtering process of patients based on missing laboratory data during the four years preceding NIAD initiation. In total, 100325 patients were excluded due to missing data for key laboratory values.

\begin{table}[!ht]

    \centering
    \resizebox{\textwidth}{!}{\begin{tabular}{c|cc|cc}
    \toprule
     Sex & \multicolumn{2}{c}{Men}& \multicolumn{2}{c}{Women}\\
         Age group & $< 65$& $65+$&  $< 65$& $65+$ \\
         \midrule
         Number of patients &27191&21989&20792&18229\\
         Average age at NIAD (y) $\pm$ SD &54.0 $\pm$ 8.0&73.7 $\pm$ 6.3&52.5 $\pm$ 9.3&74.9 $\pm$ 6.7\\
         Median follow-up duration in years post-NIAD [IQR] &4.5 [5.0]&4.0 [4.5]&4.0 [5.0]&4.0 [4.5]\\
         Average BMI at NIAD (kg/m\textsuperscript{2}) $\pm$ SD &32.9 $\pm$ 6.3&30.1 $\pm$ 5.1&35.4 $\pm$ 7.9&31.0 $\pm$ 6.4\\
         Average HbA1c at NIAD (\%) $\pm$ SD &7.4 $\pm$ 1.7&7.1 $\pm$ 1.4&7.2 $\pm$ 1.6&7.0 $\pm$ 1.3\\
         \% Ever smoked & 45.2 & 34.4 & 39.4 & 28.8\\
         \midrule
         \multicolumn{5}{l}{\textit{Comorbidities at NIAD initiation (\%)}} \\
Stroke & 5.3 & 15.6 & 4.5 & 13.9 \\
Myocardial infarction & 8.0 & 15.8 & 2.6 & 7.1 \\
Congestive heart failure & 3.8 & 11.9 & 1.9 & 9.2 \\
Peripheral vascular disease & 3.3 & 10.1 & 1.4 & 5.0 \\
Diabetic retinopathy & 32.3 & 35.3 & 29.8 & 34.7 \\
Hypertension & 57.6 & 70.7 & 53.9 & 77.4 \\
Chronic kidney disease & 5.4 & 10.6 & 6.6 & 9.6 \\
Osteoporosis & 0.6 & 1.9 & 2.2 & 10.6 \\
COPD & 6.4 & 15.6 & 6.7 & 13.5 \\
Chronic liver disease & 5.1 & 3.0 & 6.2 & 3.7 \\
\midrule
\multicolumn{5}{l}{\textit{Laboratory measurements 6 months before NIAD}} \\
Diastolic blood pressure (mmHg) & 
84.5 (31.7) & 77.4 (22.3) & 81.5 (38.0) & 77.5 (22.0) \\
Systolic blood pressure (mmHg) & 
139.1 (31.7) & 138.5 (22.3) & 134.3 (38.0) & 140.5 (22.0) \\
GFR (mL/min/1.73\,m\textsuperscript{2}) & 
76.8 (35.2) & 67.6 (29.2) & 75.3 (46.3) & 64.1 (29.3) \\
HDL (mmol/L) & 
1.1 (31.0) & 1.2 (31.5) & 1.2 (48.0) & 1.4 (33.7) \\
LDL (mmol/L) & 
3.1 (45.5) & 2.5 (43.6) & 3.3 (56.5) & 2.8 (44.8) \\
Triglycerides (mmol/L) & 
2.8 (35.3) & 2.1 (37.1) & 2.3 (51.1) & 2.1 (38.8) \\

         \bottomrule
    \end{tabular}}
    \caption{Patient characteristics at the index date (NIAD initiation) for the cohort after preprocessing. Comorbidities are reported as percentages of the cohort population. Laboratory measurements are summarized as mean values, with the proportion of missing observations reported in parentheses. Abbreviations are defined in Table \ref{tab:list_of_var}. SD denotes standard deviation and IQR denotes interquartile range.}
    \label{tab:proportion_comorbidity}
\end{table}

The median follow-up duration after NIAD initiation was approximately 4.0 years across all groups. Baseline patient characteristics at the index date (first NIAD prescription) are summarized in Table \ref{tab:proportion_comorbidity}. Continuous variables, including age, body mass index (BMI), and glycated hemoglobin (HbA1c), are reported as medians with interquartile ranges (IQRs), while categorical variables—such as sex, smoking status, and comorbidities—are presented as counts and percentages.

Smoking prevalence was higher among men than women. Hypertension and diabetic retinopathy were the most prevalent comorbidities at baseline, with diabetic retinopathy showing the greatest increase in prevalence following NIAD initiation. Comorbidities that developed after NIAD initiation are indicated in parentheses in Table \ref{tab:proportion_comorbidity}.

\subsection{Model selection}\label{sec:model_selection}

Table \ref{tab:model_selection} presents the AIC and BIC scores for various models evaluated on the validation set. We use the notation $\text{FHMM}(M,K)_d$ to denote a factorial hidden Markov model (FHMM) with $M$ chains, $K$ states per chain, and a transition matrix bandwidth of $d$. Similarly, $\text{HMM}_d$ refers to a standard hidden Markov model (HMM) with $K$ states and a transition matrix $P_{\theta^P}$ of bandwidth $d$. 

\begin{table}[ht]
    \centering
    \small
    \begin{tabular}{c|cccccccc}
        & \multicolumn{2}{c}{Men $<65$}& \multicolumn{2}{c}{Women $<65$}&\multicolumn{2}{c}{Men $65+$}& \multicolumn{2}{c}{Women $65+$}\\
        Model & BIC & AIC & BIC & AIC& BIC & AIC & BIC & AIC \\
        \midrule
        $\text{HMM}(27)_8$ &329077&314469 &236219&221959& 234704&220497 &212736&198738\\
        $\text{FHMM}(3,3)_1$& 310663&307445 & 225526 &222385&229629 &226500  &203633&200550\\
        
        $\text{FHMM}(4,4)_2$ & 292039& 286438& 216236&210769&221682&216235&195206&189839\\
        $\text{FHMM}(4,6)_2$ & 293949& 285052&211923 &203239 &220663 &212010 & 195312&186787\\
        
        $\text{FHMM}(6,4)_2$ & \textbf{264231}& \textbf{256136} & 190847 &182946&201848&193976&181320&173563\\
        $\text{FHMM}(6,6)_2$ & 273908& 260869&196096 &183369 &  196499& 183819& 181122 &168628\\
        
        $\text{FHMM}(8,4)_2$ &{273337}&{262748 }&\textbf{178959}&\textbf{168623}&\textbf{181179}&\textbf{170881}&\textbf{162739}&\textbf{152592}\\
        
        $\text{FHMM}(10,2)_1$ &{276474}& {270739} & 211039 &192272&{186698}&188498&{174079}&168585\\
        
        \bottomrule
    \end{tabular}
    \caption{Comparison between models with various number of chains. Metrics are evaluated on the validation set and drugs are not used as input.}
    \label{tab:model_selection}
\end{table}

For comparison, an HMM with $K=27=3^3$ states is evaluated against an FHMM of equivalent size. In the FHMM, each chain has 2 possible transitions ($d=1$), which corresponds to an HMM bandwidth of $d=8=2^3$. Because the relationship between HMM state values and worsening health conditions is less direct (where a single disease progression may be too restrictive), the HMM emission probabilities are parameterized without monotonicity constraints, and drug information is excluded from the transition matrix input.

Memory constraints limit our comparison to a state dimension of $\tilde{K}=K^3=27$, with $K=3$. In the HMM, the full matrix 
\[
\xi_t = \mathbb{E}\left[\tilde{H}_t \tilde{H}_{t-1}^\top\right] \in \mathbb{R}^{K^3 \times K^3}
\] 
must be computed for all patients and timesteps, where $\tilde{H}_t$ is the one-hot encoding of $H_t$. In contrast, the FHMM primarily requires computing
$$\xi_t^{(m)}=\E\left[\tilde{H}_t^{(m)} {\tilde{H}_{t-1}^{(m)}\ }^\top \right]\in \R^{K\times K}$$
for all patients, timesteps $t$, and chains $m$. Since memory complexity scales quadratically with $K$, we kept $K$ small. In particular, increasing the number of states from 4 to 6 did not necessarily improve the BIC on the validation set for larger models (see Table \ref{tab:model_selection}).

The top-performing model was $\text{FHMM}(8,4)_2$ across all sex and age groups, except among \emph{Men $<65$}, for whom $\text{FHMM}(6,4)_2$ performed better according to both BIC and AIC. The model $\text{FHMM}(8,4)_2$ corresponds to the largest latent space, capable of encoding up to $4^8 = 65,536$ possible states. In what follows, we will focus on the best performing model for each age and sex group. 
Generally, FHMM models consistently outperform HMM models in terms of BIC.

\subsection{Model analysis}\label{sec:model_analysis}

Next  we provide the analysis of the results for \emph{Women $<65$} subgroup. Other subgroups can be found in the Appendix. The model selected on validation was "fine-tuned" on the testing set, meaning that the variational parameters $\phi$ were newly inferred based on the latter.

\begin{figure}[ht!]
        \centering
     \begin{subfigure}[b]{0.95\textwidth}
        \centering
         \includegraphics[width=1\textwidth]{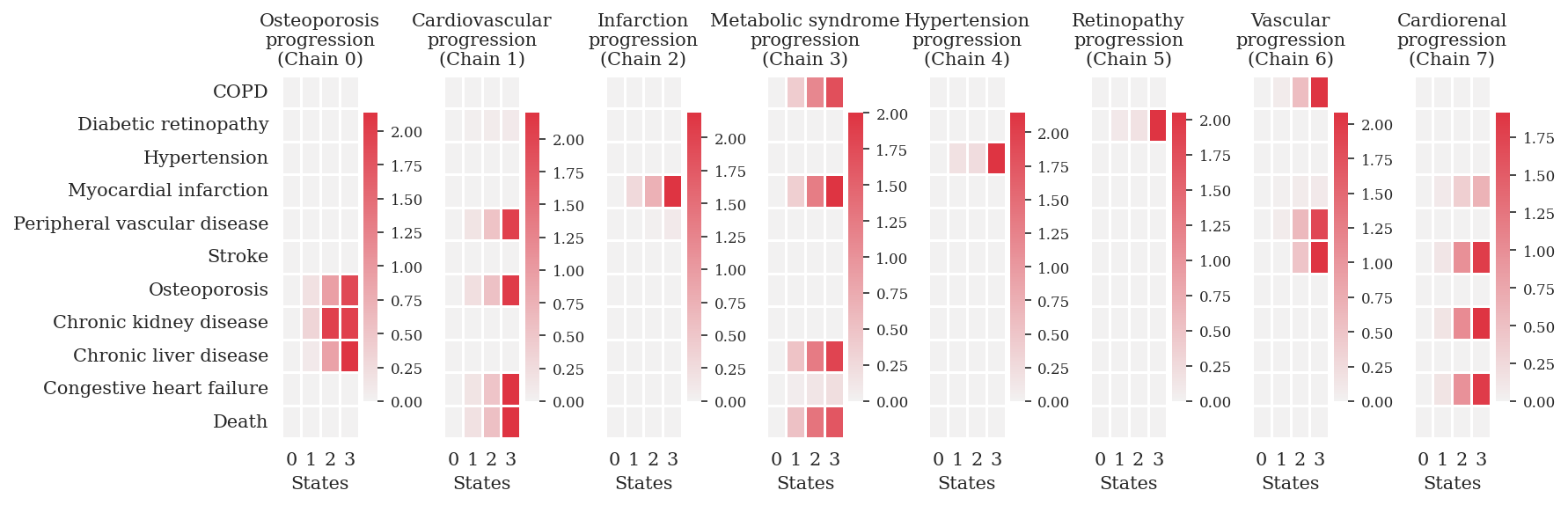}
    \end{subfigure}
    
     \begin{subfigure}[b]{0.9\textwidth}
        \centering
         \includegraphics[width=1\textwidth]{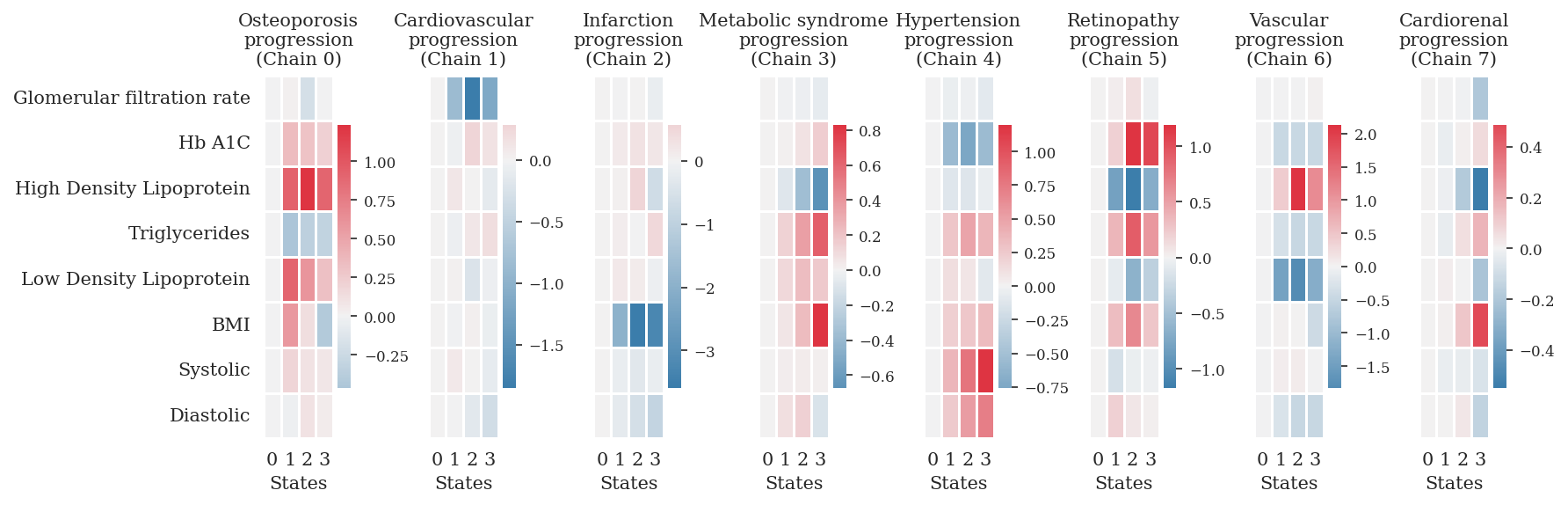}
    \end{subfigure}
    \caption{\textit{Top:} Emission parameters $\eta$ modeling the risk of developing a comorbidity for a given state. Larger values indicate a higher probability of having the comorbidity. The values are reported on a logarithmic scale. \textit{Bottom:} Mean value of lab tests with respect to the intercept (i.e., state 0). Red (resp. blue) indicates an increase (resp. decrease) in $\E[Y_t|H_t]$ relative to the intercept. The model includes 8 chains with 4 states per chain and is fitted on on \textit{Women $<65$}.}
    \label{fig:emission_param_bin}
\end{figure}

\begin{figure}[ht!]
    \centering
    \includegraphics[width=0.4\textwidth]{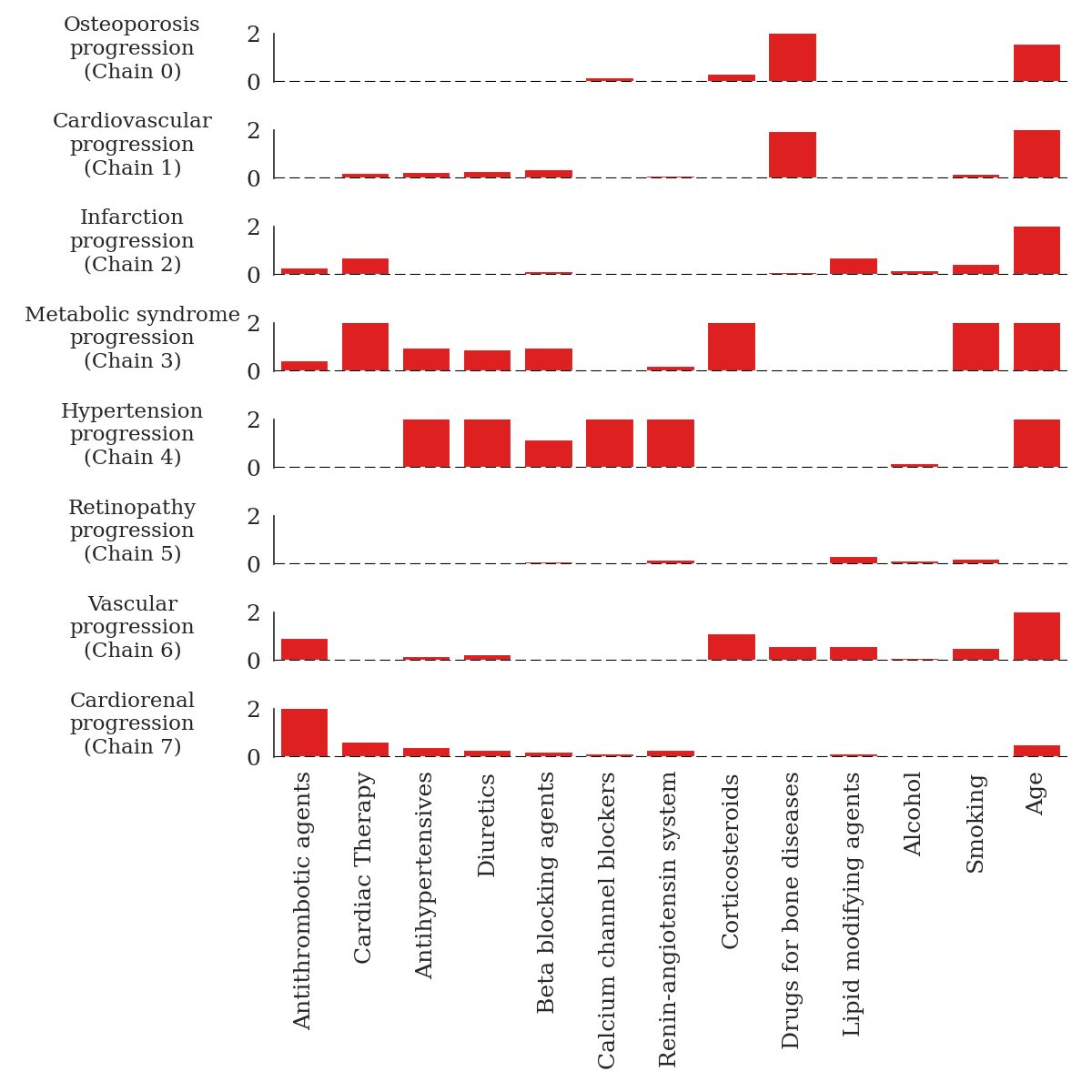}
    \caption{Estimated contribution of drug prescriptions to initial latent state allocation ($W^{\text{init}}$), based on a model with 8 chains and 4 states per chain fitted to the \emph{Women $<65$} subgroup. Higher values indicate a greater probability of starting in a more severe latent state. These estimates capture baseline correlations, not causal effects, since they describe state assignment rather than counterfactual deviation.}
    \label{fig:init_param}
\end{figure}

\paragraph{Emission parameters.}The emission parameters for each latent state are shown in Figure \ref{fig:emission_param_bin}. By construction (Section \ref{sec:structure}), higher states correspond to greater probabilities of specific comorbidities, revealing clusters of conditions that tend to co‐occur:
\begin{itemize}
    \item 
Chain 0, termed "Osteoporosis progression," combines elevated risks of osteoporosis, chronic kidney disease and fatty liver disease with a high‐HDL/LDL and low‐triglyceride profile. 
 \item Chain 1, "Cardiovascular progression," captures macrovascular outcomes (peripheral vascular disease, heart failure, death) and renal failure (low GFR), along with osteoporosis. 
  \item Chain 2, "Infarction progression," is marked by myocardial infarction risk accompanied by BMI reduction. 
   \item  Chain 3, "Metabolic syndrome progression," features obesity, dyslipidemia, chronic obstructive pulmonary disease, liver disease and death risk. 
    \item Chain 4, "Hypertension progression," reflects isolated elevation of systolic and diastolic blood pressure. 
     \item Chain 5, "Retinopathy progression," shows increased diabetic retinopathy probability alongside poor glycemic and lipid control (elevated HbA1c, triglycerides and BMI, reduced HDL). 
      \item Chain 6, "Vascular progression," combines stroke, peripheral vascular disease and COPD with a comparatively favorable lipid profile. 
       \item Chain 7, "Cardiorenal progression," integrates myocardial infarction, stroke, chronic heart failure, chronic kidney disease and dyslipidemia with obesity. Hypertension and diabetic retinopathy tend to develop independently of other comorbidities in our model.
\end{itemize}

 \paragraph{Initial latent state.} The relationship between the initial latent state and the input variables is shown in Figure \ref{fig:init_param}. Positive values (in red) correspond to a higher probability of beginning in a more severe state. Smoking shows a marked positive value in chain 3, reflecting its association with increased COPD risk. The age parameter is positive across several chains, indicating that older patients tend to start in worse health states. This pattern is not observed in chain 5, where age has no significant effect on the initial state. Receiving treatment for bone diseases is associated with an increased likelihood of being in the osteoporosis state at the start.

\begin{table}[!ht]
    \centering
\resizebox{\textwidth}{!}{
\begin{tabular}{l|ccccccccccccc}
    \toprule
Cluster & 0 & 1 & 2 & 3 & 4 & 5 & 6 & 7 & 8 & 9 & 10 & 11 & 12\\
    \midrule
\# of patients & 3373 & 1468 & 1549 & 792 & 2774 & 1491 & 1101 & 1431 & 1006 & 1421 & 594 & 1143 & 1362\\
Avg. age at NIAD (y) $\pm$ SD & 47.01 $\pm$ 10.68 & 47.1 $\pm$ 10.3 & 54.01 $\pm$ 7.19 & 57.49 $\pm$ 6.09 & 54.27 $\pm$ 7.5 & 54.53 $\pm$ 8.24 & 54.61 $\pm$ 7.71 & 55.52 $\pm$ 7.38 & 56.12 $\pm$ 7.1 & 53.19 $\pm$ 7.93 & 57.12 $\pm$ 6.27 & 48.75 $\pm$ 9.98 & 55.23 $\pm$ 7.86\\
Avg. follow-up post-NIAD (y) $\pm$ SD & 3.73 $\pm$ 2.68 & 4.52 $\pm$ 2.98 & 5.88 $\pm$ 3.15 & 4.87 $\pm$ 2.99 & 4.63 $\pm$ 3.13 & 5.23 $\pm$ 3.28 & 5.23 $\pm$ 3.19 & 6.21 $\pm$ 3.27 & 5.37 $\pm$ 3.24 & 5.81 $\pm$ 3.25 & 6.79 $\pm$ 3.23 & 5.08 $\pm$ 3.0 & 6.52 $\pm$ 3.35\\
Avg. BMI at NIAD (kg/m\textsuperscript{2}) $\pm$ SD & 33.15 $\pm$ 7.55 & 37.25 $\pm$ 7.79 & 34.07 $\pm$ 6.16 & 35.74 $\pm$ 7.81 & 34.54 $\pm$ 6.94 & 34.31 $\pm$ 6.73 & 35.04 $\pm$ 6.93 & 40.96 $\pm$ 9.87 & 36.64 $\pm$ 7.38 & 38.35 $\pm$ 7.72 & 31.23 $\pm$ 6.45 & 32.96 $\pm$ 6.62 & 38.04 $\pm$ 8.34\\
Avg. HbA1c at NIAD (\%) $\pm$ SD & 6.83 $\pm$ 1.56 & 7.26 $\pm$ 1.66 & 7.48 $\pm$ 1.61 & 7.24 $\pm$ 1.47 & 6.89 $\pm$ 1.37 & 7.15 $\pm$ 1.49 & 7.25 $\pm$ 1.48 & 7.51 $\pm$ 1.61 & 7.35 $\pm$ 1.5 & 7.05 $\pm$ 1.46 & 7.16 $\pm$ 1.59 & 7.94 $\pm$ 1.83 & 6.92 $\pm$ 1.37\\
\% Ever smoked & 23.04 & 79.56 & 20.46 & 78.41 & 20.01 & 29.51 & 43.23 & 56.81 & 55.07 & 59.25 & 30.13 & 24.41 & 50.07\\
\% Ever alcohol & 70.53 & 82.63 & 85.15 & 85.61 & 78.98 & 79.61 & 82.29 & 84.35 & 82.9 & 86.77 & 82.15 & 82.06 & 82.53\\
    \midrule
\multicolumn{14}{l}{\textit{Comorbidities at NIAD [N,(\%)]}}\\
Stroke & $<1$ \% & $<1$ \% & $<1$ \% & $<1$ \% & $<1$ \% & 294 (19.72) & $<1$ \% & 178 (12.44) & $<1$ \% & $<1$ \% & $<1$ \% & $<1$ \% & 104 (7.64)\\
Myocardial infarction & $<1$ \% & $<1$ \% & $<1$ \% & 16 (2.02) & $<1$ \% & $<1$ \% & 12 (1.09) & 158 (11.04) & $<1$ \% & $<1$ \% & $<1$ \% & $<1$ \% & 171 (12.56)\\
Congestive heart failure & $<1$ \% & $<1$ \% & $<1$ \% & $<1$ \% & $<1$ \% & 49 (3.29) & $<1$ \% & 89 (6.22) & $<1$ \% & $<1$ \% & $<1$ \% & $<1$ \% & 50 (3.67)\\
Peripheral vascular disease & $<1$ \% & $<1$ \% & $<1$ \% & 43 (5.43) & $<1$ \% & 19 (1.27) & $<1$ \% & 25 (1.75) & $<1$ \% & $<1$ \% & 65 (10.94) & $<1$ \% & $<1$ \%\\
Diabetic retinopathy & $<1$ \% & 116 (7.90) & 520 (33.57) & 118 (14.90) & $<1$ \% & 213 (14.29) & 188 (17.08) & 436 (30.47) & 191 (18.99) & 69 (4.86) & 67 (11.28) & 445 (38.93) & $<1$ \%\\
Hypertension & 79 (2.34) & 23 (1.57) & 1155 (74.56) & 384 (48.48) & 2230 (80.39) & 821 (55.06) & 593 (53.86) & 938 (65.55) & 769 (76.44) & 1113 (78.33) & 288 (48.48) & $<1$ \% & 843 (61.89)\\
Chronic kidney disease & $<1$ \% & $<1$ \% & $<1$ \% & $<1$ \% & $<1$ \% & 399 (26.76) & 18 (1.63) & 191 (13.35) & $<1$ \% & $<1$ \% & $<1$ \% & $<1$ \% & 170 (12.48)\\
Osteoporosis & $<1$ \% & $<1$ \% & $<1$ \% & $<1$ \% & $<1$ \% & 35 (2.35) & 65 (5.90) & 20 (1.40) & $<1$ \% & $<1$ \% & 150 (25.25) & $<1$ \% & $<1$ \%\\
COPD & $<1$ \% & $<1$ \% & $<1$ \% & 497 (62.75) & $<1$ \% & $<1$ \% & 78 (7.08) & 149 (10.41) & $<1$ \% & $<1$ \% & 6 (1.01) & $<1$ \% & 82 (6.02)\\
Chronic liver disease & $<1$ \% & $<1$ \% & $<1$ \% & $<1$ \% & $<1$ \% & $<1$ \% & 531 (48.23) & 65 (4.54) & $<1$ \% & $<1$ \% & $<1$ \% & $<1$ \% & 39 (2.86)\\
    \midrule
\multicolumn{14}{l}{\textit{Comorbidities during follow-up [N,(\%)]}}\\
Stroke & $<1$ \% & $<1$ \% & $<1$ \% & 10 (1.26) & $<1$ \% & 102 (6.84) & 14 (1.27) & 59 (4.12) & $<1$ \% & $<1$ \% & 14 (2.36) & $<1$ \% & 58 (4.26)\\
Myocardial infarction & $<1$ \% & $<1$ \% & $<1$ \% & 10 (1.26) & $<1$ \% & $<1$ \% & 11 (1.00) & 27 (1.89) & $<1$ \% & 20 (1.41) & $<1$ \% & $<1$ \% & 29 (2.13)\\
Congestive heart failure & $<1$ \% & $<1$ \% & $<1$ \% & $<1$ \% & $<1$ \% & 31 (2.08) & $<1$ \% & 59 (4.12) & $<1$ \% & $<1$ \% & 6 (1.01) & $<1$ \% & 55 (4.04)\\
Peripheral vascular disease & $<1$ \% & $<1$ \% & $<1$ \% & 20 (2.53) & $<1$ \% & $<1$ \% & $<1$ \% & $<1$ \% & $<1$ \% & $<1$ \% & 22 (3.70) & $<1$ \% & $<1$ \%\\
Diabetic retinopathy & 163 (4.83) & 224 (15.26) & 610 (39.38) & 166 (20.96) & 202 (7.28) & 283 (18.98) & 222 (20.16) & 462 (32.29) & 219 (21.77) & 247 (17.38) & 153 (25.76) & 486 (42.52) & 176 (12.92)\\
Hypertension & 126 (3.74) & 36 (2.45) & 175 (11.30) & 66 (8.33) & 309 (11.14) & 138 (9.26) & 70 (6.36) & 76 (5.31) & 76 (7.55) & 148 (10.42) & 72 (12.12) & 40 (3.50) & 89 (6.53)\\
Chronic kidney disease & $<1$ \% & $<1$ \% & $<1$ \% & 13 (1.64) & $<1$ \% & 166 (11.13) & 39 (3.54) & 124 (8.67) & $<1$ \% & $<1$ \% & 21 (3.54) & $<1$ \% & 88 (6.46)\\
Osteoporosis & $<1$ \% & $<1$ \% & $<1$ \% & 15 (1.89) & $<1$ \% & 19 (1.27) & 21 (1.91) & $<1$ \% & $<1$ \% & $<1$ \% & 50 (8.42) & $<1$ \% & $<1$ \%\\
COPD & $<1$ \% & 31 (2.11) & $<1$ \% & 136 (17.17) & $<1$ \% & 15 (1.01) & 49 (4.45) & 94 (6.57) & 29 (2.88) & 46 (3.24) & 11 (1.85) & $<1$ \% & 58 (4.26)\\
Chronic liver disease & $<1$ \% & 28 (1.91) & $<1$ \% & 8 (1.01) & $<1$ \% & 16 (1.07) & 296 (26.88) & 75 (5.24) & 26 (2.58) & 36 (2.53) & 19 (3.20) & 14 (1.22) & 49 (3.60)\\
Deaths before 2020& $<1$ \% & 42 (2.86) & $<1$ \% & 79 (9.97) & $<1$ \% & 21 (1.41) & 60 (5.45) & 132 (9.22) & 148 (14.71) & 30 (2.11) & 12 (2.02) & $<1$ \% & 93 (6.83)\\
    \bottomrule
\end{tabular}
}
     \caption{Patient characteristics in each cluster for the \textit{Women $<65$} cohort with average values and standard deviation where applicable. Comorbidities are presented as absolute counts at NIAD initiation, along with the number of new comorbidities developed post-NIAD. Proportions, expressed as percentages of the cohort population, are reported in parentheses.}
    \label{tab:cluster_descr}
\end{table}

\begin{figure}[!h]
    \centering
    \includegraphics[width=0.8\linewidth]{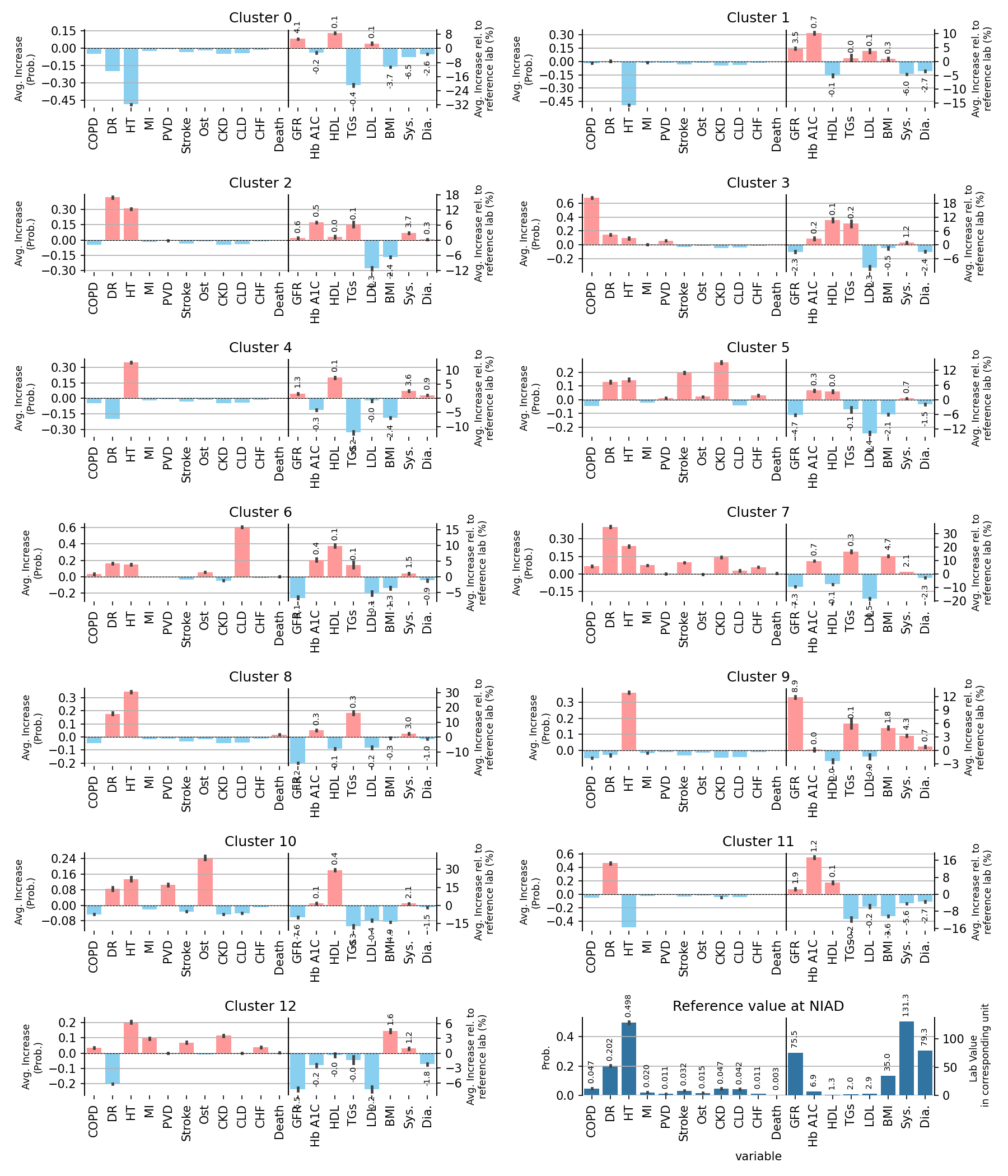}
    \caption{
    Average cluster-specific probabilities and lab values for \textit{Women $<65$}. Values have been centered by subtracting the average at the time of NIAD initiation to improve readability. Red bars indicate increases, and blue bars indicate decreases relative to the average patient at NIAD. Comorbidity risks are shown as changes in probability (left axis), and lab values as percentage changes relative to the reference value (right axis). Annotated numbers indicate absolute changes in the original units (see Table \ref{tab:list_of_var}). Error bars represent the standard error of the mean.}
    \label{fig:female_m65_proba_avg}
\end{figure}

\paragraph{State Clustering.}
Following Section \ref{sec:trajectory_analysis}, we identified 13 clusters in the latent representation according to the selection criterion (see Figure \ref{fig:cluster_selection_women<65} in the Appendix). Table \ref{tab:cluster_descr} details the baseline characteristics of each cluster at NIAD initiation, and Figure \ref{fig:female_m65_proba_avg} presents their time-averaged clinical profiles.

Several clusters demonstrate similarities at NIAD initiation. Clusters 0, 1, and 11 share characteristics of younger patients (age $<50$) with low comorbidity burden. However, their clinical trajectories differ significantly - particularly cluster 1, which despite its younger profile shows higher smoking prevalence, elevated BMI, and increased mortality risk (see Table \ref{tab:cluster_descr}).

The clusters can be divided based on their mortality risk. At one end of the spectrum, clusters 0, 2, 4, 5, and 11 form a lower-risk group characterized by relatively low death rates. An intermediate category includes clusters 1, 9, and 10, which show progressing disease with moderate mortality risk. The highest-risk group comprises clusters 3, 6, 7, 8, and 12, showing substantially elevated mortality rates that indicate advanced disease states.
Within this high-risk category, cluster 7 is particularly notable:  it is defined by the highest BMI (morbid obesity with BMI $>40$) and carries an exceptionally high comorbidity burden, indicating particularly severe disease progression.

The healthiest subgroup, cluster 0 ("young low-risk"), is characterized by preserved metabolic and renal function and low cardiovascular risk. Cluster 3 ("pulmonary-metabolic") combines metabolic dysfunction and pulmonary disease with high smoking prevalence and renal impairment. This cluster's profile, with increased risk in nearly all comorbidities, positions it as a transition toward severe multimorbidity. On the other hand, cluster 10 ("bone-vascular") features high osteoporosis prevalence and increased peripheral vascular disease, suggesting shared pathophysiological pathways between skeletal and vascular systems.

Among the most severe phenotypes, cluster 8 ("advanced nephropathy") shows severe GFR loss, persistent hypertension, and high mortality. Cluster 12 ("cardio-pulmonary multimorbidity") is defined by COPD, CHF, and elevated stroke risk. Cluster 6 ("hepatorenal syndrome") represents a very severe situation, linking chronic liver disease, renal dysfunction (low GFR), and poor glycemic control, consistent with fatty liver disease trends \cite{younossi2016global}.

\begin{figure}[ht]
    \centering
    \includegraphics[width=0.35\linewidth]{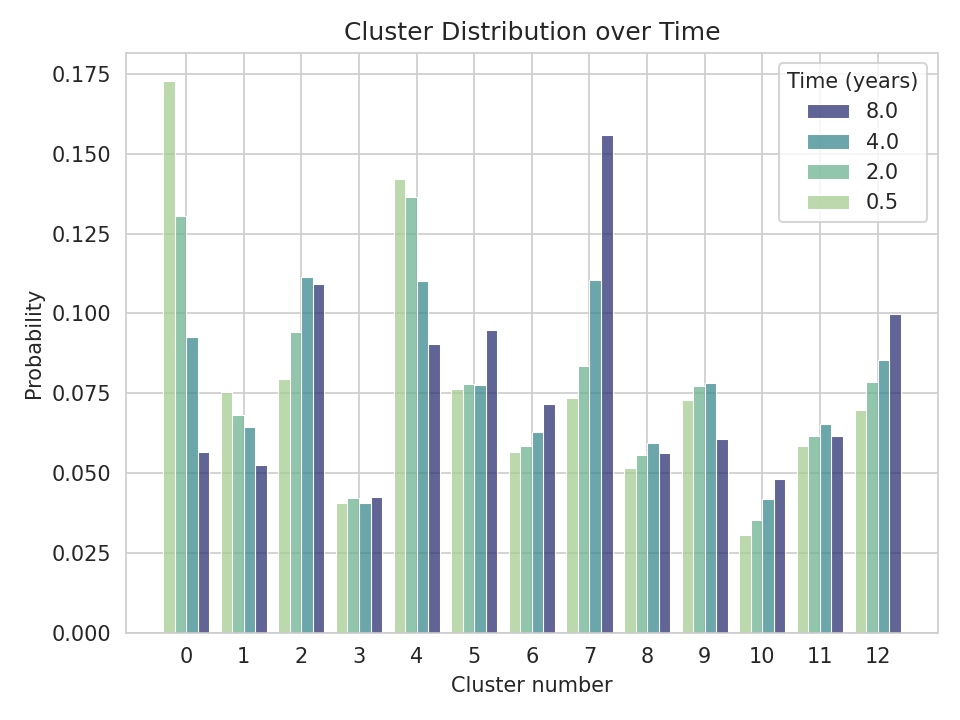}
    \caption{Distribution of the clusters over time.}
    \label{fig:female_m65_distribution}
\end{figure}

\paragraph{State progression.}
The temporal distribution of the clusters is shown in Figure \ref{fig:female_m65_distribution}. Clusters associated with advanced or progressing disease (e.g., cluster 7 and transitional clusters 2, 6, 10, 12) gain members over time, while healthier clusters (0, 1, 4) decline. 

\begin{figure}[!ht]
    \centering
    \includegraphics[width=0.8\textwidth]{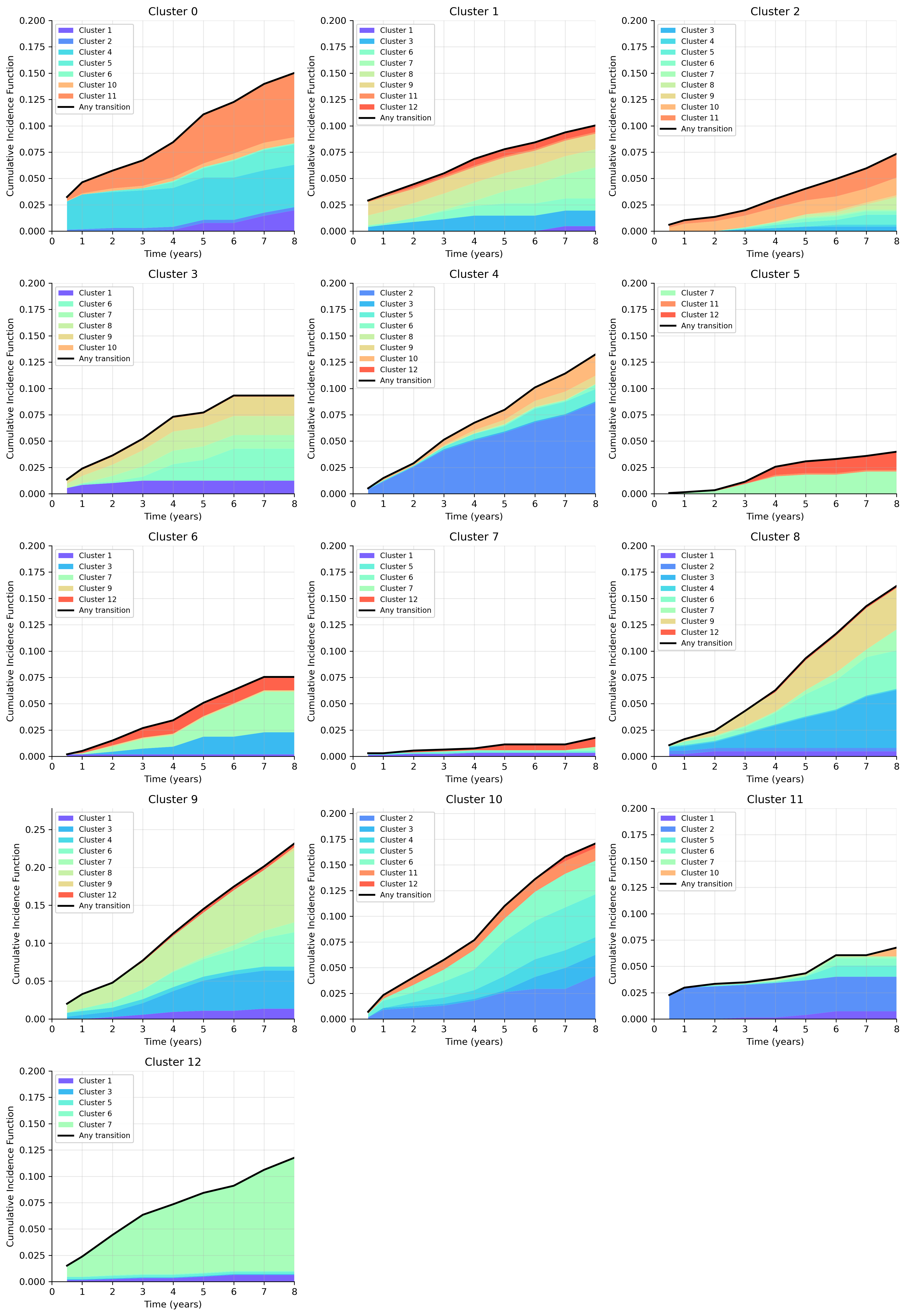}
    \caption{
Cumulative Incidence Function (CIF) for the time to first cluster transition in \textit{Women $<65$}, stratified by initial cluster assignment at 6 months post-NIAD.
The estimates account for censoring due to loss to follow-up and study termination, computed using the Aalen nonparametric estimator \cite{aalen1978}.
}
    \label{fig:female_m65_transition_prob}
\end{figure}

Figure \ref{fig:female_m65_transition_prob} presents the cumulative incidence functions for time to first cluster transition, stratified by initial cluster assignment. The transition dynamics are further illustrated in the appendix (Figure \ref{fig:female_m65_evolution}). Patients in cluster 0 often progress into early microvascular states, particularly cluster 11 or cluster 4, defined by higher prevalence of diabetic retinopathy and hypertension, respectively. After approximately four years, these states may progress to early multimorbidity (cluster 5) or a combination of both diseases (cluster 2), as confirmed by Figure \ref{fig:female_m65_transition_prob} showing most cluster 4 patients transitioning to cluster 2.
In contrast, cluster 1, despite representing younger patients, exhibits higher mortality risk and demonstrates progression to advanced disease states. These patients typically evolve into clusters 6 or 7 in later stages, indicating more severe disease trajectories.

Cluster 10 (bone-vascular) represents an intermediate severity profile. Patients in this cluster tend to progress toward broader multimorbidity through two primary pathways: either via clusters 2 or 5, or occasionally involving hepatic complications (cluster 6).
The advanced disease profiles represented by cluster 12 (cardio-pulmonary multimorbidity) typically transition to cluster 7, the most severe multimorbid state, that serves as a frequent absorbing state. Similarly, clusters 2, 5, and 11 demonstrate low transition probabilities, functioning as stable states where patients tend to remain over time (at least when starting in those clusters).

\begin{figure}[!ht]
    \centering
    \includegraphics[width=0.9\linewidth]{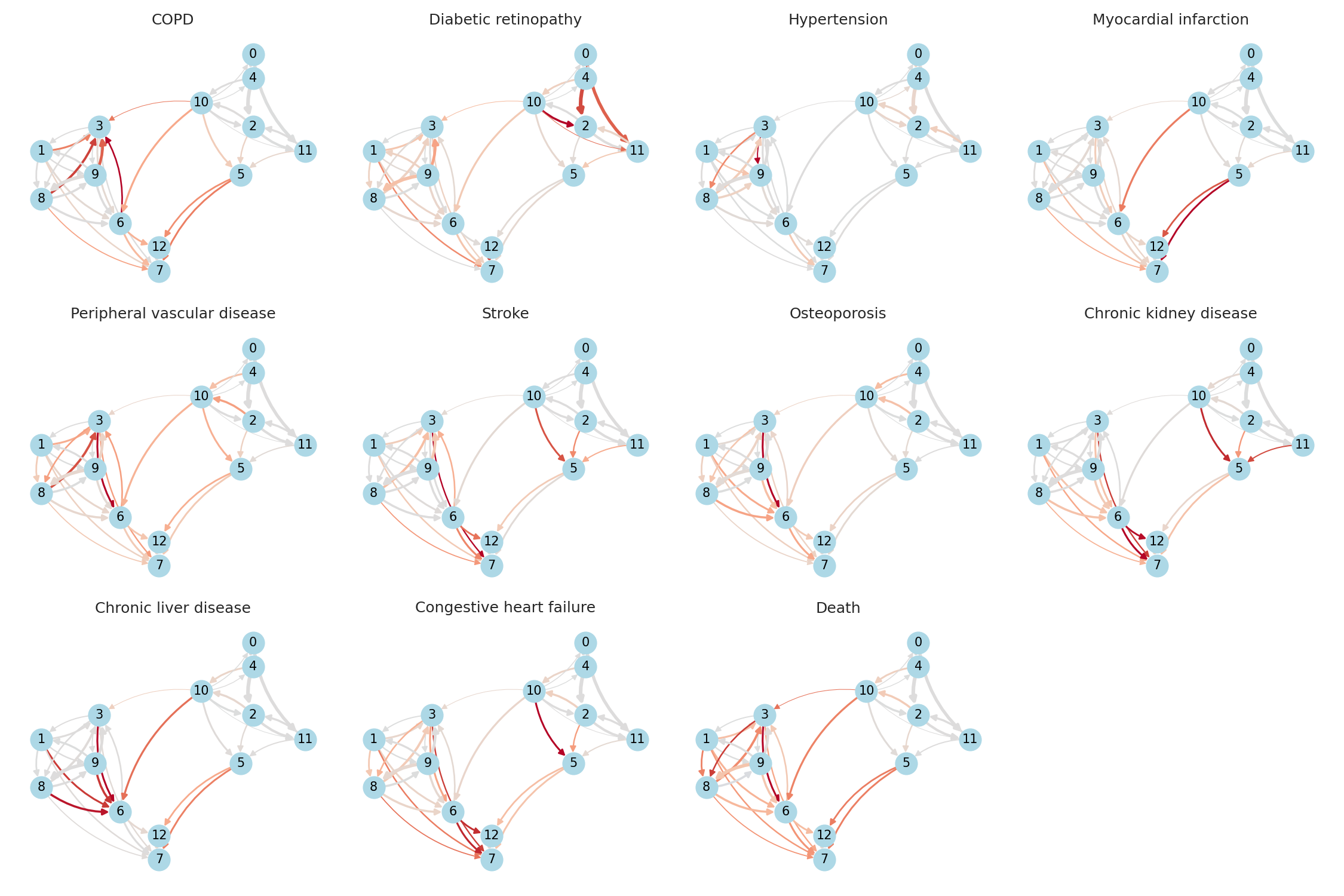}
    \caption{Transition graph. Red arrows (resp. blue) means an increase (resp. decrease) of the comorbidity risk (\textit{Women $<65$}). Arrow thickness is proportional to the number of observed transitions. Transitions with less than 1\% probability are not shown.}
    \label{fig:female_m65_graph}
\end{figure}

The transition graph in Figure \ref{fig:female_m65_graph} (showing only transitions with less than 1\% probability) reveals that clusters 0, 2, 4, 5, 10, and 11 form a transient subgraph. This configuration indicates that once patients transition out of these clusters, they rarely return, suggesting a unidirectional progression toward more severe disease states. The observed pattern reflects the clinical reality that as patients develop serious comorbidities, their condition tends to evolve toward increasingly complex and severe multimorbidity profiles rather than reverting to less severe states.

\section{Discussion}

We identified eight latent components that capture distinct patterns of comorbidity in type 2 diabetes in women under 65 years of age (with additional population analyses provided in the Appendix).
At the interpretation level, the inferred latent chains correspond to recognizable diabetes-related complications, spanning microvascular, macrovascular, metabolic, skeletal, pulmonary, and cardiorenal involvement. Several components reflect well-established comorbidity patterns in T2DM; for example, chains dominated by macrovascular and renal outcomes align with the known coupling between cardiovascular disease and chronic kidney disease in diabetes, often described as a cardiorenal syndrome \cite{ronco2008cardiorenal}. Similarly, components emphasizing microvascular complications, such as retinopathy and hypertension, appear to develop independently of other complications and follow expected clinical patterns, including associations with elevated blood pressure and poor glycemic and lipid control, respectively \cite{Cheung2010, lee2015epidemiology}. The skeletal involvement in our osteoporosis-related component co-occurred with chronic kidney and liver disease, reflecting the clustering of multisystem complications in T2DM. This pattern is consistent with the understanding that diabetes affects multiple organ systems through interconnected metabolic and vascular pathways \cite{Vestergaard2007,Napoli2017}.

While many of these associations are individually well documented, our model learns them jointly from longitudinal data without prespecifying disease groupings or assuming independence between outcomes. This yields an interpretable decomposition of multimorbidity into parallel disease processes that may co-occur and progress at different rates within individuals.

At the patient level, clustering of latent-state trajectories highlights how these components combine into progression pathways. Patients often transition from relatively healthy or low-risk metabolic states (clusters 0, 2, 4, 5, 10, 11) toward advanced multimorbid states, with clusters 6 and 7 representing terminal disease profiles associated with elevated mortality. These trajectories broadly separate into at least two modes of progression: one dominated by microvascular complications (clusters 2, 4, 5, 10, 11), and another driven by pulmonary, hepatic, renal, and cardio-pulmonary involvement (clusters 3, 6, 8, 9, 12).

From a methodological perspective, our model reveals patterns that are difficult to extract with alternative approaches. While static clustering approaches summarize patient similarity but do not encode temporal dynamics \cite{martinez2022comorbidity}, and deep generative models can learn flexible latent representations from longitudinal clinical data but often require additional assumptions or post hoc alignment to relate latent dimensions to clinical concepts \cite{trottet2023generative}, our approach provides a structured factorization that enables direct clinical interpretation. Classical HMM-based models capture temporal dynamics but often rely on a single latent process, potentially mixing distinct disease mechanisms \cite{alaa2017learning,wang2014unsupervised}. The factorial structure adopted here enables multimorbidity to be represented as the superposition of multiple latent chains, providing a natural framework for chronic diseases characterized by overlapping and asynchronous complications.

This study has several limitations. Real-world electronic health records are collected based on routine clinical care, resulting in unequally spaced measurements that often required forward imputation to maintain complete information at each time-point. Some variables, particularly BMI and other lifestyle variables, exhibit missingness that is likely not missing at random. While forward imputation avoids using future information, alternative imputation schemes might be more appropriate for handling missing values. The current model uses Gaussian emissions for scalability, though extensions to other distributions (e.g., count-based likelihoods) are possible. The data were grouped into binned time steps, imposing an assumption of regular observation intervals that does not reflect the irregular nature of real-world measurements. The analysis was restricted to a relatively small set of commonly occurring chronic comorbidities and laboratory values, excluding less frequent conditions that might contribute to multimorbidity patterns. Issues of data imbalance persist, as many comorbidities represent rare events, and treatment adherence was not explicitly modeled, adding uncertainty to treatment-related effects. Finally, the requirement for HbA1c measurements prior to the index date may introduce selection bias, as tested individuals are more likely to have suspected or established comorbidities. As a result, the study cohort may represent a systematically sicker subset of the broader population, potentially limiting generalizability.

\section{Conclusion}

Our study demonstrates that comorbidities in T2DM can be effectively modeled as the interaction of multiple latent disease processes evolving over time. The structural FHMM approach identified eight coherent latent components that explain the different disease trajectories, providing insights into the complex progression patterns of T2DM and its complications.

The advantages of our approach include its ability to learn meaningful patterns of comorbidities jointly from longitudinal data, without requiring pre-specified disease groupings. This makes it broadly applicable, particularly in contexts where prior knowledge is unavailable or cannot be encoded. The model captures temporal dynamics often missed by static clustering approaches and delivers more interpretable results compared to deep generative models. The factorial structure allows for the representation of comorbidities as a superposition of multiple concurrent disease processes, offering a better understanding of disease progression than single-process HMMs.

Looking forward, several directions could improve the clinical utility of this approach. While our model effectively characterizes T2DM progression patterns, real-time clinical implementation would require incorporation of additional comorbidities beyond the core set analyzed. Although prescribed medications were included in our initial analysis, they did not significantly improve model performance, likely due to challenges in capturing complex treatment patterns from EHR data. Future work could explore more sophisticated feature engineering of medication data. Nevertheless the interpretability demonstrated by our approach suggests strong potential for integration into real-world clinical workflows, where it could provide valuable insights into disease progression patterns and inform personalized care strategies for T2DM patients.

\section{Funding}
This research was funded by a Swiss Data Science Centre Collaboration Grant (C19-09).
\clearpage
\bibliographystyle{ieeetr} 
\bibliography{references}
\clearpage

\appendix

\section{Data preparation}\label{sec:data_preparation}
\begin{table}[ht]
    \centering
    \begin{tabular}{lccc}
    \toprule
    Lab Value & Low & High & Comment\\
    \midrule
         Glomerular filtration rate &0&150& - \\
         Hb A1C& 0&16&-\\
         High Density Lipoprotein &0&5&-\\
         Low Density Lipoprotein &0&10&-\\
         Triglycerides&0&15&-\\
         BMI &0&80&-\\
         Diastolic&30&150&Lower than systolic\\
         Systolic&80&240&Higher than diastolic\\
         \bottomrule
    \end{tabular}
    \caption{Predefined thresholds used to remove outliers. Values are removed if they are not included in the lower and upper limits.}
    \label{tab:thresholds}
\end{table}

\begin{table}[ht]
    \centering
    \footnotesize
    \begin{tabular}{l|rrr|rrr|rrr|rrr}
\toprule
 & \multicolumn{3}{c}{Men $< 65$} & \multicolumn{3}{c}{Men $65 +$} & \multicolumn{3}{c}{Women $< 65$} & \multicolumn{3}{c}{Women $65 +$} \\
 & Train & Val & Test & Train & Val & Test & Train & Val & Test & Train & Val & Test \\
\midrule
COPD & 7.7 & 7.6 & 8.0 & 16.4 & 16.1 & 17.0 & 7.8 & 7.9 & 7.9 & 13.3 & 12.8 & 15.6 \\
Diabetic retinopathy  & 32.8 & 33.4 & 32.9 & 34.8 & 36.3 & 36.4 & 30.4 & 31.5 & 31.1 & 34.6 & 33.8 & 35.7 \\
Hypertension & 59.5 & 59.7 & 61.3 & 71.2 & 71.0 & 71.9 & 56.9 & 57.0 & 57.1 & 78.6 & 79.9 & 79.2 \\
Myocardial infarction & 8.8 & 9.2 & 9.0 & 16.6 & 17.1 & 18.5 & 3.0 & 2.8 & 2.8 & 7.9 & 6.8 & 7.3 \\
 Peripheral vascular disease & 4.2 & 3.8 & 4.5 & 11.0 & 11.1 & 10.5 & 1.7 & 1.8 & 1.6 & 5.5 & 6.2 & 5.7 \\
Stroke & 6.2 & 6.4 & 6.7 & 17.6 & 17.6 & 17.6 & 5.1 & 5.1 & 5.0 & 15.8 & 14.2 & 14.4 \\
Osteoporosis & 0.7 & 0.6 & 0.7 & 2.2 & 2.3 & 2.1 & 2.9 & 3.1 & 2.6 & 11.8 & 11.8 & 11.8 \\
Chronic kidney disease  & 5.9 & 5.9 & 5.9 & 11.5 & 12.7 & 10.7 & 6.9 & 6.5 & 6.5 & 10.3 & 9.1 & 9.6 \\
Chronic liver disease  & 4.8 & 4.8 & 4.8 & 2.7 & 2.4 & 3.3 & 6.2 & 6.2 & 6.2 & 2.9 & 3.5 & 3.3 \\
Congestive heart failure   & 4.5 & 4.5 & 4.5 & 13.3 & 14.0 & 14.2 & 2.3 & 2.2 & 2.4 & 10.6 & 10.6 & 10.5 \\
Death & 5.2 & 5.4 & 5.2 & 22.3 & 22.2 & 22.2 & 4.1 & 4.1 & 3.9 & 19.4 & 19.1 & 20.1 \\
\bottomrule
\end{tabular}
    \caption{Number of patient diagnosed with the comorbidity for each data split. Numbers are given in percentage.}
    \label{tab:proportion_split}
\end{table}

\section{Variational approximation}\label{sec:variational}

This section provides details on the fitting algorithm used to train the FHMM. The fitting algorithm mainly follows \cite{ghahramani1995factorial}, where the authors reduce the inference complexity to $O(K^2MT)$ by exploiting a structured mean field variational approximation \cite{Wainwright08} for the posterior distribution $q_\phi$ used in the E step of their EM procedure. In particular, due to their specific choice, \eqref{eqn:elbo} can be written as 
\begin{equation*}
   F(q_\phi,\theta) = \E_{q_\phi}\left[\sum_{t=1}^T \log p_\theta\left(y_t|\tilde{H}_t^{(1:M)}\right) -\sum_{t=1}^T\sum_{m=1}^M (\tilde{H}_{t}^{(m)})^\top \log e_t^{(m)} \right] - Z + Z_{q_\phi},
\end{equation*}
where $Z$ and $Z_{q_\phi}$ are respectively normalizing constants of $p_\theta(y_{1:T}|h_{1:T},x_{1:T})$ and $q_\phi(h_{1:T}\mid x_{1:T})$ and where $\tilde{H}_t^{(m)}$ is a one-hot encoded version of $H_t^{(m)}$.

The E step requires the optimization of the parameters $\phi$, which can be computing by solving 
\begin{equation*}
        \left. \frac{\partial F}{\partial \log e_\tau^{(n)}}\right|_{q_{\hat{\phi}}, \theta}=0,
\end{equation*}
where the gradient is 
\begin{equation}\label{eqn:gradient_e_step}
    \left. \frac{\partial F}{\partial \log e_\tau^{(n)}}\right|_{q_{{\phi}}, \theta}=\sum_{t=1}^T \E_{q_\phi}\left[\left\{ \log p_\theta\left(y_t|\tilde{H}_t^{(1:M)}\right)  - \sum_{m=1}^M \left(\tilde{H}_{t}^{(m)}\right)^\top \log e_t^{(m)}\right\}\right. \left. \times \left\{{\tilde{H}_\tau^{(n)}}-\E_{q_\phi}\left[\tilde{H}_\tau^{(n)}\right]\right\}\right].
\end{equation}

The M step requires the optimization of $\E_{q_\phi}\left[\log p_\theta(y_{1:T}, h_{1:T}) \right]$ with respect to $\theta^{\pi},\theta^{P},\eta$. Each parameter can be optimized individually by solving
\begin{align}
   \hat{\theta}^{\pi}&=\argmax_{\theta^{\pi}} \sum_m  {\gamma_1^{(m)}}^\top \log \pi_{\theta^{\pi}}^{(m)}, \nonumber\\
    \hat{\theta}^{P}&=\argmax_{\theta^P} \sum_m \sum_t \log P_{\theta^P}^{(m)}\Tr(\xi_t^{(m)}), \nonumber\\
    \hat{\eta} &=\argmax_{\eta} \sum_t \E_{q_\phi}\left[\log p_\eta(y_t\mid {h_t^{(1:M)})} \right],\label{eqn:m_step_opt}
\end{align}
where
\begin{align*}
    \gamma_t^{(m)}&\df\E_{q_\phi}\left[\tilde{H}_t^{(m)} \right] \in \R^K,\\
    \xi_t^{(m)}&\df\E_{q_\phi}\left[\tilde{H}_t^{(m)} {\tilde{H}_{t-1}^{(m)}\ }^\top \right] \in \R^{K\times K},
\end{align*}
are computed using the Baum–Welch algorithm for HMMs (see \cite{ghahramani1995factorial}), whose complexity is $O(K^2T)$.

Equations \eqref{eqn:gradient_e_step} and \eqref{eqn:m_step_opt} involves expectations that are expensive to compute. In particular, it requires the computation of $q(H_{t_1}^{(m)},H_{t_2}^{(m)}\mid x_{1:T})$ for any $t_1,t_2$, which has a memory and computational complexity of $O(MT^2K^2)$. Inspired by the Gaussian case, for which analytical expressions exists (see \cite{ghahramani1995factorial}), a second order approximation of the emission log-likelihood $\log p_\eta(y_t \mid {h_t^{(1:M)}})$ is derived using the inequality \eqref{eqn:lower_bound_elbo}.
In that case, the gradient is reduced to
\begin{multline}\label{eqn:grad_tanh}
   \left. \frac{\partial F}{\partial \log e_\tau^{(n)}}\right|_{q, \theta}\approx \sum_t \cov\left(\tilde{H}_\tau^{(n)},\tilde{H}_t^{(n)}\right)\times\left\{y_t\eta^{(n)}-\eta^{(n)}\left(\frac{1}{2}+\lambda(z)\eta^{(n)}\eta^\top\gamma_t\right)\right.\\\left.-\frac{\lambda(z)}{2}\left(\diag(\eta^{(n)}{\eta^{(n)}}^\top)-2\eta^{(n)}{\eta^{(n)}}^\top \gamma^{(n)}\right)-\log e_t^{(n)}\right\}.
\end{multline}
Finding the zeros of \eqref{eqn:grad_tanh} (E step) can then be easily done without computing the expensive terms $\cov\left(\tilde{H}_\tau^{(n)},\tilde{H}_t^{(n)}\right)$.
On the other hand, the M step for binary outputs becomes
\begin{multline}\label{eqn:m_step_tanh}
   \max_\eta \E_{q_\phi}\left[\log p_\eta(y_t\mid h_t^{(1:M)}) \right]=  \max_\eta \E_{q_\phi}\left[y_t\tilde{H}_t^\top\eta - \log(1+\exp(\tilde{H}_t^\top\eta ))\right]\geq\\  \max_\eta \left(y_t-\frac{1}{2}\right)\gamma_t^\top \eta - \log\left\{1+\exp(-z)\right\} - z/2 - \frac{\lambda(z)}{2}\left(\eta^\top \E\left[\tilde{H}_t{\tilde{H}_t}^\top\right] \eta-z^2\right),
\end{multline}
where $\tilde{H}_t=\left(\tilde{H}_t^{(1)},\dots,\tilde{H}_t^{(M)}\right)$ and $\gamma_t=\left(\gamma_t^{(1)},\dots,\gamma_t^{(M)}\right)$.
Note that Equation \ref{eqn:m_step_tanh} is a lower bound of the ELBO, which can be optimized with resepect to $z$, which admits a closed form solution
$$z_t=\sqrt{\eta^\top \E\left[\tilde{H}_t{\tilde{H}_t}^\top\right] \eta}.$$

\section{Input medications}\label{sec:drug}

    When drugs are used as input, the entries of the matrix $P_{\theta^P}$ are given 
    $$[P_{\theta^P}]_{ij}\propto \exp (\alpha_{ij}^{\text{drug}} + f_l(X_t,i) \indicator{i>j} + f_r(X_t,i)  \indicator{j>i}),$$
    where $f_l(X_t,i)=W_{l,i}^{\text{drug}} \cdot X_t\geq 0$ and $f_r(X_t,i)=W_{r,i}^{\text{drug}}  \cdot X_t$. This parametrization captures the fact that drugs can either add probability mass on lower state values ($W_{l,i}^{\text{drug}}>0$) or add probability mass on higher state values ($W_{r,i}^{\text{drug}}>0$). It is important to note that adding probability mass to lower state values will automatically reduce the probability mass on higher state values, and vice versa. In addition, $W_{l,i}^{\text{drug}},W_{r,i}^{\text{drug}}$, follow a priori a Gaussian distribution centered at 0, which is equivalent to add the penalizations $\|W_{l,i}^{\text{drug}}\|_2^2$ to the objective function. The underlying assumptions is that the drug effect is a priori small (i.e., $W_{l,i}^{\text{drug}}$ is likely to be small).

\begin{figure}[ht]
    \centering
    \begin{subfigure}[b]{0.7\textwidth}
        \includegraphics[width=1\textwidth]{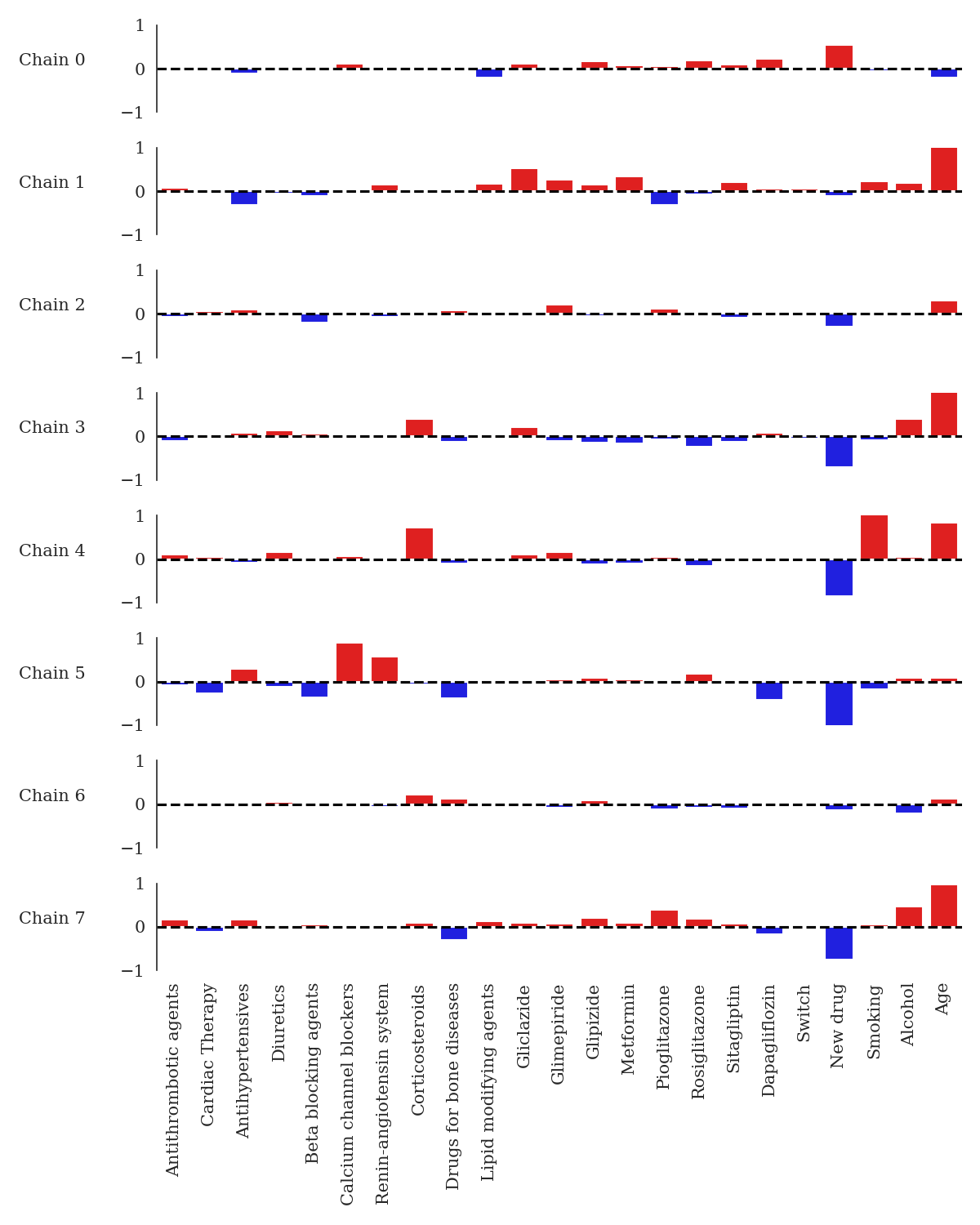}
    \end{subfigure}
    \begin{subfigure}[b]{0.8\textwidth}
        \includegraphics[width=1\textwidth]{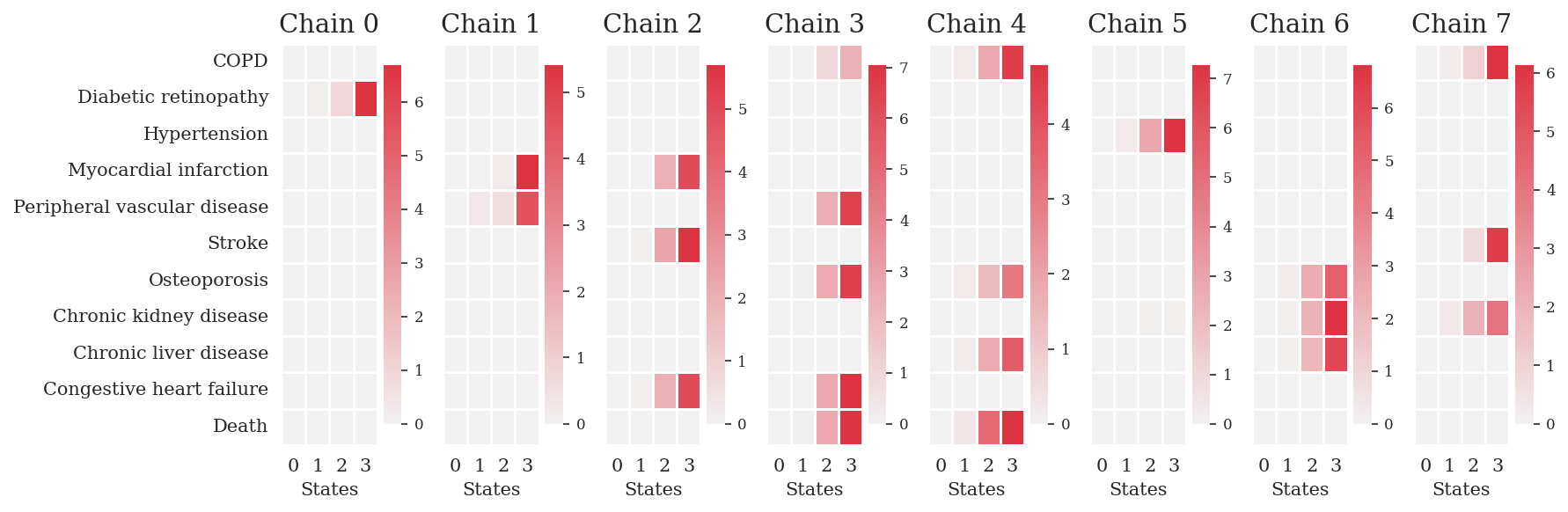}
    \end{subfigure}
    \caption{\textit{Top:} Average effect of drugs on the transition matrix, estimated as $1/K\sum_{i=1}^K \left(W_{r,i}^{\text{drug}} -W_{l,i}^{\text{drug}}\right)$. Red (resp. blue) means an increase of the probability to transition to a higher (resp. lower) state value. \textit{Bottom:} Emission parameters $\eta$ modeling the risk of developing a comorbidity for a given state. Larger values indicate a higher probability of having the comorbidity. The model includes 8 chains with 4 states per chain and is fitted on \emph{Men $<65$}.}
    \label{fig:input_param}
\end{figure}

Table \ref{tab:drugs} shows the performance of the model when drugs are used as input. The use of drugs in the model, is marginal and it does not improve the BIC score. The AIC is improved in few cases. 
Figure \ref{fig:input_param} show how the input variables affect the evolution of the latent state for the model $\text{FHMM}(8,4)_2$ fitted on the male population younger than 65. Age generally has a negative effect on chain transitions, increasing the probability of moving to a worsening state. Smoking also accelerates progression along chain 4, which is associated with a higher risk of COPD. The use of calcium channel blockers is linked to chain 5, which models hypertension. Here we observe an apparent worsening effect, although this may simply reflect that the model associates higher states with greater use of calcium channel blockers. By contrast, the addition of a new antidiabetic drug typically produces a positive effect on the patient’s state. 

\begin{table}[ht]
    \centering
    \begin{tabular}{cc|cc|cc}
    &Metric & \multicolumn{2}{c}{BIC}&\multicolumn{2}{c}{AIC} \\
     Category&Model & Drugs& No Drugs&Drugs& No Drugs \\
    \midrule
     \multirow{5}{*}{Men $<65$}&  $\text{FHMM}(3,3)_1$ &315914 & \textbf{310663} & {308086} &  \textbf{307445} \\
      &  $\text{FHMM}(4,4)_2$ & 308458& \textbf{292039}&294662 &  \textbf{286438} \\
       &  $\text{FHMM}(6,4)_2$  & {291660} &  \textbf{264231}& {271273} &   \textbf{256136}\\
        &  $\text{FHMM}(8,4)_2$ & 280861& \textbf{273337} &\textbf{253882} &   {262748} \\
         &  $\text{FHMM}(10,2)_1$ & 286149&  \textbf{276474}& \textbf{270171} &270739   \\
      \midrule
 \multirow{5}{*}{Women $<65$}&  $\text{FHMM}(3,3)_1$  &231096 & \textbf{225526} &223455 &  \textbf{222385} \\
      &  $\text{FHMM}(4,4)_2$  &231223 & \textbf{216236} & {217757} & \textbf{210769}  \\
       & $\text{FHMM}(6,4)_2$ & 212362& \textbf{190847} & {192462} &\textbf{182946}  \\
        &  $\text{FHMM}(8,4)_2$  & {203653} &\textbf{178959} & {177320} &  \textbf{168623} \\
         & $\text{FHMM}(10,2)_1$ & 218704&  \textbf{211039}&\textbf{203108} & {205441}  \\
      \midrule
       \multirow{5}{*}{Men $65+$}&  $\text{FHMM}(3,3)_1$  & {229811} & \textbf{229629}& \textbf{222198} &  226500\\
      &  $\text{FHMM}(4,4)_2$  &  {228468}& \textbf{221682}& \textbf{215051} & 216235 \\
       & $\text{FHMM}(6,4)_2$  &218800 & \textbf{201848} &  {198973}& \textbf{193976} \\
        &  $\text{FHMM}(8,4)_2$   &203550 & \textbf{181179} &177312 &  \textbf{170881} \\
         & $\text{FHMM}(10,2)_1$ & 204037&  \textbf{192274}&188498 & \textbf{186698}   \\
      \midrule
       \multirow{5}{*}{Women $65+$}&  $\text{FHMM}(3,3)_1$  & {208526} &\textbf{203633} & {201026} &  \textbf{200550}\\
      &  $\text{FHMM}(4,4)_2$ &204033 & \textbf{195206} & 190813&    \textbf{189839}\\
       & $\text{FHMM}(6,4)_2$  & 195974&  \textbf{181320}&  {176439}&\textbf{173563}  \\
        &  $\text{FHMM}(8,4)_2$ &180633& \textbf{162739} &154781 &  \textbf{152592} \\
         & $\text{FHMM}(10,2)_1$  &184247 & \textbf{174079} &168937 & \textbf{168585} \\
     \bottomrule
    \end{tabular}
    \caption{Comparison between models with drugs ($X_t$) and without drugs. Metrics are evaluated on the validation set.} 
    \label{tab:drugs}
\end{table}

\section{Additional Figures}

The number of clusters is determined using the Davies-Bouldin index \cite{Davies1979Cluster} and the prediction strength introduced by \cite{Tib2005Cluster}. The prediction strength is estimated using a 3-fold cross-validation setup, where two-thirds of the data is used for training and one-third for testing. 

\begin{figure}[ht]
    \centering
    \includegraphics[width=0.8\textwidth]{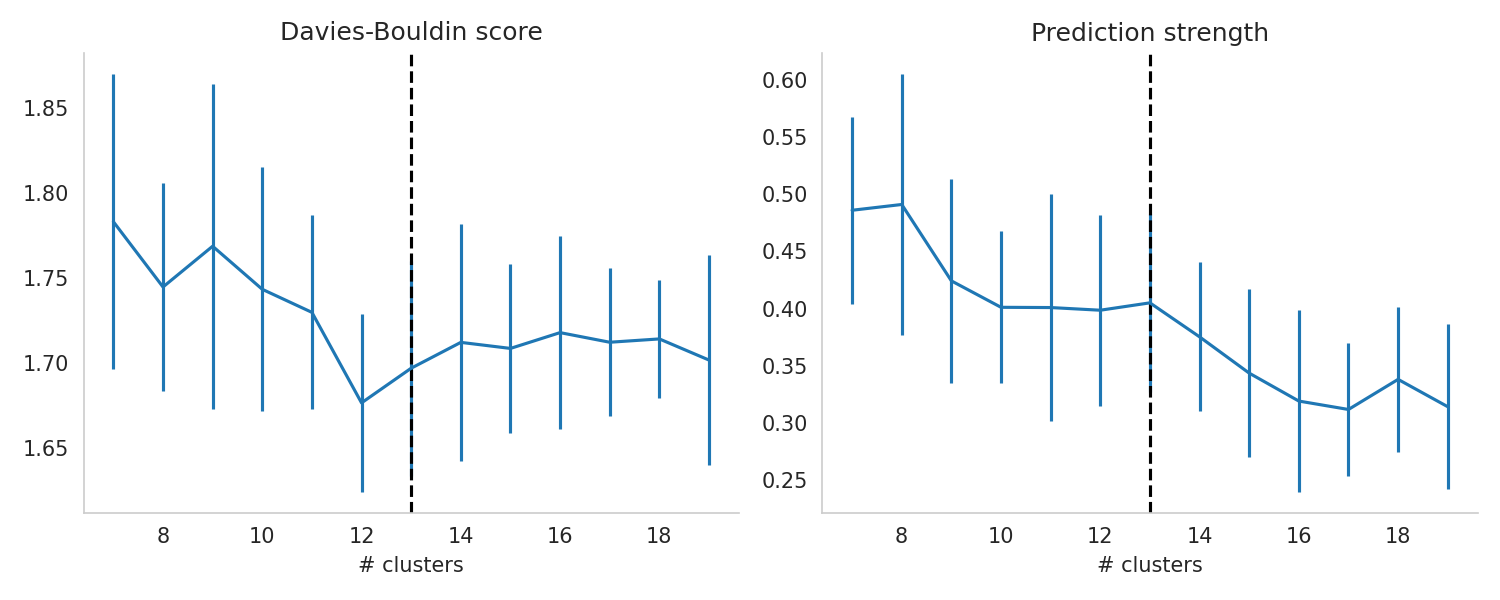}
    \caption{Cluster selection using Davies-Bouldin index and the prediction strength. The optimal number of cluster is indicated by the black dashed line. This is fitted on the population of \textit{Women $<65$}.}
    \label{fig:cluster_selection_women<65}
\end{figure}

\begin{figure}[ht!]
    \centering
   \begin{subfigure}[b]{0.32\textwidth}
        \centering
         \includegraphics[width=1\textwidth]{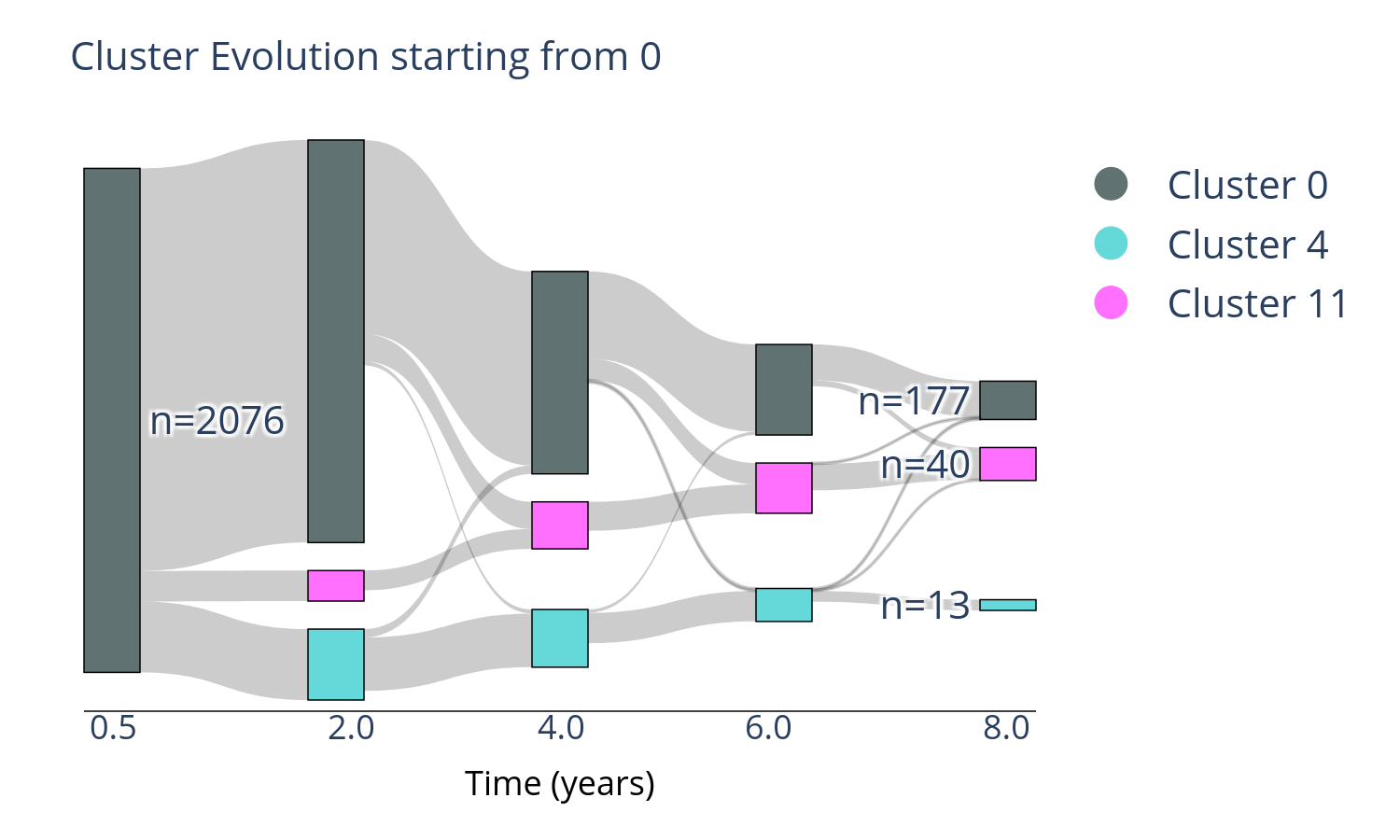}
    \end{subfigure}
         \begin{subfigure}[b]{0.32\textwidth}
        \centering
         \includegraphics[width=1\textwidth]{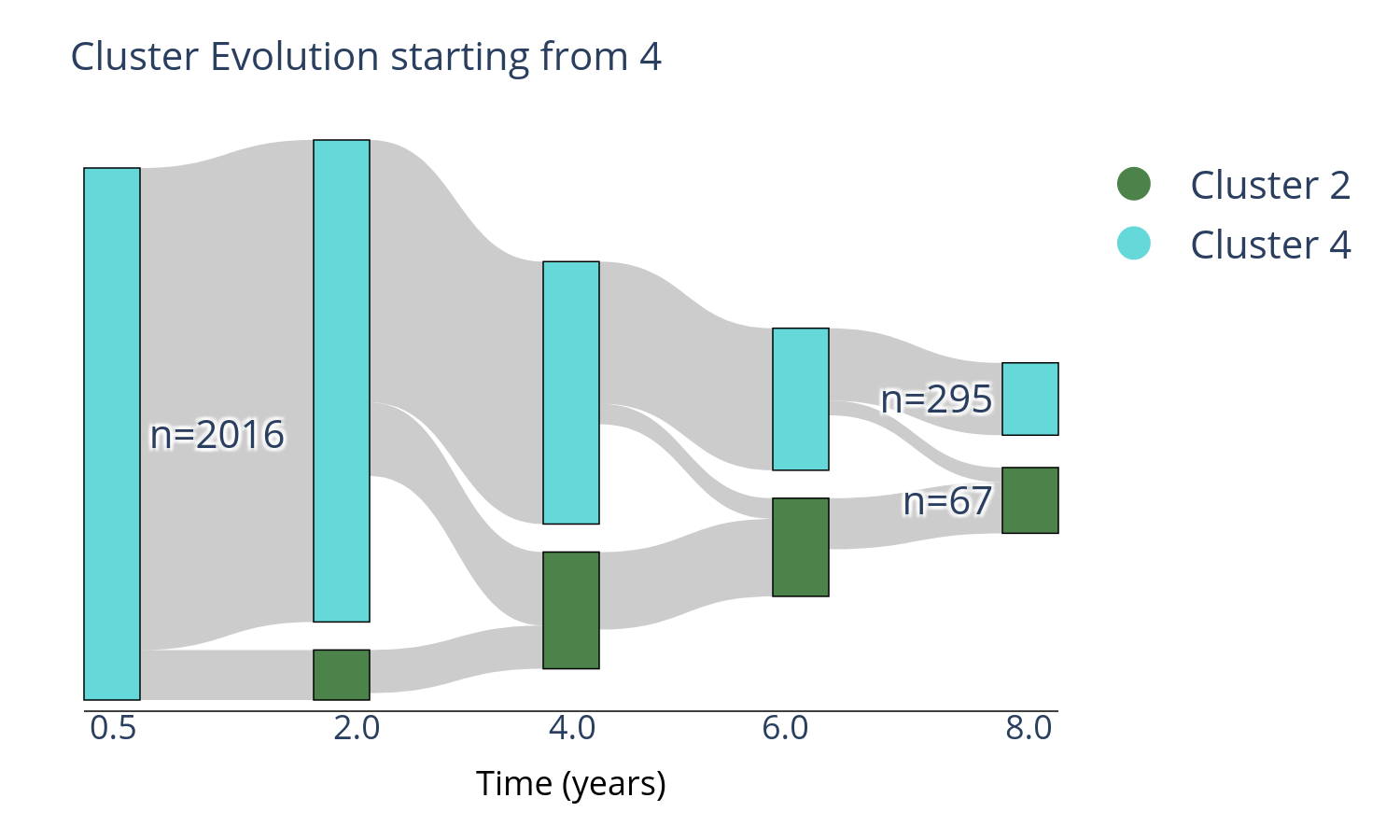}
    \end{subfigure} 
    \begin{subfigure}[b]{0.33\textwidth}
        \centering
         \includegraphics[width=1\textwidth]{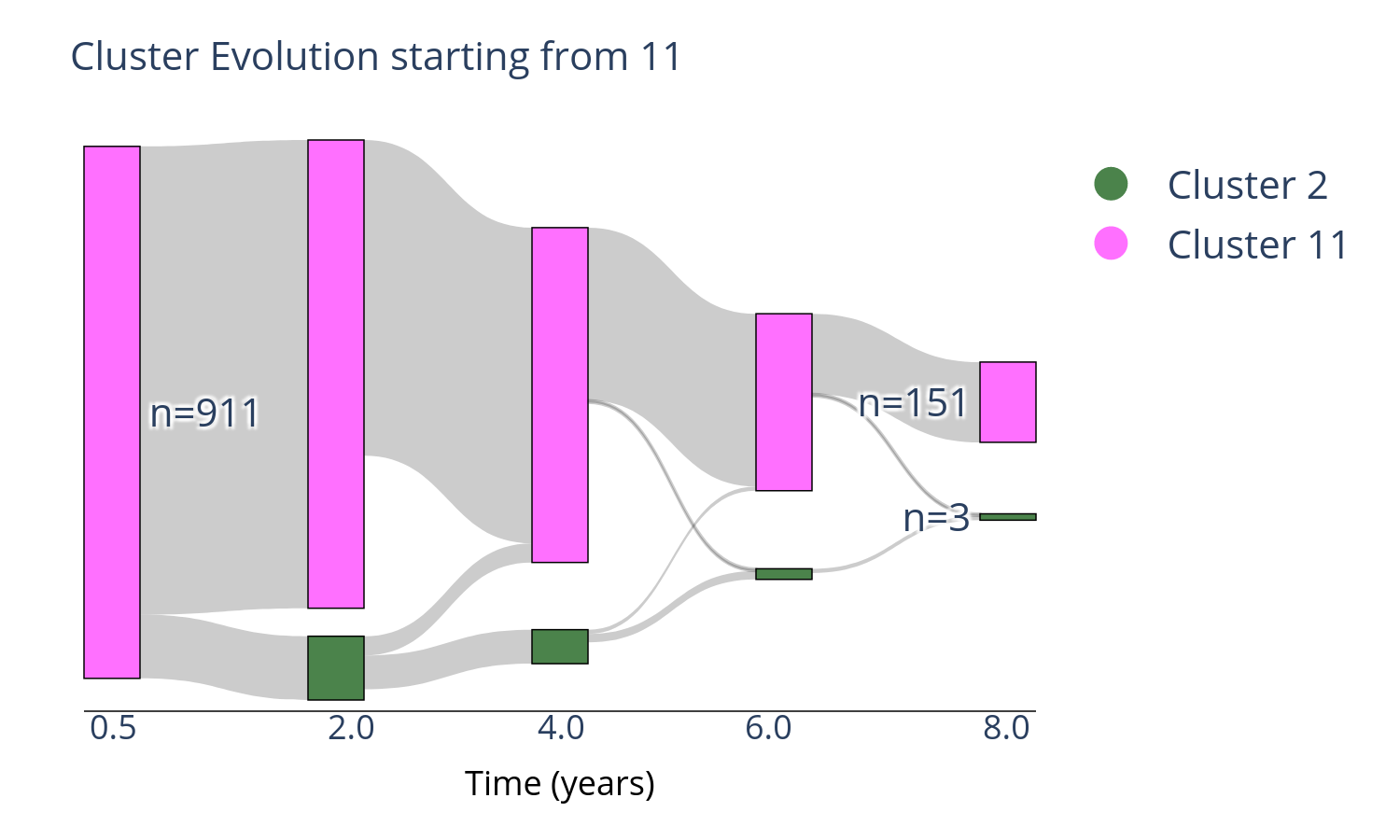}
    \end{subfigure}
    
    \begin{subfigure}[b]{0.32\textwidth}
        \centering
         \includegraphics[width=1\textwidth]{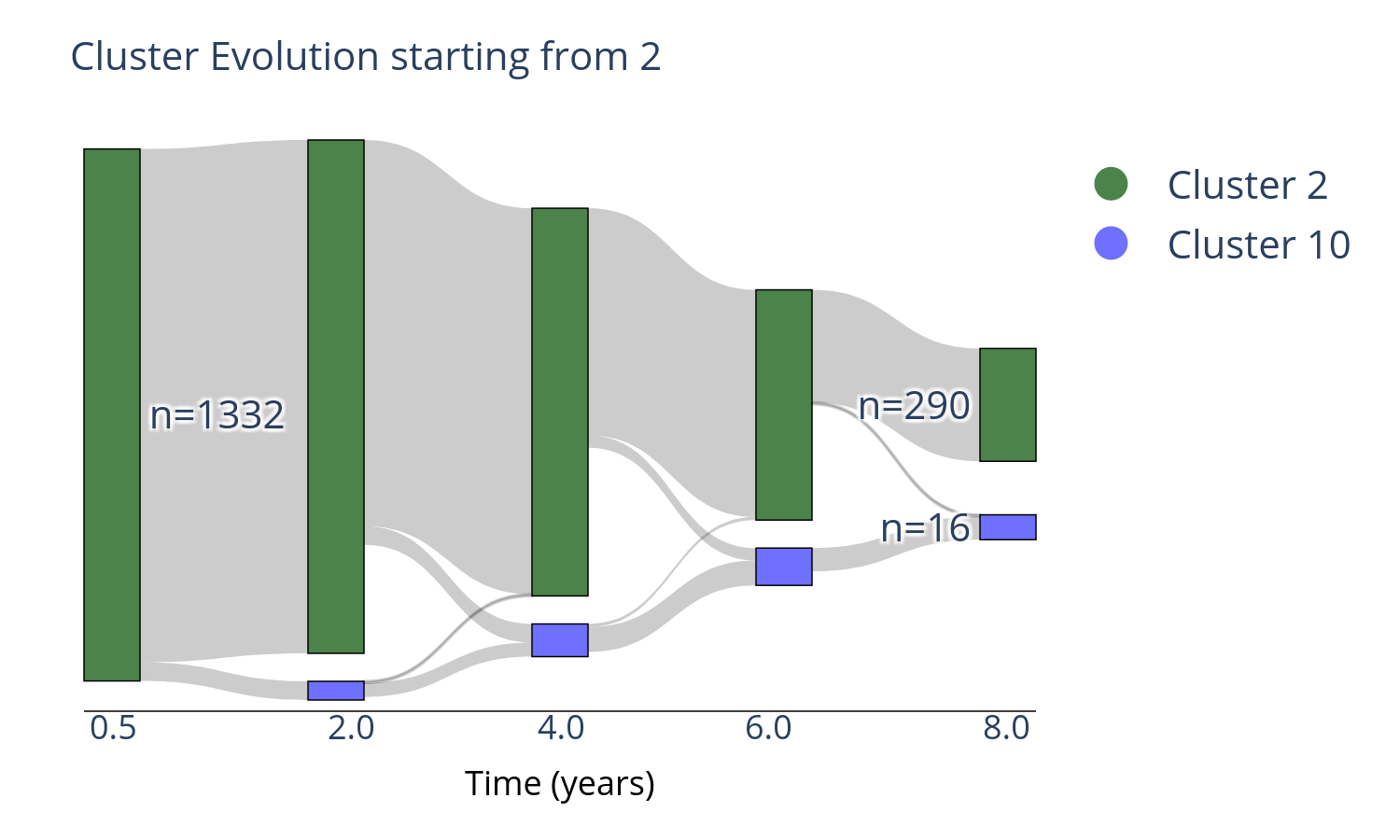}
    \end{subfigure}
     \begin{subfigure}[b]{0.33\textwidth}
        \centering
         \includegraphics[width=1\textwidth]{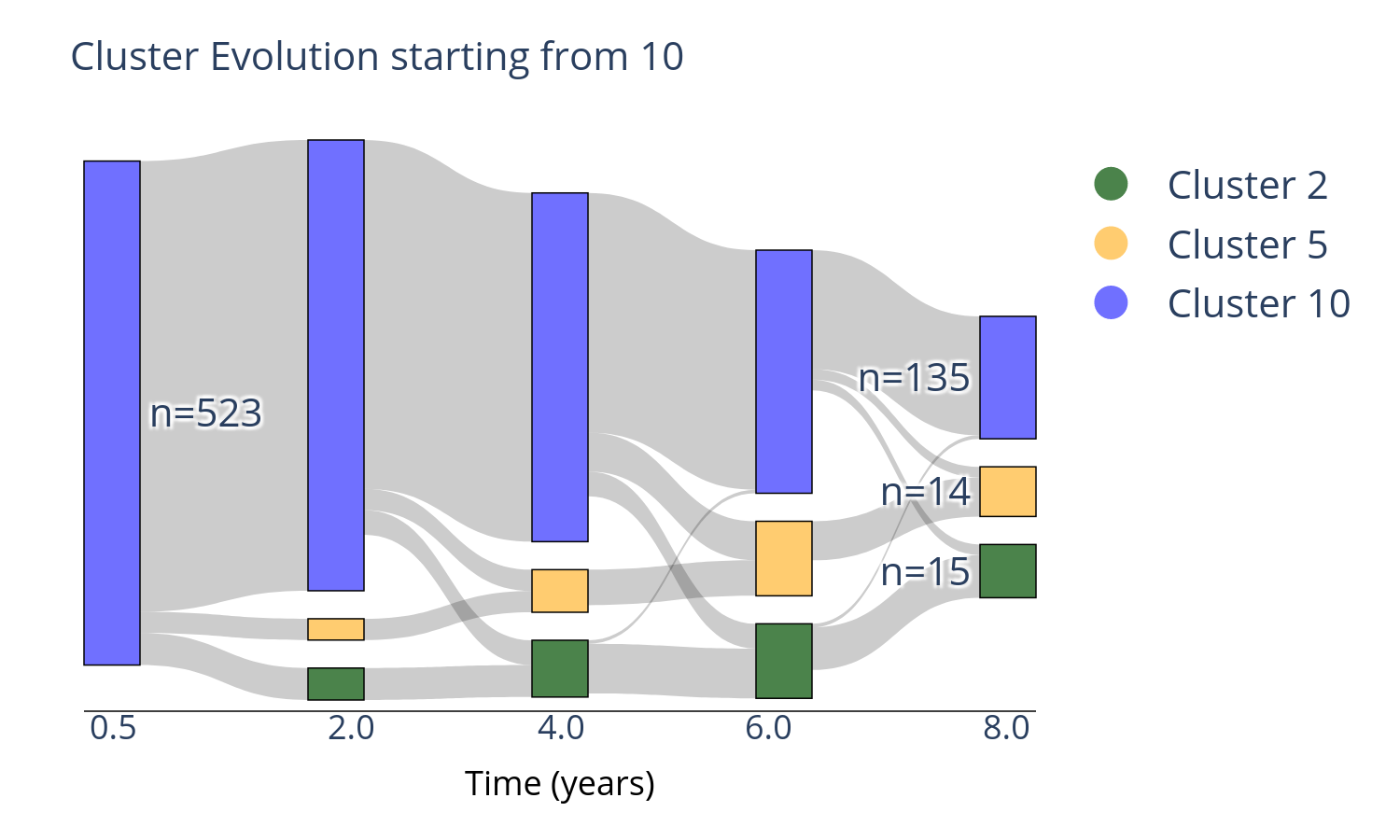}
    \end{subfigure}
    \centering 
     \begin{subfigure}[b]{0.31\textwidth}
        \centering
         \includegraphics[width=1\textwidth]{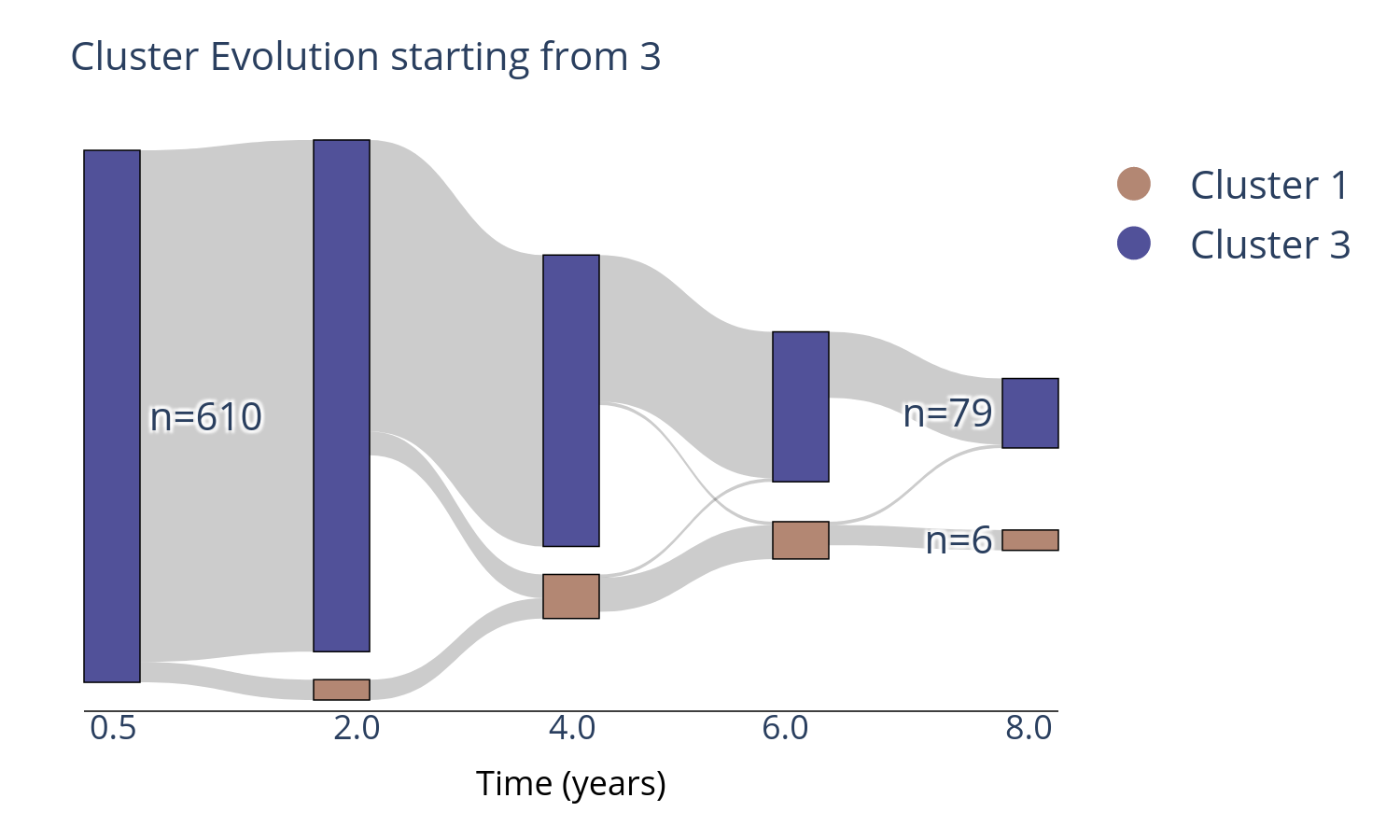}
    \end{subfigure} 
     \begin{subfigure}[b]{0.31\textwidth}
        \centering
         \includegraphics[width=1\textwidth]{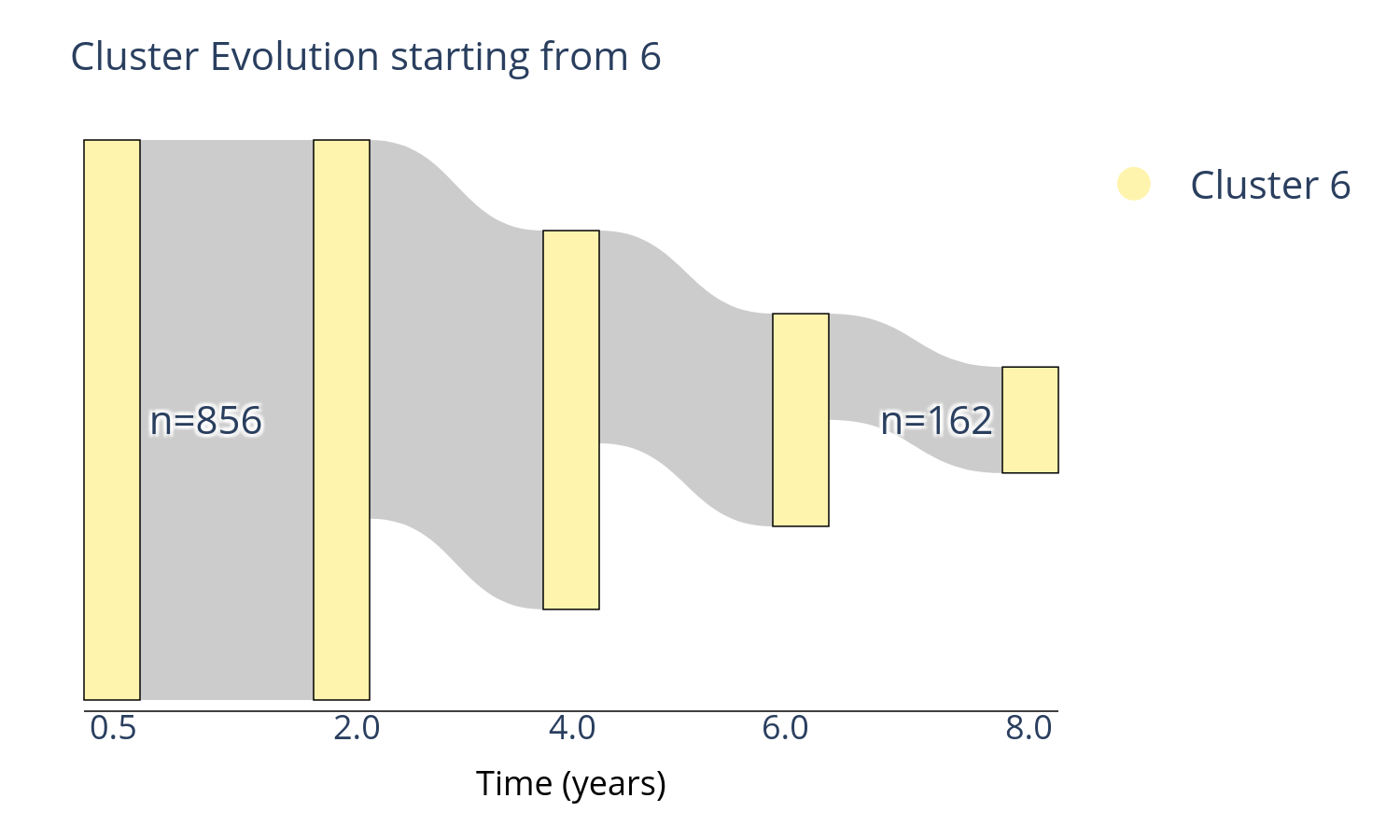}
    \end{subfigure}
     \begin{subfigure}[b]{0.31\textwidth}
        \centering
         \includegraphics[width=1\textwidth]{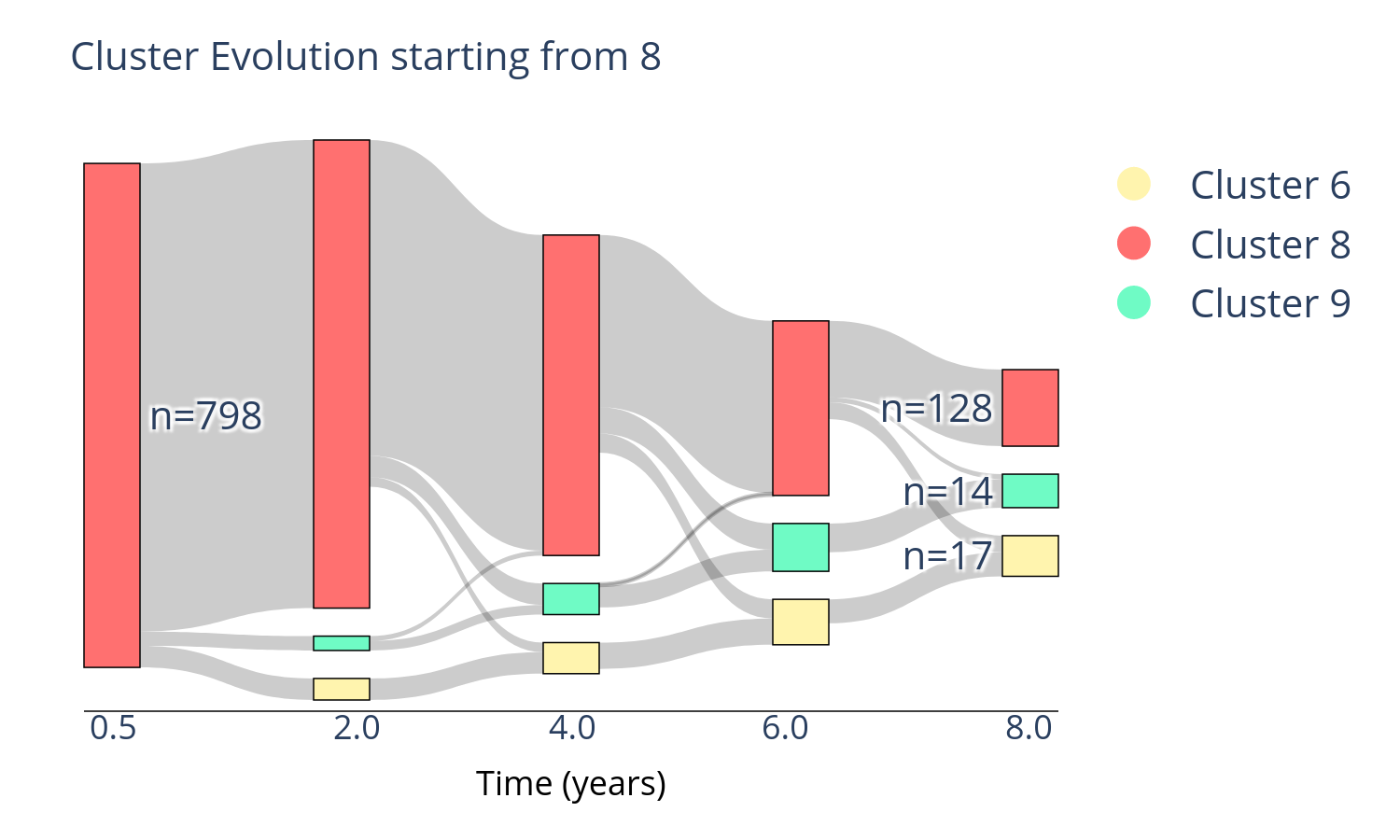}
    \end{subfigure}
     \begin{subfigure}[b]{0.31\textwidth}
        \centering
         \includegraphics[width=1\textwidth]{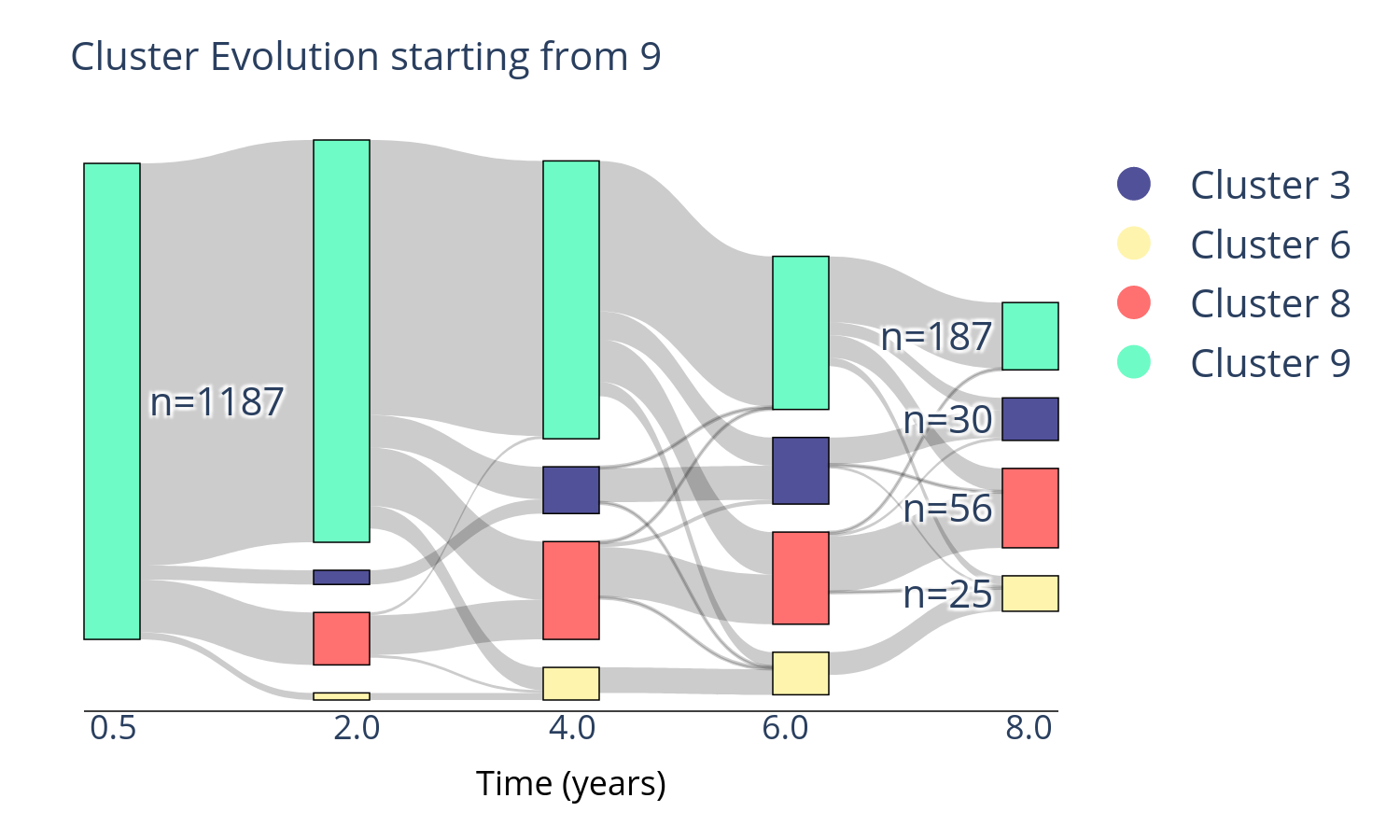}
    \end{subfigure} 
     \begin{subfigure}[b]{0.31\textwidth}
        \centering
         \includegraphics[width=1\textwidth]{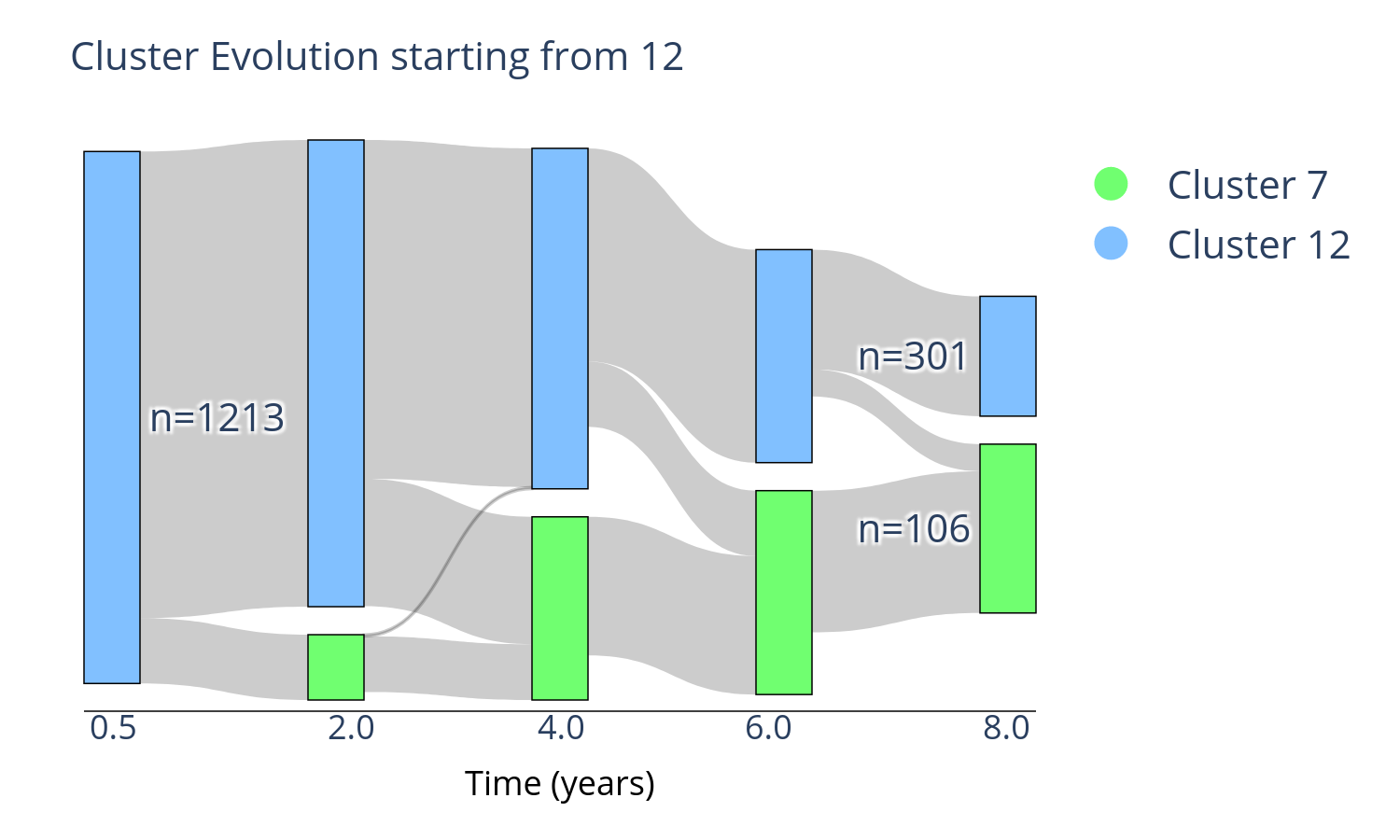}
    \end{subfigure}
        \caption{
Evolution of patient clusters 0 (healthy), 2 (pathway to multimorbidity), 3 (pulmonary-metabolic), 4 (isolated hypertension), 6 (hepatic), 8 (advanced nephropathy), 9 (obese hypertension), 10 (bone-vascular), 11 (isolated DR), and 12 (cardio-pulmonary multimorbidity) over time (\textit{Women $<65$}), conditional on the initial cluster assignment (6 months post-NIAD).
The number of patients in each cluster is indicated alongside the corresponding bars.
For clarity, flows representing less than 1\% of the initial cluster size, as well as transitions to no available follow-up data, are not displayed.
Clusters 1, 5, and 7 are excluded due to the low number of transitions observed from these clusters.
}
    \label{fig:female_m65_evolution}
\end{figure}

\clearpage

\section{Male $<65$}
\begin{table}[ht!]
    \centering
    \resizebox{\textwidth}{!}{
\begin{tabular}{l|ccccccccccccccc}
    \toprule
Cluster & 0 & 1 & 2 & 3 & 4 & 5 & 6 & 7 & 8 & 9 & 10 & 11 & 12 & 13 & 14\\
    \midrule
\# of patients & 6504 & 1788 & 1698 & 1166 & 3911 & 1522 & 360 & 1212 & 957 & 1534 & 1010 & 1511 & 1070 & 703 & 685\\
Avg. age at NIAD (y) $\pm$ SD & 50.88 $\pm$ 8.68 & 53.66 $\pm$ 8.11 & 55.28 $\pm$ 6.95 & 57.41 $\pm$ 6.39 & 53.89 $\pm$ 7.32 & 56.18 $\pm$ 6.73 & 58.28 $\pm$ 5.63 & 53.21 $\pm$ 7.78 & 57.69 $\pm$ 5.92 & 57.42 $\pm$ 6.14 & 55.41 $\pm$ 7.21 & 51.53 $\pm$ 8.35 & 56.68 $\pm$ 6.42 & 55.89 $\pm$ 7.02 & 54.0 $\pm$ 7.97\\
Avg. follow-up post-NIAD (y) $\pm$ SD & 4.28 $\pm$ 3.1 & 6.85 $\pm$ 3.25 & 5.78 $\pm$ 2.86 & 5.38 $\pm$ 3.11 & 4.72 $\pm$ 3.22 & 6.52 $\pm$ 3.27 & 4.49 $\pm$ 2.89 & 5.97 $\pm$ 3.06 & 5.46 $\pm$ 3.59 & 5.58 $\pm$ 3.57 & 3.63 $\pm$ 2.85 & 5.29 $\pm$ 2.75 & 6.25 $\pm$ 2.88 & 5.05 $\pm$ 3.08 & 6.29 $\pm$ 2.64\\
Avg. BMI at NIAD (kg/m\textsuperscript{2}) $\pm$ SD & 30.85 $\pm$ 5.52 & 34.96 $\pm$ 6.67 & 31.58 $\pm$ 4.79 & 34.31 $\pm$ 6.68 & 32.47 $\pm$ 5.32 & 30.86 $\pm$ 4.48 & 32.25 $\pm$ 6.67 & 43.79 $\pm$ 5.7 & 32.82 $\pm$ 4.93 & 33.54 $\pm$ 4.23 & 33.08 $\pm$ 6.39 & 30.38 $\pm$ 4.95 & 36.46 $\pm$ 6.85 & 32.01 $\pm$ 6.18 & 34.96 $\pm$ 6.77\\
Avg. HbA1c at NIAD (\%) $\pm$ SD & 7.45 $\pm$ 1.73 & 7.39 $\pm$ 1.61 & 7.47 $\pm$ 1.59 & 7.31 $\pm$ 1.64 & 7.23 $\pm$ 1.56 & 7.19 $\pm$ 1.6 & 7.31 $\pm$ 1.61 & 7.44 $\pm$ 1.63 & 7.36 $\pm$ 1.7 & 7.2 $\pm$ 1.45 & 7.4 $\pm$ 1.71 & 7.69 $\pm$ 1.7 & 7.48 $\pm$ 1.59 & 7.47 $\pm$ 1.63 & 7.59 $\pm$ 1.65\\
\% Ever smoked & 43.48 & 53.75 & 32.63 & 62.86 & 39.43 & 34.82 & 71.67 & 39.69 & 53.81 & 44.72 & 55.15 & 39.38 & 44.11 & 70.13 & 53.87\\
\% Ever alcohol & 84.62 & 88.59 & 92.34 & 92.02 & 91.2 & 89.49 & 93.06 & 90.68 & 89.97 & 92.57 & 90.79 & 89.61 & 93.74 & 87.77 & 91.82\\
    \midrule
\multicolumn{16}{l}{\textit{Comorbidities at NIAD [N,(\%)]}}\\
Stroke & 86 (1.32) & 46 (2.57) & 62 (3.65) & 82 (7.03) & $<1$ \% & 34 (2.23) & 37 (10.28) & $<1$ \% & 32 (3.34) & 258 (16.82) & 55 (5.45) & 18 (1.19) & 57 (5.33) & 25 (3.56) & 30 (4.38)\\
Myocardial infarction & $<1$ \% & 370 (20.69) & $<1$ \% & 99 (8.49) & $<1$ \% & $<1$ \% & 63 (17.50) & 76 (6.27) & 517 (54.02) & 30 (1.96) & 61 (6.04) & $<1$ \% & 201 (18.79) & 33 (4.69) & 162 (23.65)\\
Congestive heart failure & $<1$ \% & $<1$ \% & $<1$ \% & 95 (8.15) & $<1$ \% & $<1$ \% & 35 (9.72) & 46 (3.80) & 231 (24.14) & $<1$ \% & 24 (2.38) & $<1$ \% & 103 (9.63) & $<1$ \% & $<1$ \%\\
Peripheral vascular disease & $<1$ \% & 54 (3.02) & $<1$ \% & $<1$ \% & $<1$ \% & $<1$ \% & 108 (30.00) & $<1$ \% & $<1$ \% & $<1$ \% & 230 (22.77) & $<1$ \% & 15 (1.40) & 7 (1.00) & 21 (3.07)\\
Diabetic retinopathy & 159 (2.44) & $<1$ \% & 931 (54.83) & 188 (16.12) & 97 (2.48) & $<1$ \% & 61 (16.94) & $<1$ \% & 22 (2.30) & 50 (3.26) & 144 (14.26) & 714 (47.25) & 556 (51.96) & 87 (12.38) & 352 (51.39)\\
Hypertension & 310 (4.77) & 19 (1.06) & 1573 (92.64) & 1012 (86.79) & 3398 (86.88) & 1428 (93.82) & 207 (57.50) & 1080 (89.11) & 721 (75.34) & 1363 (88.85) & 700 (69.31) & 29 (1.92) & 945 (88.32) & 27 (3.84) & $<1$ \%\\
Chronic kidney disease & $<1$ \% & $<1$ \% & $<1$ \% & 434 (37.22) & $<1$ \% & $<1$ \% & 57 (15.83) & $<1$ \% & $<1$ \% & $<1$ \% & $<1$ \% & $<1$ \% & $<1$ \% & 198 (28.17) & $<1$ \%\\
Osteoporosis & $<1$ \% & $<1$ \% & $<1$ \% & $<1$ \% & $<1$ \% & $<1$ \% & 96 (26.67) & $<1$ \% & $<1$ \% & $<1$ \% & $<1$ \% & $<1$ \% & $<1$ \% & $<1$ \% & $<1$ \%\\
COPD & $<1$ \% & $<1$ \% & $<1$ \% & 500 (42.88) & $<1$ \% & $<1$ \% & 158 (43.89) & $<1$ \% & $<1$ \% & $<1$ \% & $<1$ \% & $<1$ \% & $<1$ \% & 343 (48.79) & 7 (1.02)\\
Chronic liver disease & 81 (1.25) & 38 (2.13) & $<1$ \% & $<1$ \% & $<1$ \% & $<1$ \% & 75 (20.83) & $<1$ \% & $<1$ \% & $<1$ \% & 411 (40.69) & 36 (2.38) & $<1$ \% & 15 (2.13) & 10 (1.46)\\
    \midrule
\multicolumn{16}{l}{\textit{Comorbidities during follow-up [N,(\%)]}}\\
Stroke & $<1$ \% & 47 (2.63) & 34 (2.00) & 43 (3.69) & 42 (1.07) & 35 (2.30) & 10 (2.78) & $<1$ \% & 39 (4.08) & 84 (5.48) & 19 (1.88) & 18 (1.19) & 32 (2.99) & 14 (1.99) & 22 (3.21)\\
Myocardial infarction & $<1$ \% & 107 (5.98) & $<1$ \% & 25 (2.14) & $<1$ \% & $<1$ \% & 8 (2.22) & 39 (3.22) & 58 (6.06) & 58 (3.78) & 25 (2.48) & $<1$ \% & 58 (5.42) & 11 (1.56) & 33 (4.82)\\
Congestive heart failure & $<1$ \% & 58 (3.24) & $<1$ \% & 65 (5.57) & $<1$ \% & $<1$ \% & 24 (6.67) & 42 (3.47) & 95 (9.93) & 31 (2.02) & 13 (1.29) & $<1$ \% & 79 (7.38) & 11 (1.56) & 16 (2.34)\\
Peripheral vascular disease & $<1$ \% & 47 (2.63) & 20 (1.18) & 32 (2.74) & $<1$ \% & 21 (1.38) & 27 (7.50) & $<1$ \% & 30 (3.13) & 22 (1.43) & 72 (7.13) & $<1$ \% & 21 (1.96) & 9 (1.28) & 19 (2.77)\\
Diabetic retinopathy & 700 (10.76) & 263 (14.71) & 767 (45.17) & 256 (21.96) & 457 (11.68) & 224 (14.72) & 72 (20.00) & 171 (14.11) & 116 (12.12) & 192 (12.52) & 169 (16.73) & 797 (52.75) & 514 (48.04) & 130 (18.49) & 333 (48.61)\\
Hypertension & 585 (8.99) & 202 (11.30) & 113 (6.65) & 75 (6.43) & 395 (10.10) & 90 (5.91) & 24 (6.67) & 88 (7.26) & 38 (3.97) & 91 (5.93) & 75 (7.43) & 160 (10.59) & 54 (5.05) & 58 (8.25) & 54 (7.88)\\
Chronic kidney disease & 66 (1.01) & 35 (1.96) & 32 (1.88) & 165 (14.15) & 44 (1.13) & 36 (2.37) & 40 (11.11) & 39 (3.22) & 31 (3.24) & 27 (1.76) & 13 (1.29) & 22 (1.46) & 47 (4.39) & 102 (14.51) & 21 (3.07)\\
Osteoporosis & $<1$ \% & $<1$ \% & $<1$ \% & $<1$ \% & $<1$ \% & $<1$ \% & 20 (5.56) & $<1$ \% & $<1$ \% & $<1$ \% & $<1$ \% & $<1$ \% & $<1$ \% & $<1$ \% & $<1$ \%\\
COPD & $<1$ \% & 62 (3.47) & 21 (1.24) & 134 (11.49) & 50 (1.28) & 21 (1.38) & 36 (10.00) & 26 (2.15) & 33 (3.45) & 39 (2.54) & 24 (2.38) & $<1$ \% & 39 (3.64) & 85 (12.09) & 17 (2.48)\\
Chronic liver disease & 108 (1.66) & 45 (2.52) & 37 (2.18) & 23 (1.97) & 48 (1.23) & 24 (1.58) & 27 (7.50) & 27 (2.23) & 16 (1.67) & 20 (1.30) & 140 (13.86) & 43 (2.85) & 29 (2.71) & 18 (2.56) & 30 (4.38)\\
Deaths before 2020 & 97 (1.49) & 67 (3.75) & 30 (1.77) & 88 (7.55) & 78 (1.99) & 51 (3.35) & 85 (23.61) & 55 (4.54) & 68 (7.11) & 55 (3.59) & 238 (23.56) & 31 (2.05) & 60 (5.61) & 36 (5.12) & 28 (4.09)\\
    \bottomrule
\end{tabular}
    }
    \caption{Patient characteristics in each cluster for the \textit{Men $<65$} cohort with average values and standard deviation where applicable. Comorbidities are presented as absolute counts at NIAD initiation, along with the number of new comorbidities developed post-NIAD. Proportions, expressed as percentages of the cohort population, are reported in parentheses.}
    \label{tab:cluster_descr_male_<65}
\end{table}

\begin{figure}[ht]
    \centering
    \includegraphics[width=0.9\linewidth]{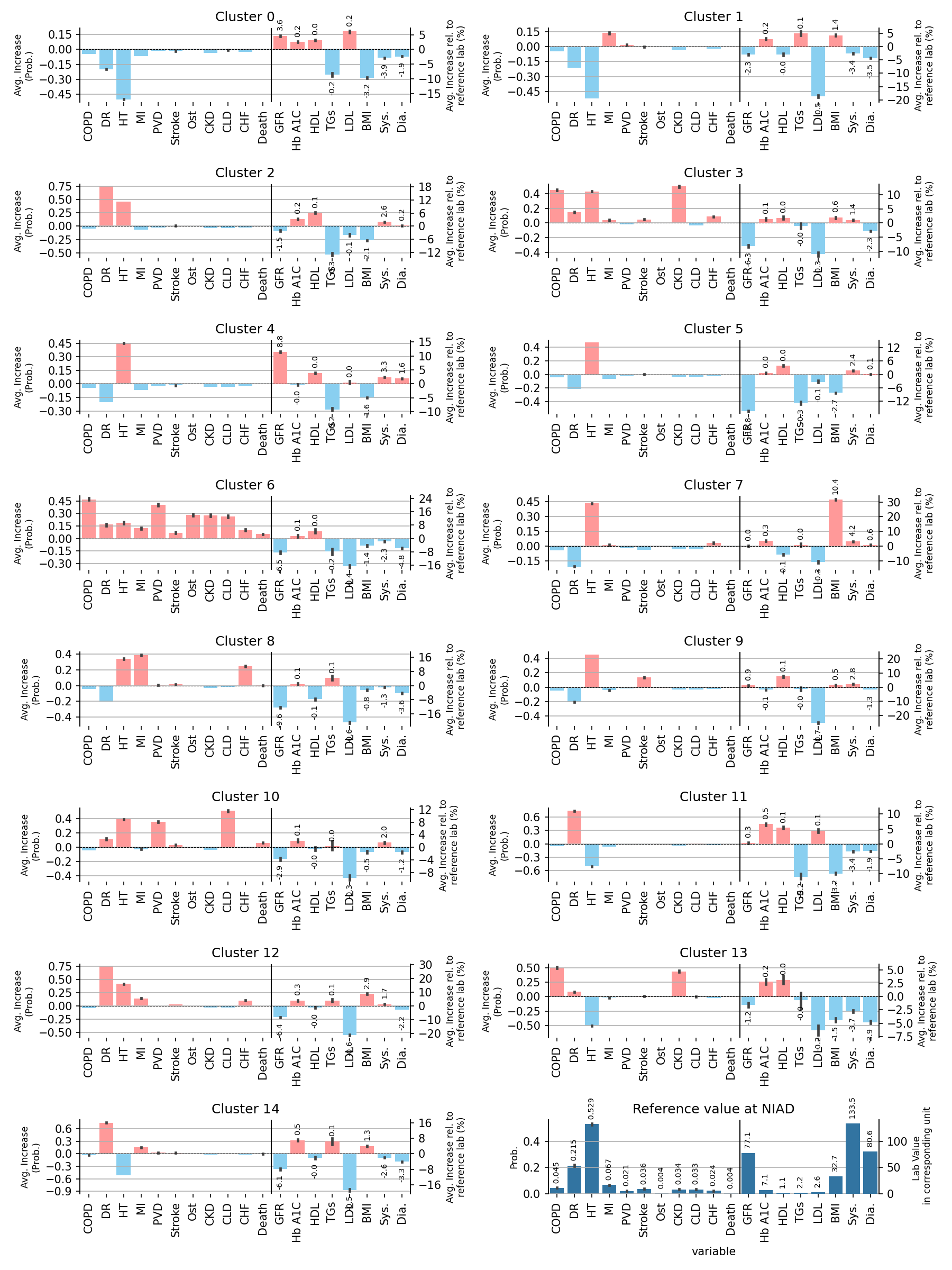}
    \caption{Average cluster-specific probabilities and lab values for \textit{Men $<65$}. Values have been centered by subtracting the average at the time of NIAD initiation to improve readability. Red bars indicate increases, and blue bars indicate decreases relative to the average patient at NIAD. Comorbidity risks are shown as changes in probability (left axis), and lab values as percentage changes relative to the reference value (right axis). Annotated numbers indicate absolute changes in the original units (see Table \ref{tab:list_of_var}). Error bars represent the standard error of the mean.}
    \label{fig:male_m65_proba_avg}
\end{figure}

\begin{figure}[ht]
    \centering
   \begin{subfigure}[b]{0.31\textwidth}
        \centering
         \includegraphics[width=1\textwidth]{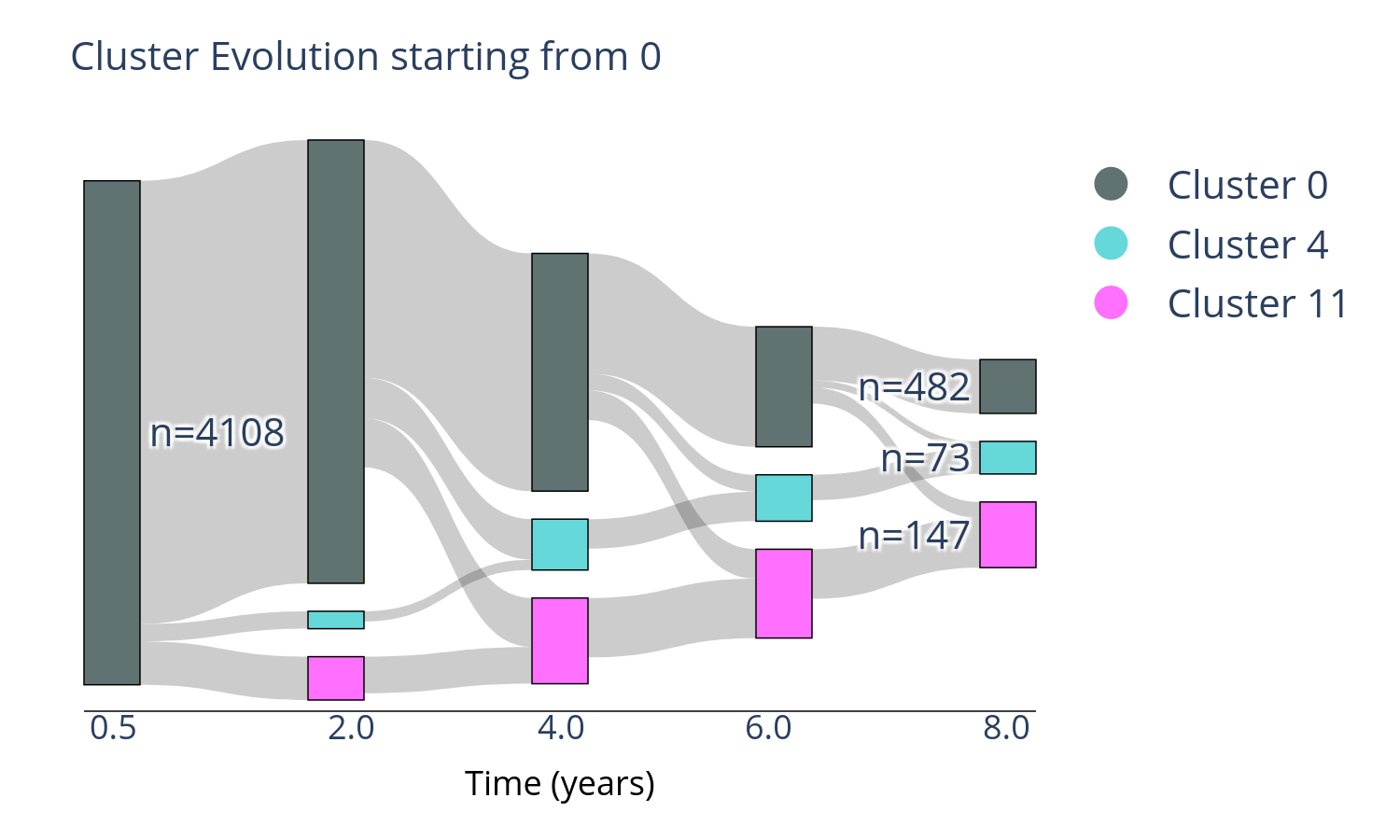}
    \end{subfigure}
    \begin{subfigure}[b]{0.31\textwidth}
        \centering
         \includegraphics[width=1\textwidth]{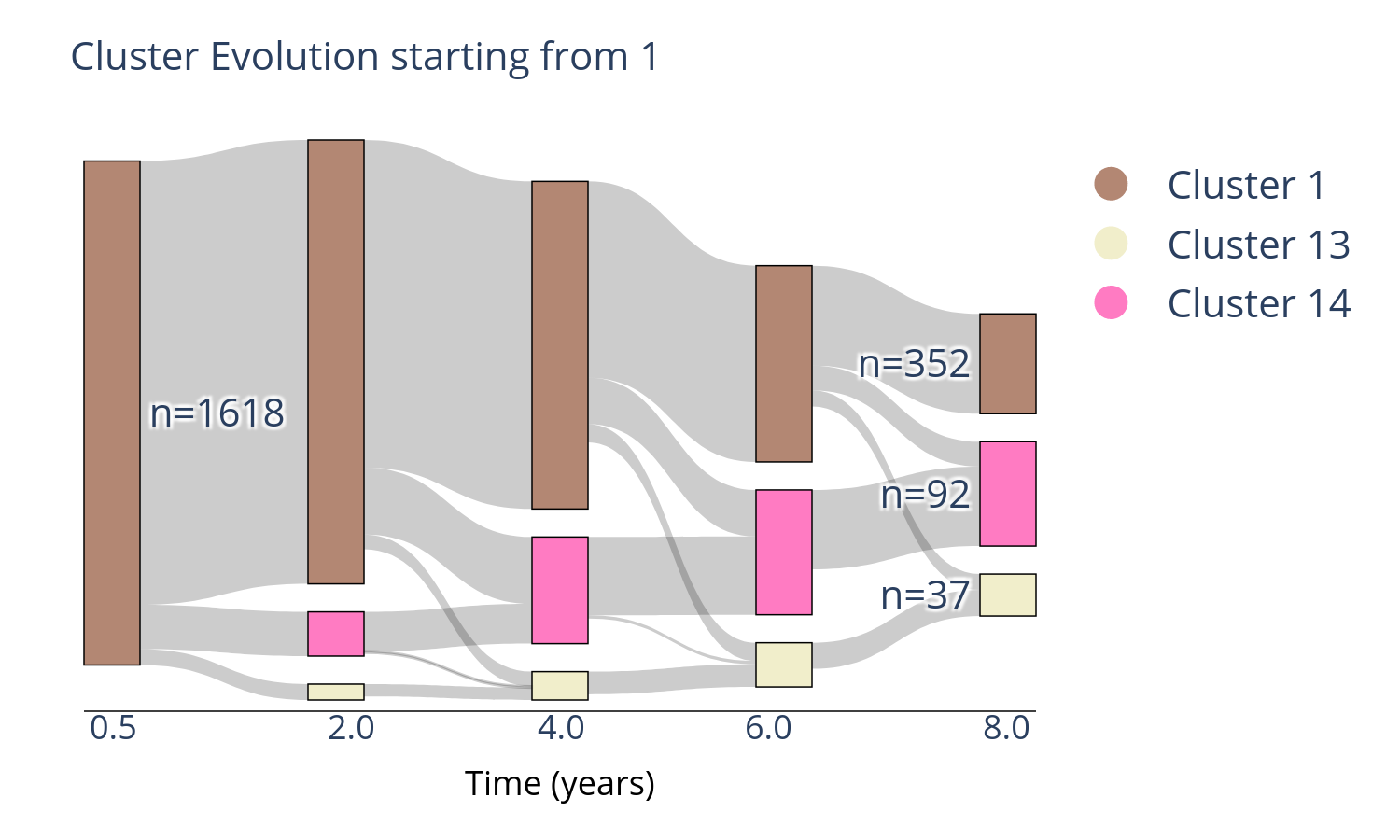}
    \end{subfigure}
     \begin{subfigure}[b]{0.31\textwidth}
        \centering
         \includegraphics[width=1\textwidth]{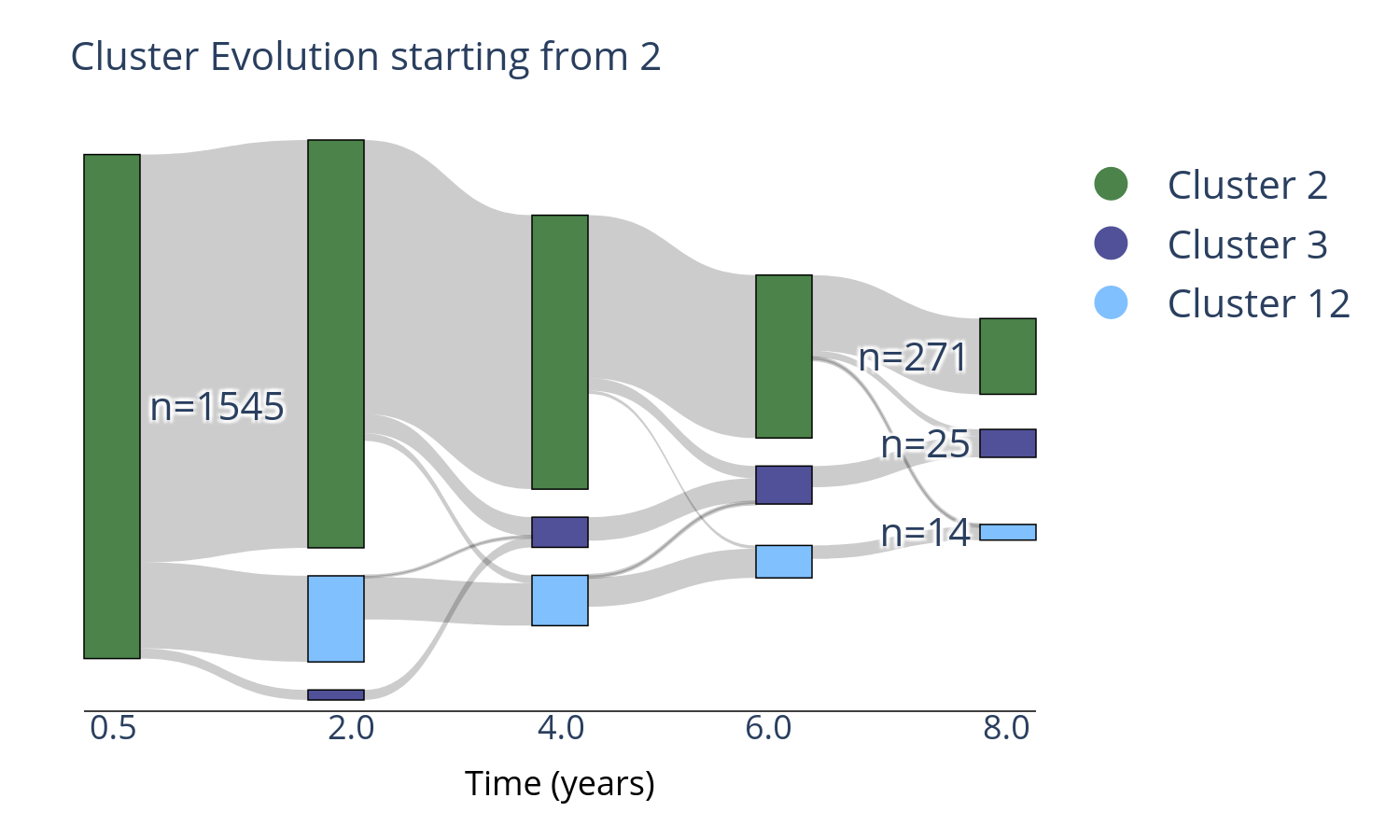}
    \end{subfigure}
     \begin{subfigure}[b]{0.31\textwidth}
        \centering
         \includegraphics[width=1\textwidth]{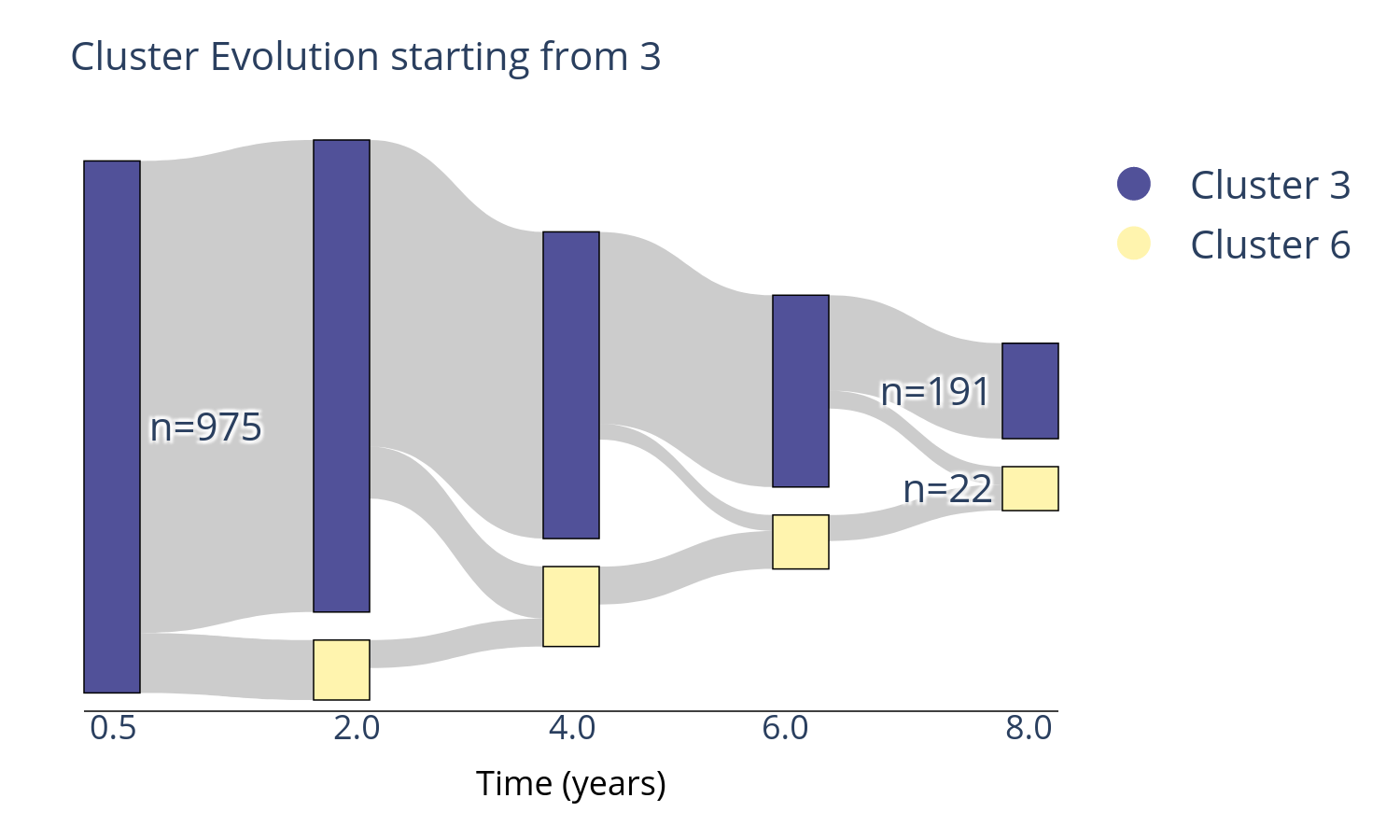}
    \end{subfigure}
     \begin{subfigure}[b]{0.31\textwidth}
        \centering
         \includegraphics[width=1\textwidth]{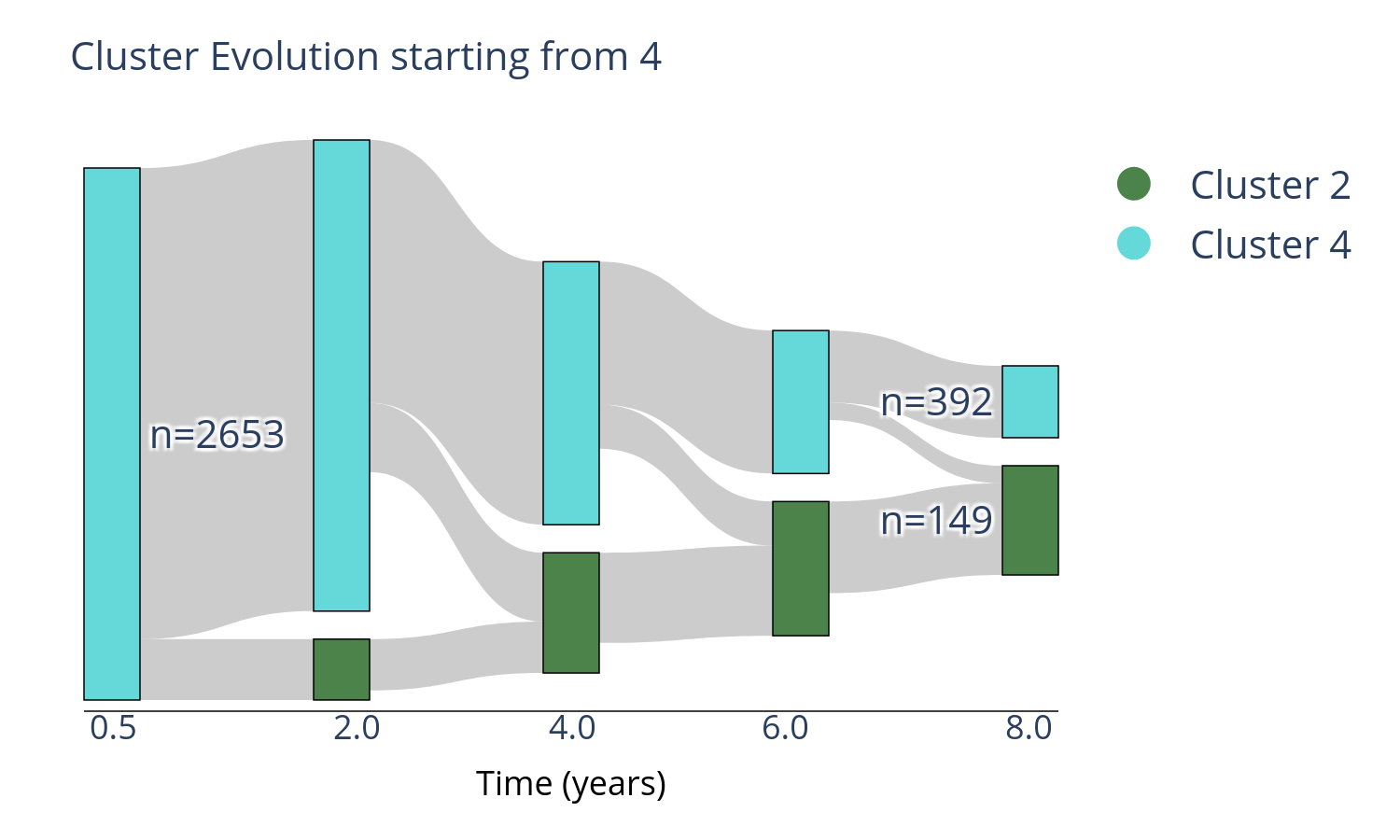}
    \end{subfigure}
     \begin{subfigure}[b]{0.31\textwidth}
        \centering
         \includegraphics[width=1\textwidth]{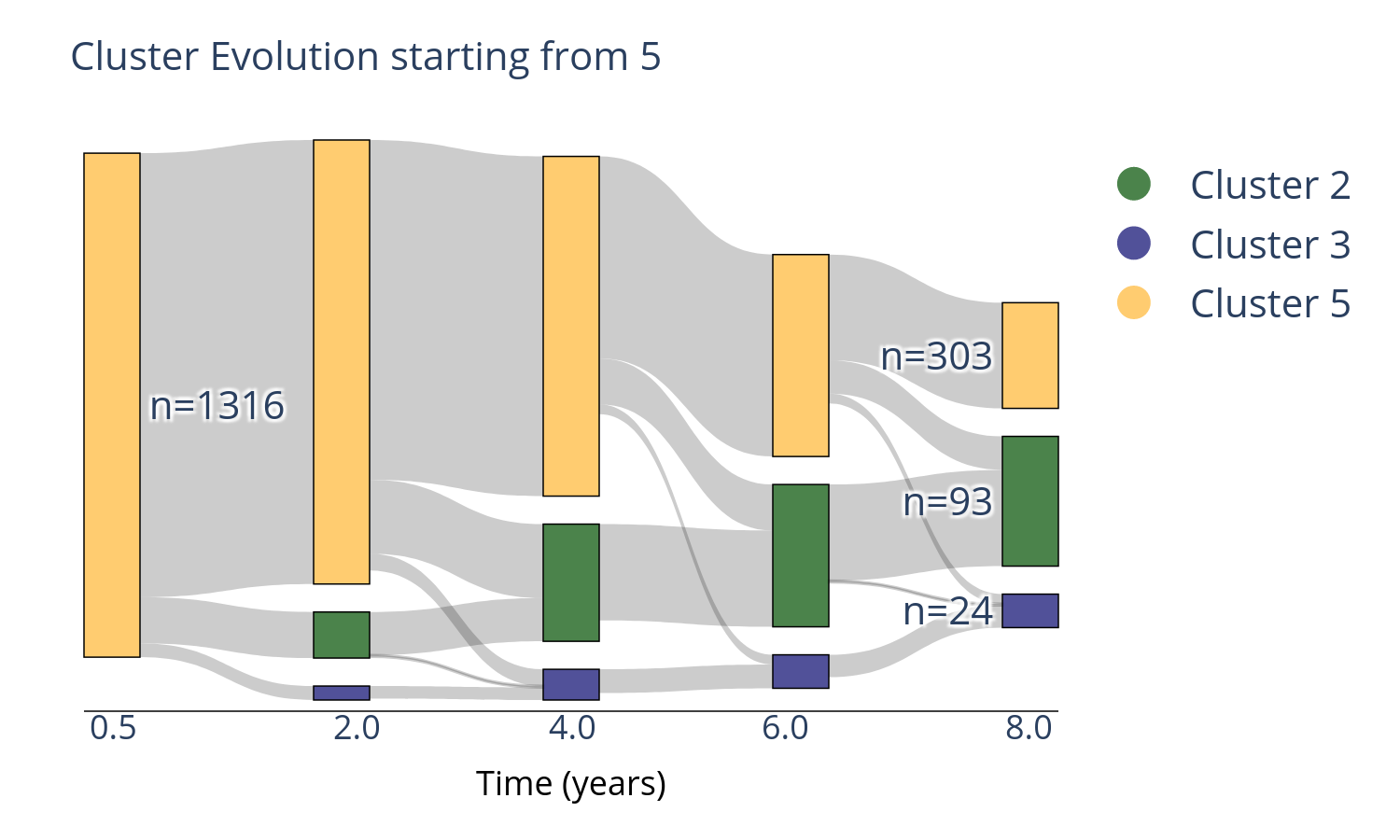}
    \end{subfigure}
     \begin{subfigure}[b]{0.31\textwidth}
        \centering
         \includegraphics[width=1\textwidth]{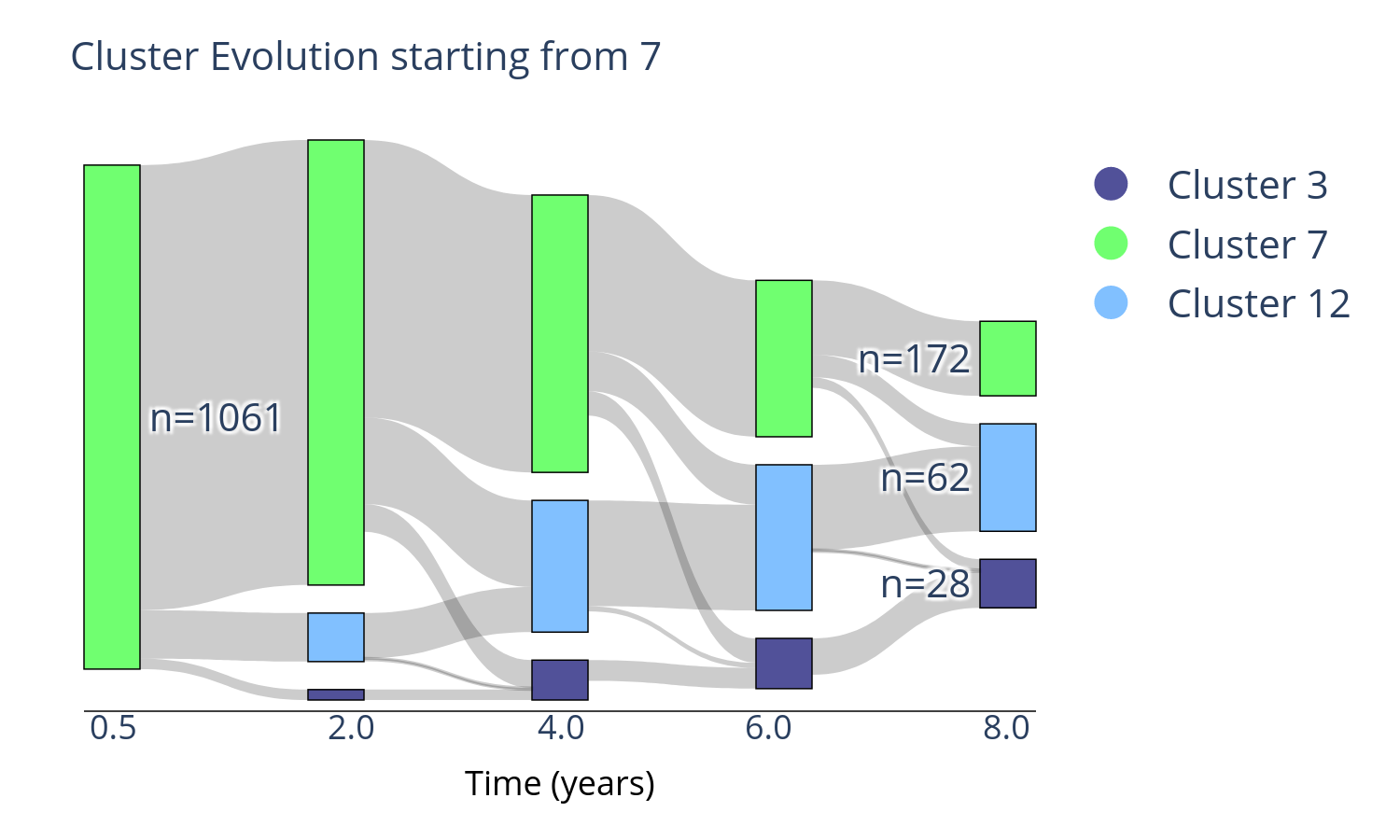}
    \end{subfigure}
     \begin{subfigure}[b]{0.31\textwidth}
        \centering
         \includegraphics[width=1\textwidth]{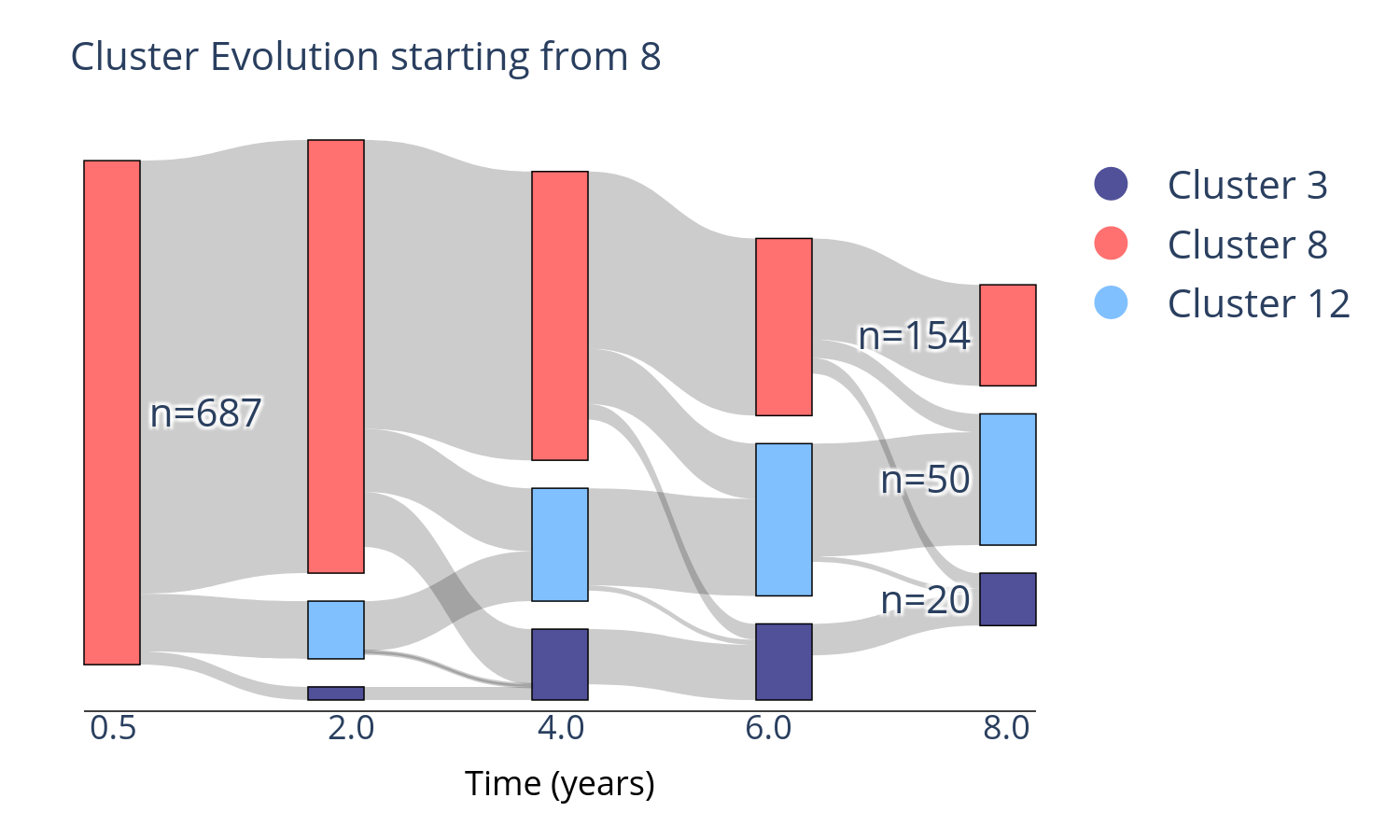}
    \end{subfigure}
     \begin{subfigure}[b]{0.31\textwidth}
        \centering
         \includegraphics[width=1\textwidth]{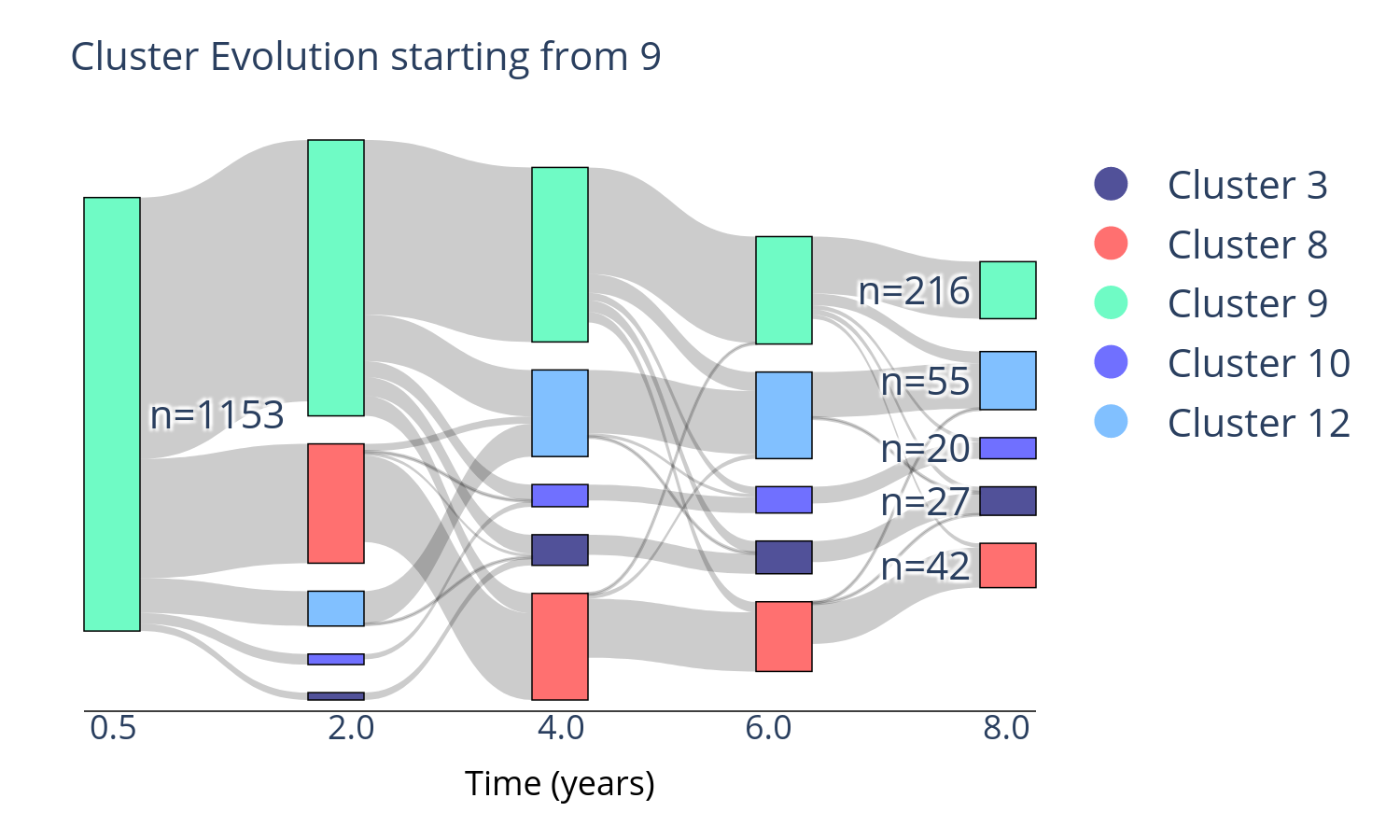}
    \end{subfigure}
    \begin{subfigure}[b]{0.31\textwidth}
        \centering
         \includegraphics[width=1\textwidth]{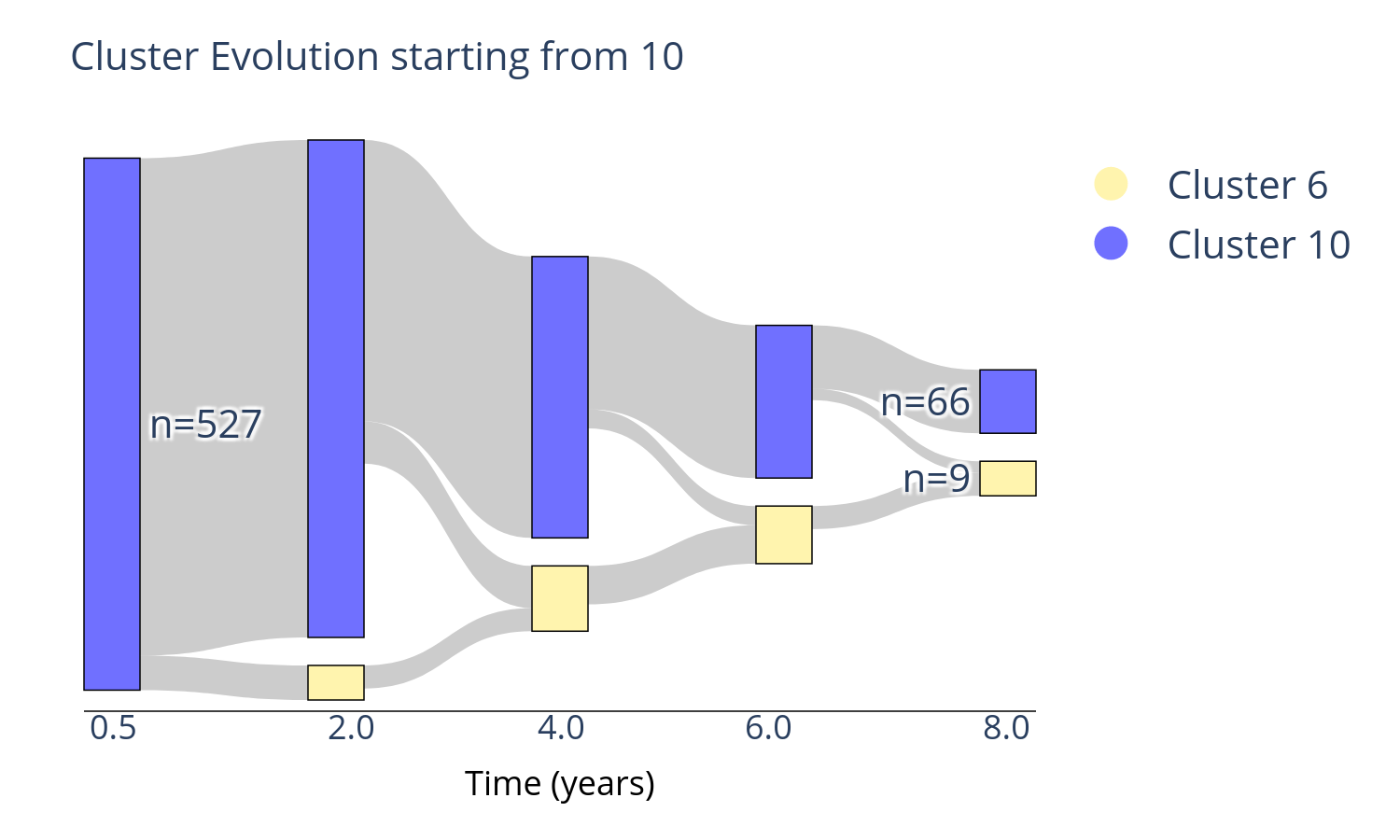}
    \end{subfigure}
    \begin{subfigure}[b]{0.31\textwidth}
        \centering
         \includegraphics[width=1\textwidth]{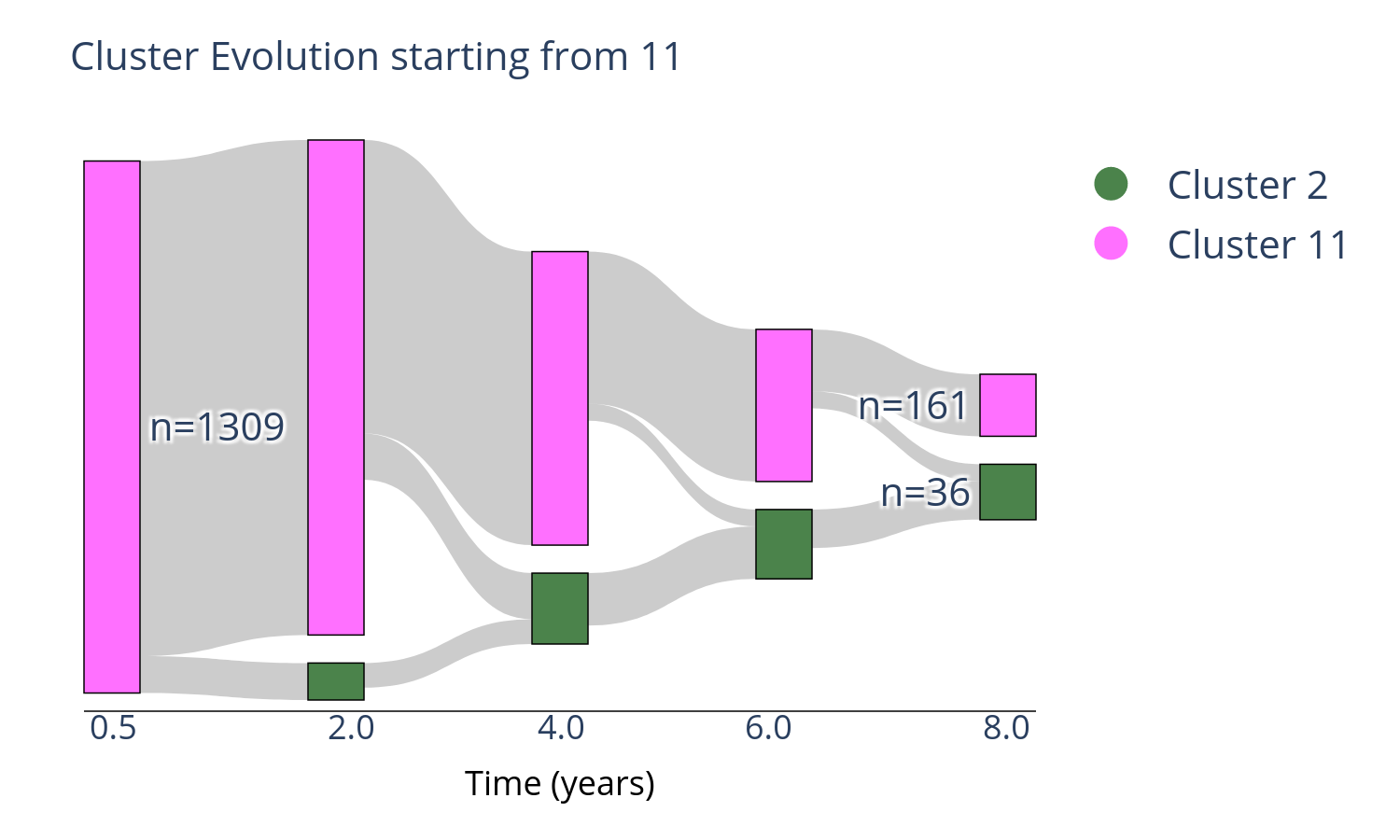}
    \end{subfigure}
    \begin{subfigure}[b]{0.31\textwidth}
        \centering
         \includegraphics[width=1\textwidth]{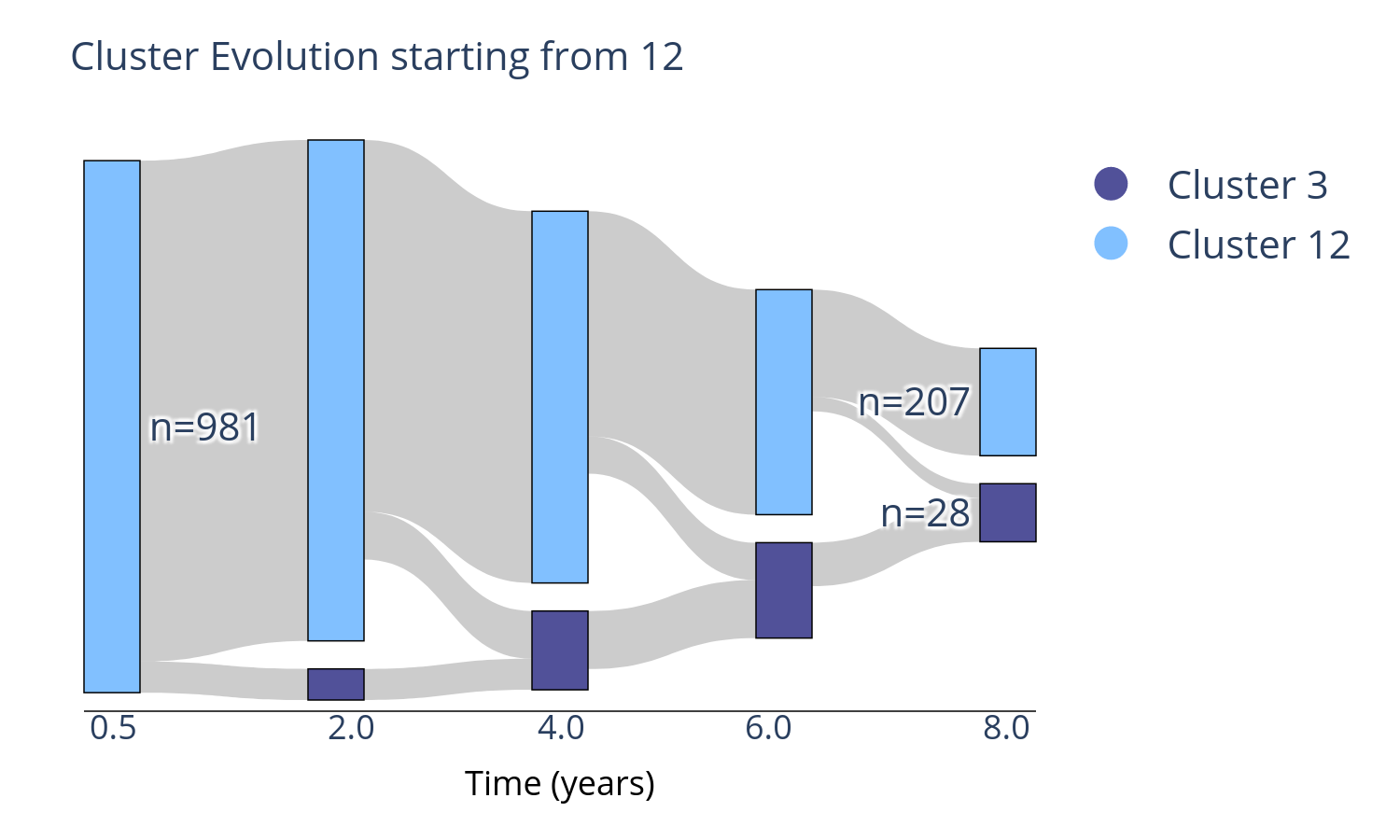}
    \end{subfigure}
    \begin{subfigure}[b]{0.31\textwidth}
        \centering
         \includegraphics[width=1\textwidth]{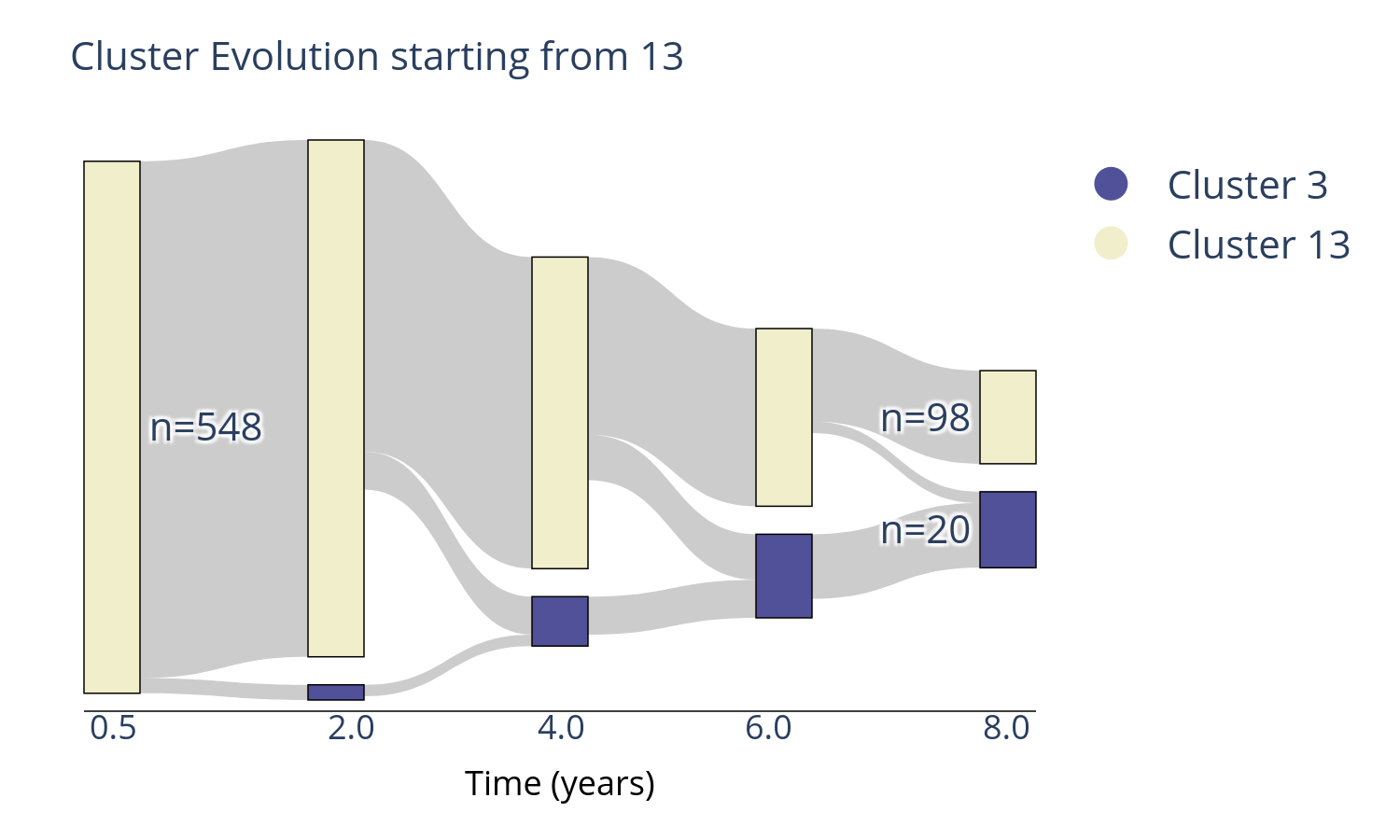}
    \end{subfigure}
     \begin{subfigure}[b]{0.31\textwidth}
        \centering
         \includegraphics[width=1\textwidth]{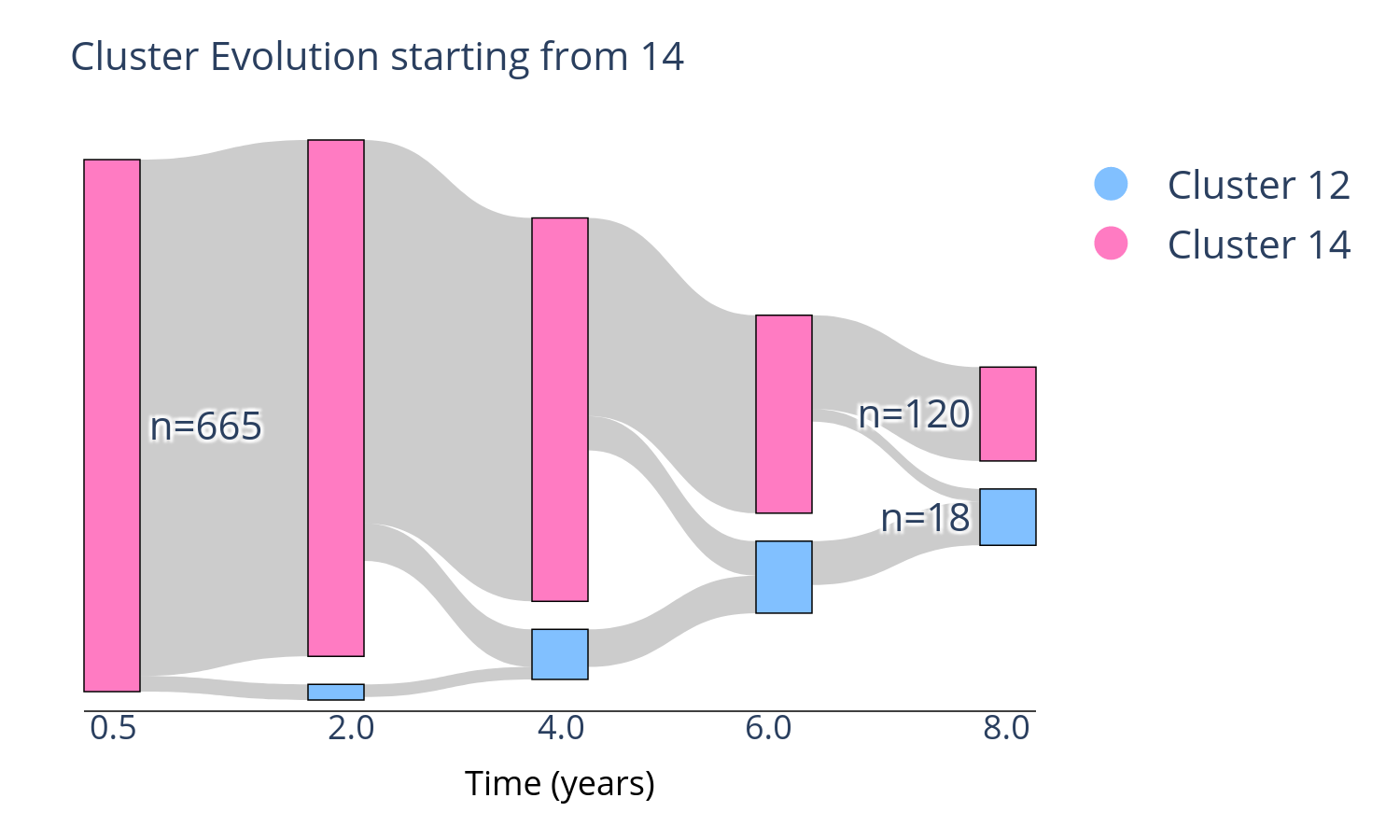}
    \end{subfigure}
    \caption{Evolution of patient clusters over time (\textit{Men $<65$}), conditional on the initial cluster assignment (6 months post-NIAD). The number of patients in each cluster is indicated alongside the corresponding bars. For clarity, only the most frequent transitions are shown (see Figure \ref{fig:male_m65_transition_prob} for more details). Cluster 6 is excluded due to the low number of transitions observed from this cluster.}
    \label{fig:male_m65_evolution}
\end{figure}

\begin{figure}[ht]
    \centering
     \includegraphics[width=0.8\textwidth]{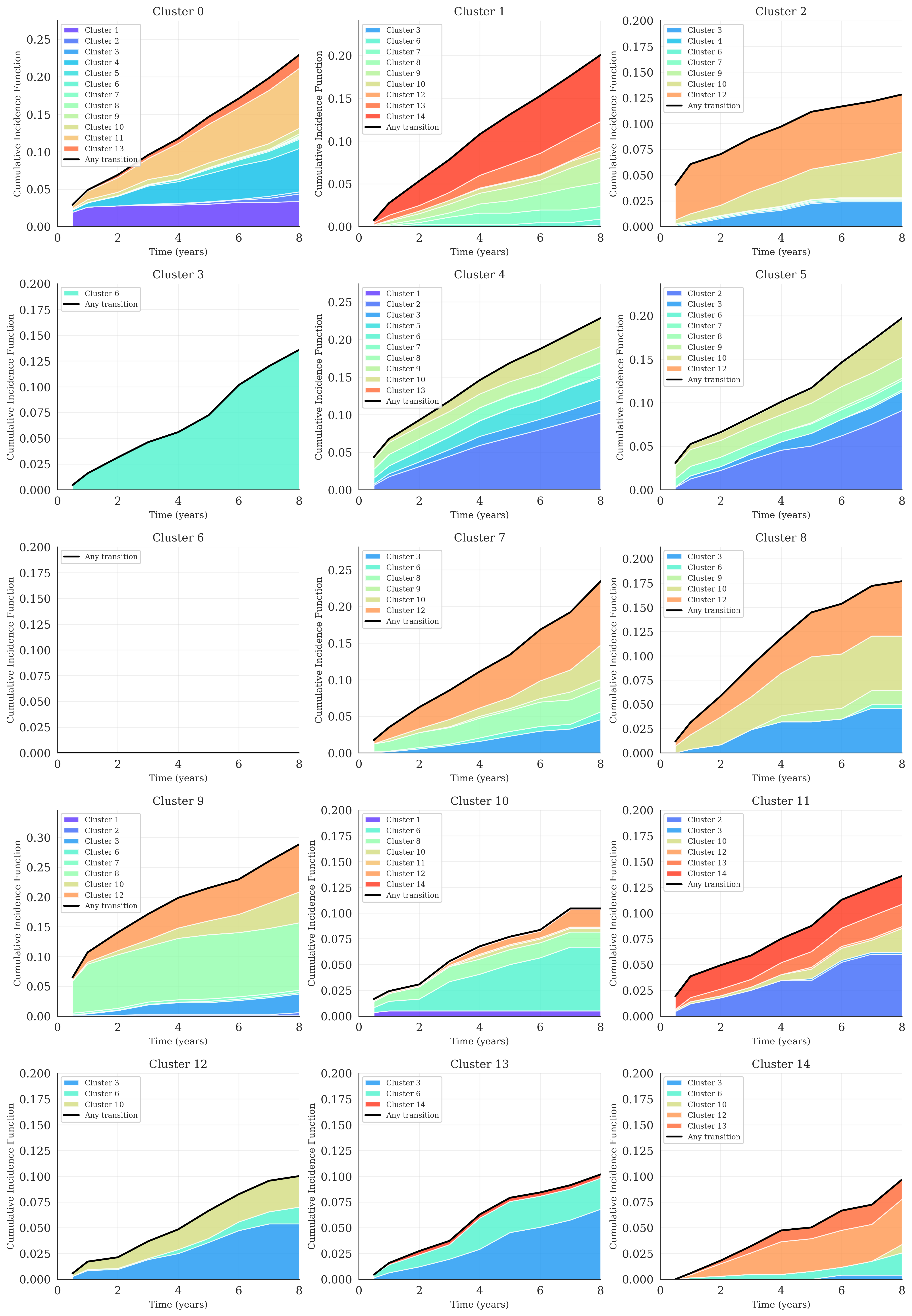}
        \caption{
Cumulative Incidence Function (CIF) for the time to first cluster transition in \textit{Men $<65$}, stratified by initial cluster assignment at 6 months post-NIAD.
The estimates account for censoring due to loss to follow-up and study termination, computed using the Aalen nonparametric estimator \cite{aalen1978}.
}
    \label{fig:male_m65_transition_prob}
\end{figure}

\begin{figure}[ht]
    \centering
    \includegraphics[width=1\linewidth]{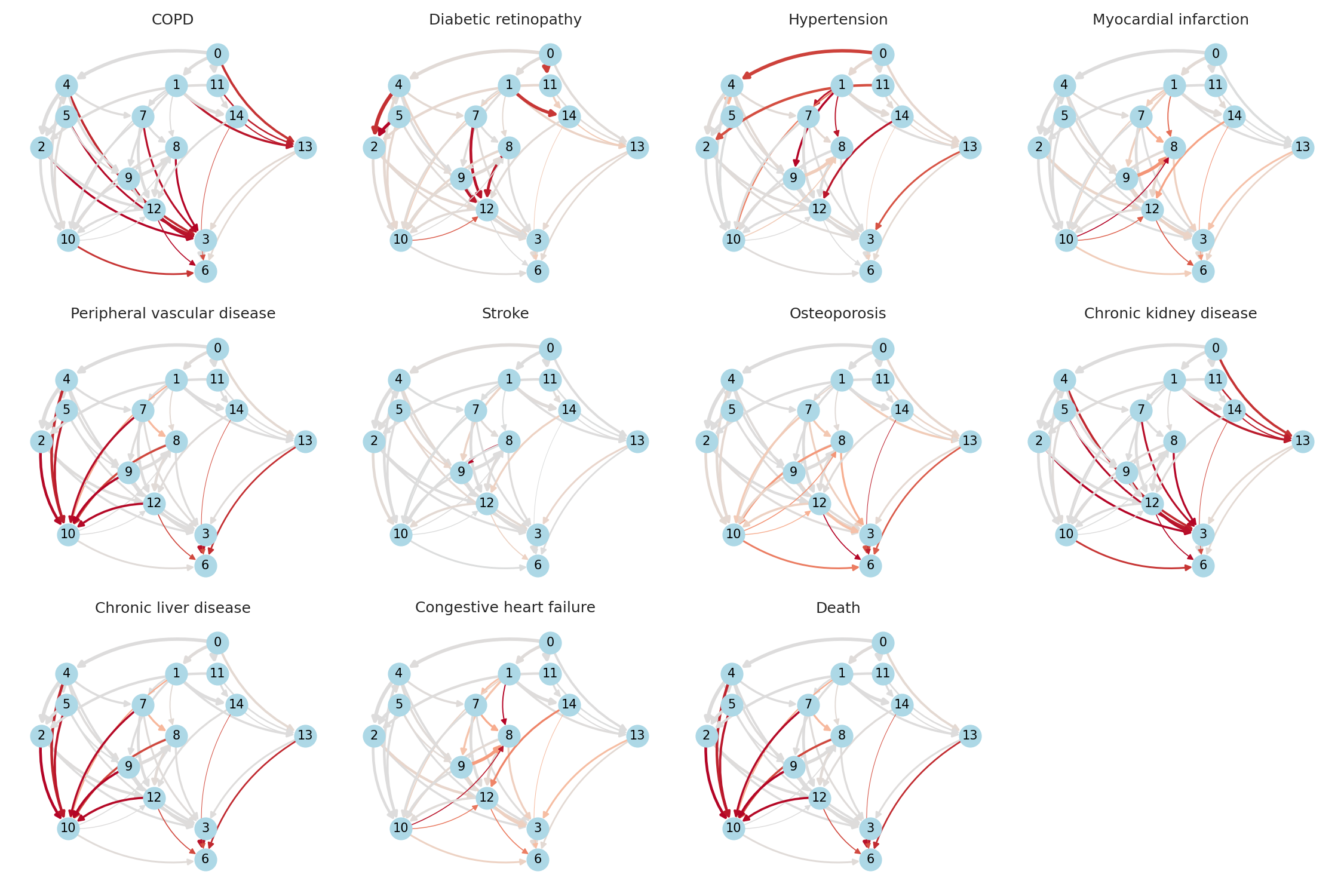}
    \caption{Transition graph. Red arrows (resp. blue) means an increase (resp. decrease) of the comorbidity risk (\textit{Men $<65$}). Arrow thickness is proportional to the number of observed transitions. Transitions with less than 1\% probability are not shown.}
    \label{fig:male_m65_graph}
\end{figure}

\clearpage
\section{Male $65+$}
\begin{table}[ht!]
    \centering
    \resizebox{\textwidth}{!}{
    \begin{tabular}{l|ccccccccccccc}
    \toprule
Cluster & 0 & 1 & 2 & 3 & 4 & 5 & 6 & 7 & 8 & 9 & 10 & 11 & 12\\
    \midrule
\# of patients & 2732 & 1515 & 1813 & 838 & 1974 & 1430 & 1125 & 1214 & 1954 & 3028 & 769 & 762 & 1319\\
Avg. age at NIAD (y) $\pm$ SD & 72.61 $\pm$ 6.01 & 72.22 $\pm$ 5.62 & 73.01 $\pm$ 5.74 & 75.48 $\pm$ 6.69 & 74.38 $\pm$ 6.09 & 74.57 $\pm$ 6.37 & 75.52 $\pm$ 6.54 & 73.3 $\pm$ 6.12 & 74.73 $\pm$ 6.57 & 72.0 $\pm$ 5.39 & 75.14 $\pm$ 6.76 & 73.46 $\pm$ 6.24 & 74.14 $\pm$ 6.3\\
Avg. follow-up post-NIAD (y) $\pm$ SD & 4.4 $\pm$ 3.1 & 5.31 $\pm$ 2.86 & 5.08 $\pm$ 2.85 & 4.45 $\pm$ 2.8 & 4.9 $\pm$ 2.9 & 5.15 $\pm$ 3.28 & 4.37 $\pm$ 2.85 & 5.79 $\pm$ 3.08 & 4.04 $\pm$ 2.78 & 4.46 $\pm$ 3.1 & 4.89 $\pm$ 2.81 & 5.35 $\pm$ 2.73 & 4.96 $\pm$ 2.97\\
Avg. BMI at NIAD (kg/m\textsuperscript{2}) $\pm$ SD & 28.68 $\pm$ 4.35 & 30.01 $\pm$ 4.87 & 29.41 $\pm$ 3.84 & 29.9 $\pm$ 5.1 & 30.54 $\pm$ 5.51 & 29.78 $\pm$ 3.97 & 30.13 $\pm$ 5.43 & 34.35 $\pm$ 6.26 & 29.0 $\pm$ 3.88 & 29.71 $\pm$ 4.17 & 32.81 $\pm$ 5.74 & 33.89 $\pm$ 6.31 & 28.94 $\pm$ 5.19\\
Avg. HbA1c at NIAD (\%) $\pm$ SD & 6.96 $\pm$ 1.33 & 7.33 $\pm$ 1.41 & 7.16 $\pm$ 1.36 & 7.09 $\pm$ 1.39 & 6.94 $\pm$ 1.33 & 6.95 $\pm$ 1.24 & 7.12 $\pm$ 1.36 & 6.87 $\pm$ 1.3 & 7.11 $\pm$ 1.37 & 6.92 $\pm$ 1.28 & 7.01 $\pm$ 1.36 & 7.21 $\pm$ 1.38 & 7.1 $\pm$ 1.36\\
\% Ever smoked & 29.61 & 31.22 & 26.86 & 27.92 & 45.54 & 27.48 & 56.44 & 27.68 & 41.91 & 27.87 & 34.72 & 29.27 & 48.6\\
\% Ever alcohol & 89.09 & 90.83 & 94.65 & 91.53 & 93.31 & 88.67 & 90.58 & 90.86 & 90.69 & 90.62 & 91.94 & 93.31 & 91.51\\
    \midrule
\multicolumn{14}{l}{\textit{Comorbidities at NIAD [N,(\%)]}}\\
Stroke & 189 (6.92) & 111 (7.33) & $<1$ \% & 748 (89.26) & 218 (11.04) & 45 (3.15) & 196 (17.42) & 88 (7.25) & 316 (16.17) & 31 (1.02) & 154 (20.03) & 60 (7.87) & 139 (10.54)\\
Myocardial infarction & 201 (7.36) & 170 (11.22) & $<1$ \% & $<1$ \% & $<1$ \% & $<1$ \% & 579 (51.47) & $<1$ \% & 1156 (59.16) & $<1$ \% & 450 (58.52) & $<1$ \% & 209 (15.85)\\
Congestive heart failure & 72 (2.64) & 77 (5.08) & 35 (1.93) & 45 (5.37) & 144 (7.29) & 175 (12.24) & 233 (20.71) & 59 (4.86) & 258 (13.20) & $<1$ \% & 166 (21.59) & 52 (6.82) & 133 (10.08)\\
Peripheral vascular disease & $<1$ \% & $<1$ \% & $<1$ \% & $<1$ \% & $<1$ \% & $<1$ \% & 478 (42.49) & $<1$ \% & 731 (37.41) & $<1$ \% & 295 (38.36) & $<1$ \% & $<1$ \%\\
Diabetic retinopathy & 28 (1.02) & 667 (44.03) & 1179 (65.03) & 207 (24.70) & 227 (11.50) & 25 (1.75) & 200 (17.78) & $<1$ \% & 393 (20.11) & 55 (1.82) & 168 (21.85) & 465 (61.02) & 222 (16.83)\\
Hypertension & 118 (4.32) & 30 (1.98) & 1774 (97.85) & 805 (96.06) & 1911 (96.81) & 1377 (96.29) & 849 (75.47) & 1159 (95.47) & 1504 (76.97) & 2893 (95.54) & 542 (70.48) & 730 (95.80) & 27 (2.05)\\
Chronic kidney disease & 53 (1.94) & 52 (3.43) & $<1$ \% & $<1$ \% & 68 (3.44) & $<1$ \% & 66 (5.87) & 345 (28.42) & $<1$ \% & $<1$ \% & 290 (37.71) & 289 (37.93) & 46 (3.49)\\
Osteoporosis & $<1$ \% & $<1$ \% & $<1$ \% & $<1$ \% & 39 (1.98) & $<1$ \% & 28 (2.49) & $<1$ \% & $<1$ \% & $<1$ \% & 13 (1.69) & 14 (1.84) & 37 (2.81)\\
COPD & $<1$ \% & $<1$ \% & $<1$ \% & $<1$ \% & 1063 (53.85) & $<1$ \% & 609 (54.13) & $<1$ \% & $<1$ \% & $<1$ \% & $<1$ \% & 29 (3.81) & 669 (50.72)\\
Chronic liver disease & 36 (1.32) & 26 (1.72) & 31 (1.71) & 14 (1.67) & 34 (1.72) & 24 (1.68) & 27 (2.40) & $<1$ \% & 28 (1.43) & 54 (1.78) & 16 (2.08) & 37 (4.86) & 21 (1.59)\\
    \midrule
\multicolumn{14}{l}{\textit{Comorbidities during follow-up [N,(\%)]}}\\
Stroke & 94 (3.44) & 63 (4.16) & 65 (3.59) & 90 (10.74) & 99 (5.02) & 50 (3.50) & 82 (7.29) & 47 (3.87) & 105 (5.37) & 89 (2.94) & 45 (5.85) & 42 (5.51) & 53 (4.02)\\
Myocardial infarction & 47 (1.72) & 38 (2.51) & 22 (1.21) & 10 (1.19) & 42 (2.13) & 22 (1.54) & 67 (5.96) & 18 (1.48) & 91 (4.66) & 40 (1.32) & 41 (5.33) & 26 (3.41) & 23 (1.74)\\
Congestive heart failure & 68 (2.49) & 68 (4.49) & 59 (3.25) & 30 (3.58) & 118 (5.98) & 104 (7.27) & 113 (10.04) & 72 (5.93) & 120 (6.14) & 51 (1.68) & 66 (8.58) & 63 (8.27) & 62 (4.70)\\
Peripheral vascular disease & $<1$ \% & 22 (1.45) & 19 (1.05) & 19 (2.27) & 51 (2.58) & 15 (1.05) & 115 (10.22) & 13 (1.07) & 150 (7.68) & $<1$ \% & 52 (6.76) & 12 (1.57) & 25 (1.90)\\
Diabetic retinopathy & 214 (7.83) & 510 (33.66) & 615 (33.92) & 152 (18.14) & 341 (17.27) & 159 (11.12) & 216 (19.20) & 132 (10.87) & 311 (15.92) & 296 (9.78) & 148 (19.25) & 221 (29.00) & 303 (22.97)\\
Hypertension & 245 (8.97) & 140 (9.24) & 32 (1.77) & 15 (1.79) & 41 (2.08) & 21 (1.47) & 47 (4.18) & 33 (2.72) & 54 (2.76) & 88 (2.91) & 19 (2.47) & 21 (2.76) & 98 (7.43)\\
Chronic kidney disease & 84 (3.07) & 62 (4.09) & 47 (2.59) & 26 (3.10) & 89 (4.51) & 46 (3.22) & 49 (4.36) & 166 (13.67) & 49 (2.51) & 71 (2.34) & 120 (15.60) & 110 (14.44) & 70 (5.31)\\
Osteoporosis & $<1$ \% & $<1$ \% & $<1$ \% & 10 (1.19) & $<1$ \% & $<1$ \% & 16 (1.42) & $<1$ \% & $<1$ \% & $<1$ \% & $<1$ \% & 14 (1.84) & $<1$ \%\\
COPD & 42 (1.54) & 33 (2.18) & 35 (1.93) & 13 (1.55) & 193 (9.78) & 31 (2.17) & 142 (12.62) & 27 (2.22) & 38 (1.94) & 41 (1.35) & 22 (2.86) & 26 (3.41) & 120 (9.10)\\
Chronic liver disease & 28 (1.02) & 30 (1.98) & 20 (1.10) & 13 (1.55) & 26 (1.32) & $<1$ \% & 17 (1.51) & 22 (1.81) & $<1$ \% & 31 (1.02) & 21 (2.73) & 18 (2.36) & 17 (1.29)\\
Deaths before 2020 & 344 (12.59) & 205 (13.53) & 200 (11.03) & 195 (23.27) & 472 (23.91) & 254 (17.76) & 384 (34.13) & 222 (18.29) & 424 (21.70) & 315 (10.40) & 219 (28.48) & 163 (21.39) & 309 (23.43)\\
    \bottomrule
\end{tabular}
    }
    \caption{Patient characteristics in each cluster for the \textit{Men $65+$} cohort with average values and standard deviation where applicable. Comorbidities are presented as absolute counts at NIAD initiation, along with the number of new comorbidities developed post-NIAD. Proportions, expressed as percentages of the cohort population, are reported in parentheses.}
    \label{tab:cluster_descr_male_65+}
\end{table}

\begin{figure}[ht]
    \centering
    \includegraphics[width=0.9\linewidth]{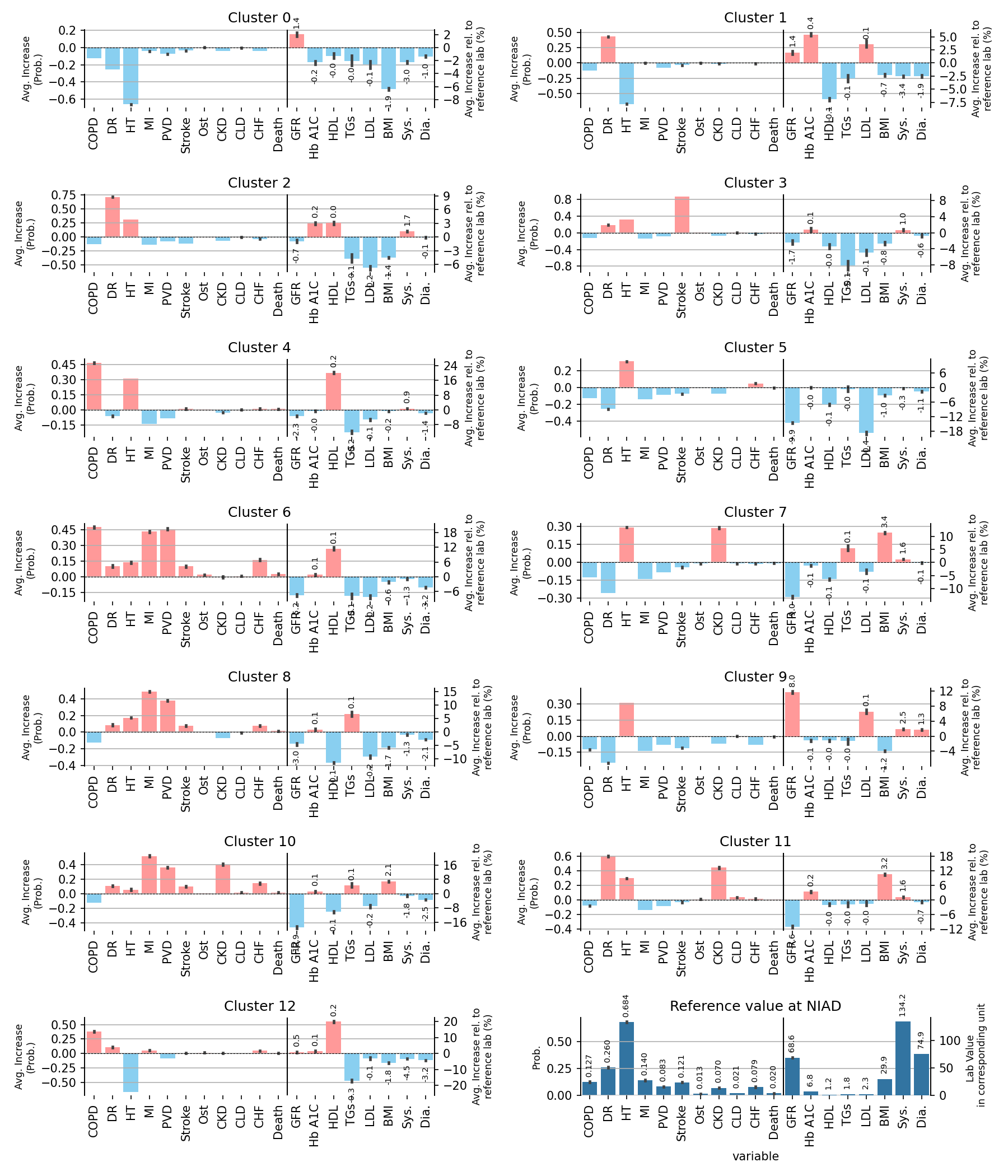}
    \caption{Average cluster-specific probabilities and lab values for \textit{Men $65+$}. Values have been centered by subtracting the average at the time of NIAD initiation to improve readability. Red bars indicate increases, and blue bars indicate decreases relative to the average patient at NIAD. Comorbidity risks are shown as changes in probability (left axis), and lab values as percentage changes relative to the reference value (right axis). Annotated numbers indicate absolute changes in the original units (see Table \ref{tab:list_of_var}). Error bars represent the standard error of the mean.}
    \label{fig:male65+_proba_avg}
\end{figure}

\begin{figure}[ht]
    \centering
   \begin{subfigure}[b]{0.31\textwidth}
        \centering
         \includegraphics[width=1\textwidth]{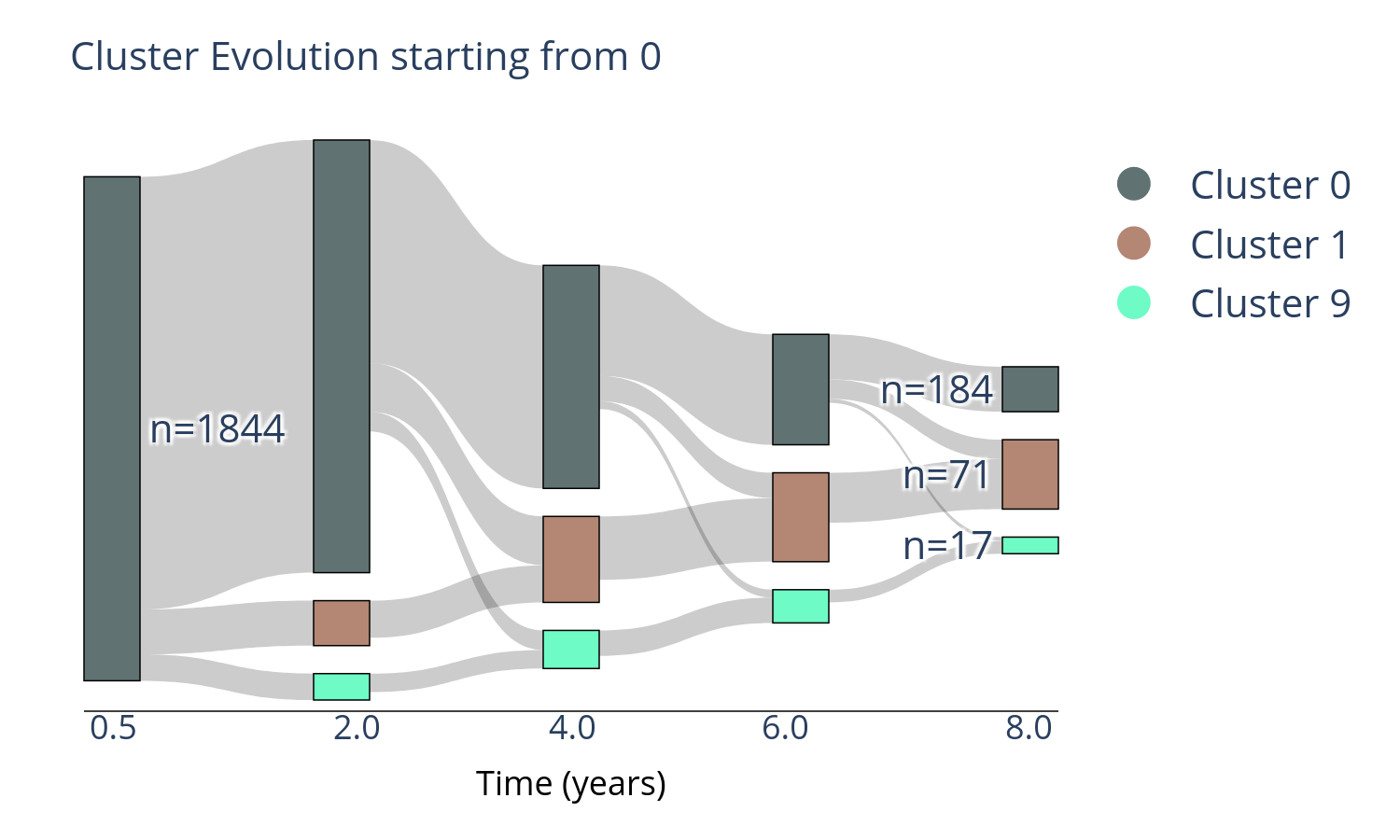}
    \end{subfigure}
    \begin{subfigure}[b]{0.31\textwidth}
        \centering
         \includegraphics[width=1\textwidth]{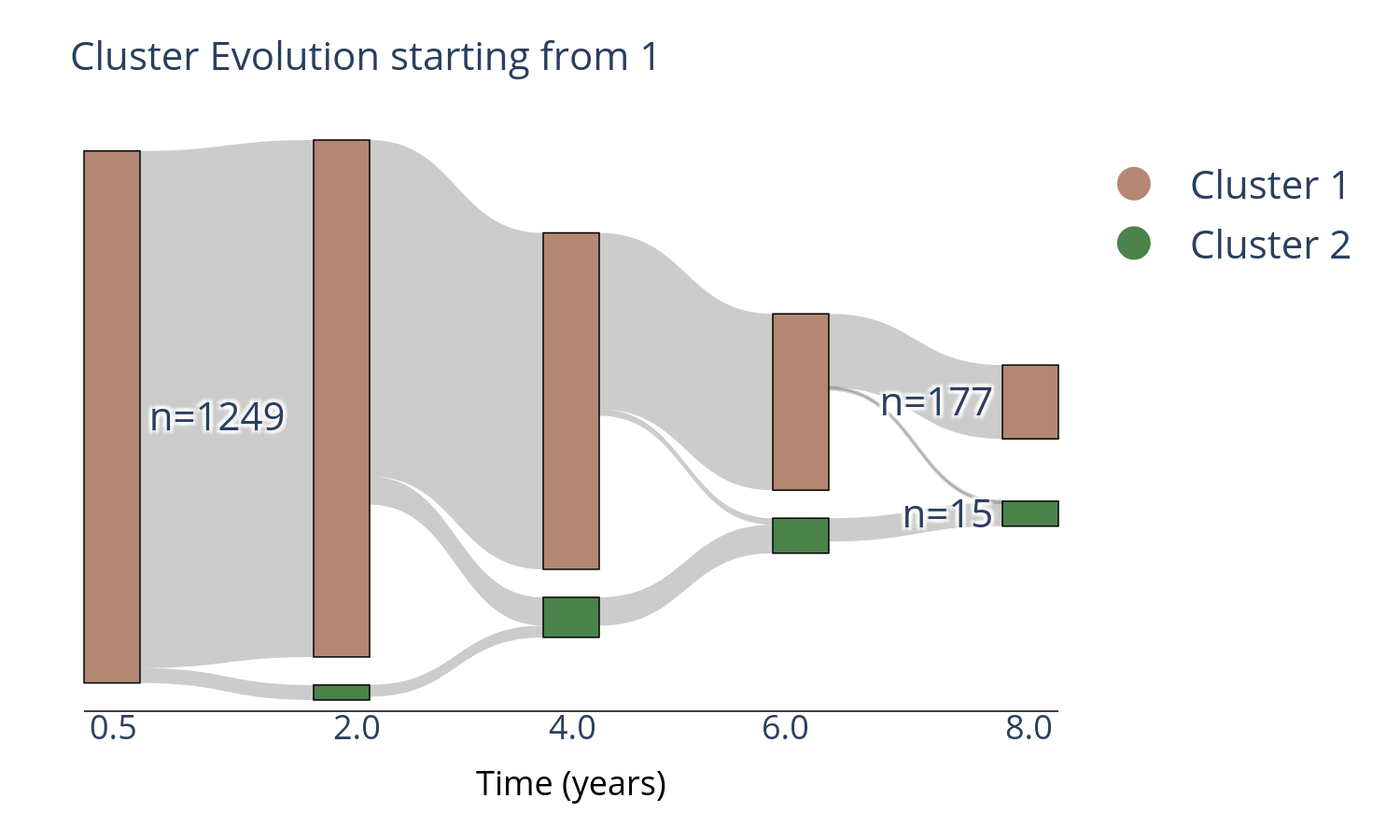}
    \end{subfigure}
     \begin{subfigure}[b]{0.31\textwidth}
        \centering
         \includegraphics[width=1\textwidth]{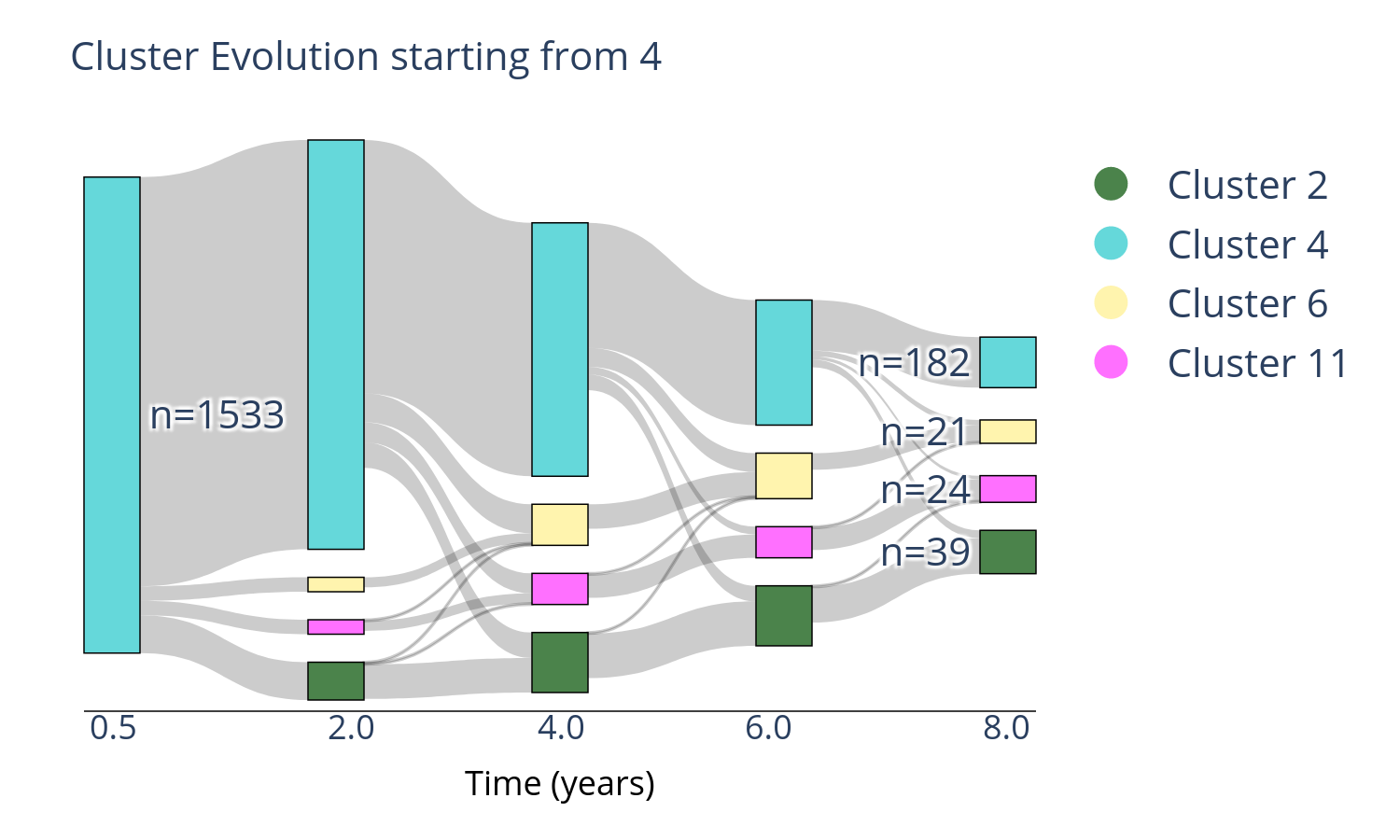}
    \end{subfigure}
     \begin{subfigure}[b]{0.31\textwidth}
        \centering
         \includegraphics[width=1\textwidth]{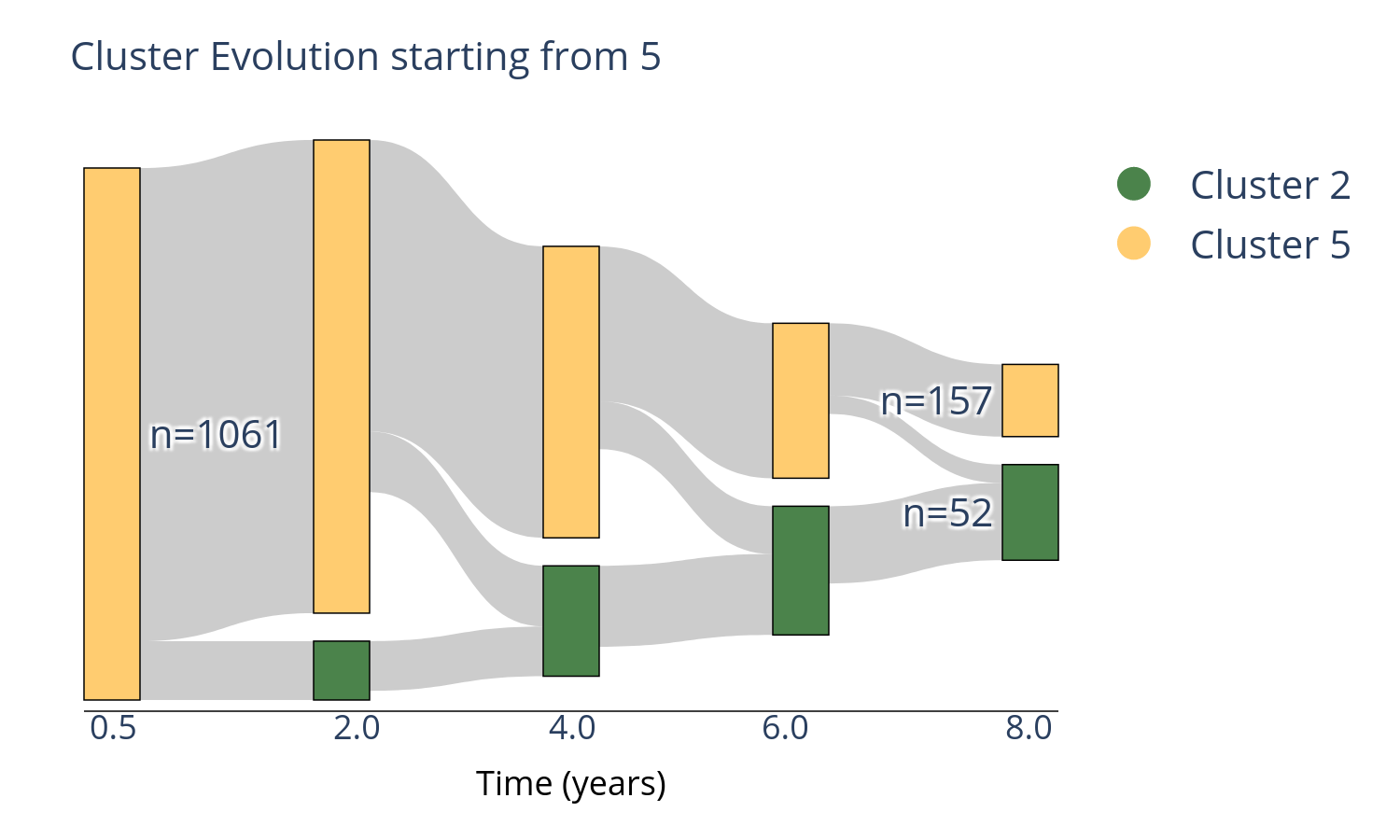}
    \end{subfigure}
     \begin{subfigure}[b]{0.31\textwidth}
        \centering
         \includegraphics[width=1\textwidth]{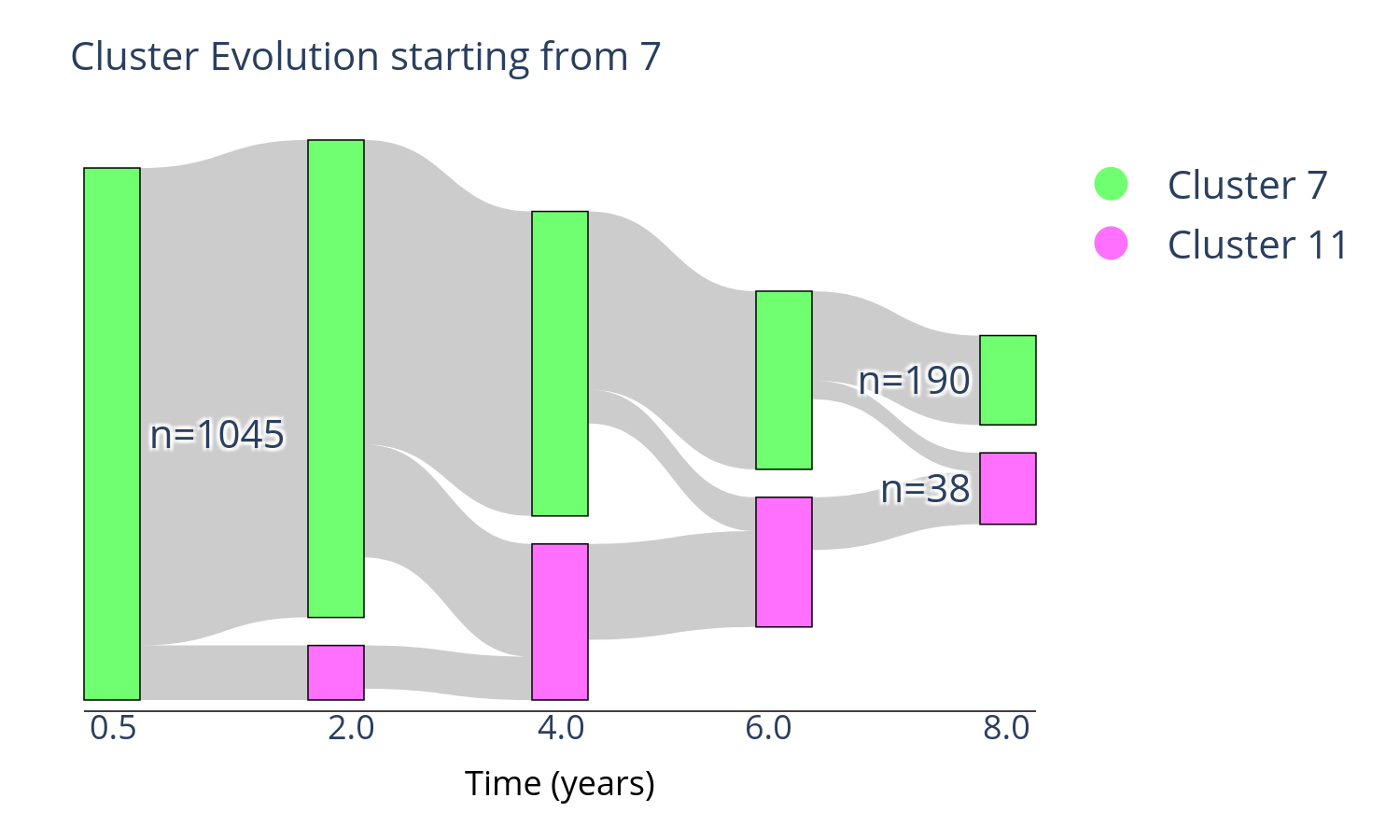}
    \end{subfigure}
    \begin{subfigure}[b]{0.31\textwidth}
        \centering
         \includegraphics[width=1\textwidth]{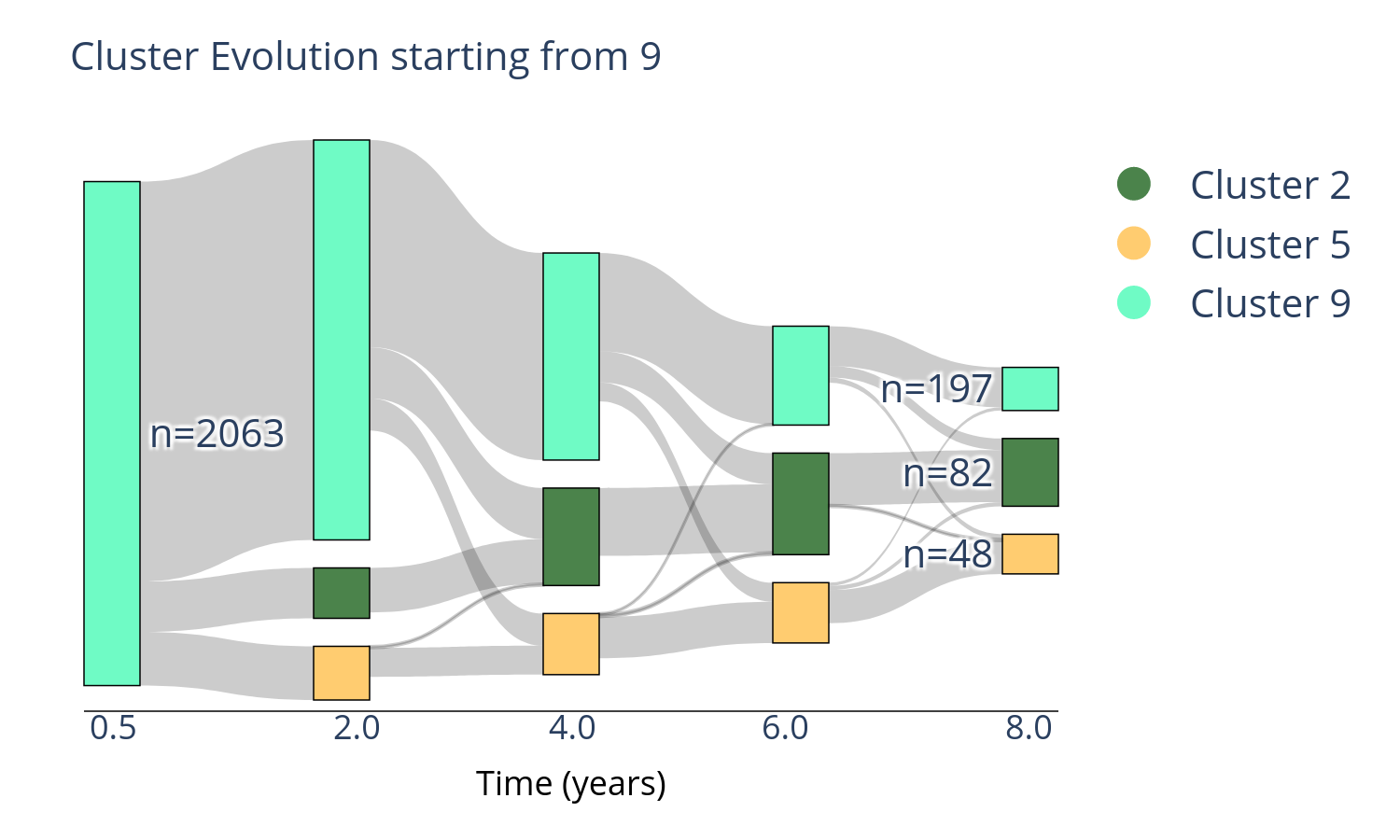}
    \end{subfigure}
    \begin{subfigure}[b]{0.31\textwidth}
        \centering
         \includegraphics[width=1\textwidth]{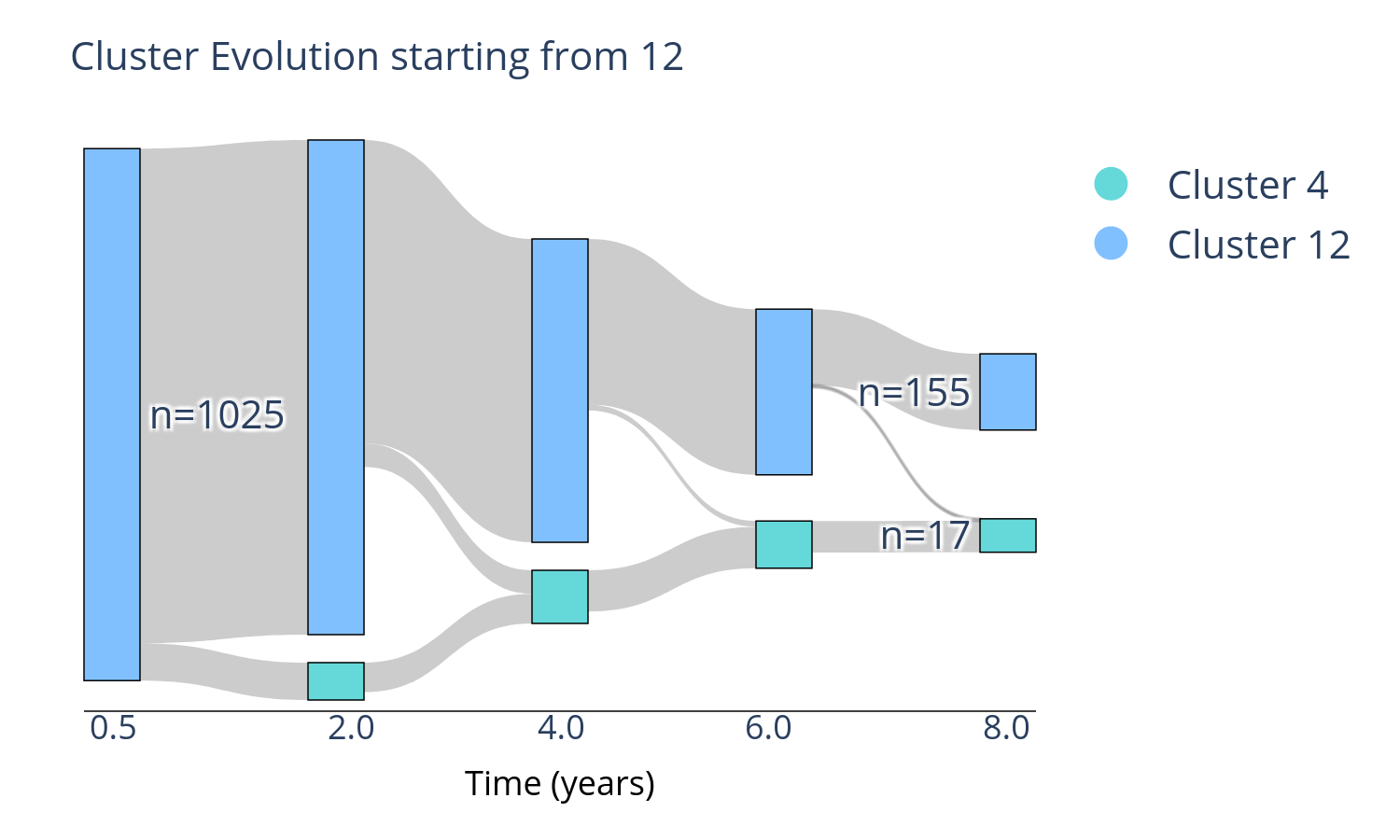}
    \end{subfigure}
    \caption{Evolution of patient clusters over time (\textit{Men $65+$}), conditional on the initial cluster assignment (6 months post-NIAD). The number of patients in each cluster is indicated alongside the corresponding bars. For clarity, only the most frequent transitions are shown (see Figure \ref{fig:male65+_transition_prob} for more details). Clusters 2,3,6,8,10,11 are excluded due to the low number of transitions observed from these clusters.}
    \label{fig:male65+_evolution}
\end{figure}

\begin{figure}[ht]
    \centering
       \includegraphics[width=0.8\textwidth]{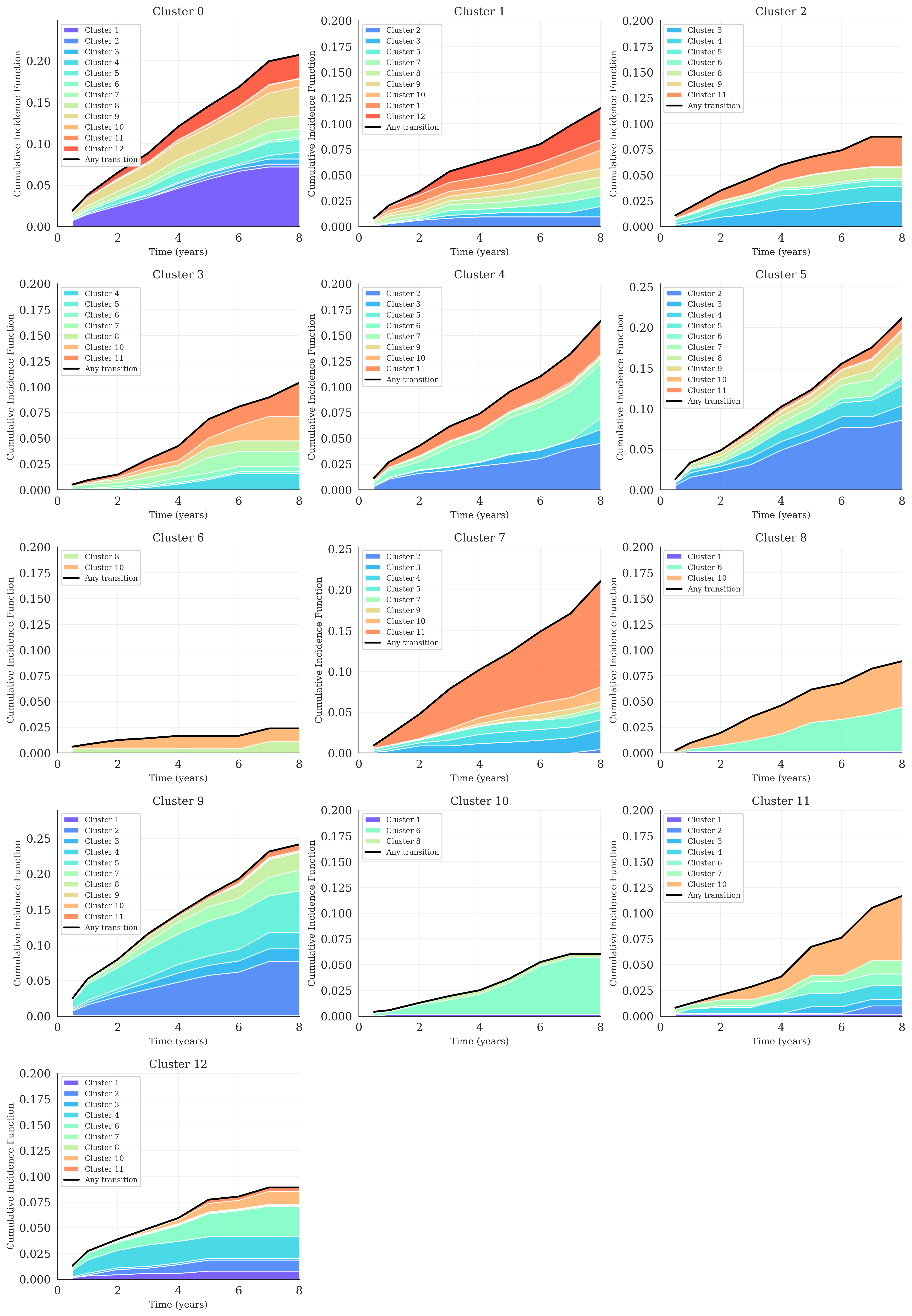}
        \caption{
Cumulative Incidence Function (CIF) for the time to first cluster transition in \textit{Men $65+$}, stratified by initial cluster assignment at 6 months post-NIAD.
The estimates account for censoring due to loss to follow-up and study termination, computed using the Aalen nonparametric estimator \cite{aalen1978}.
}
    \label{fig:male65+_transition_prob}
\end{figure}

\begin{figure}[ht]
    \centering
    \includegraphics[width=1\linewidth]{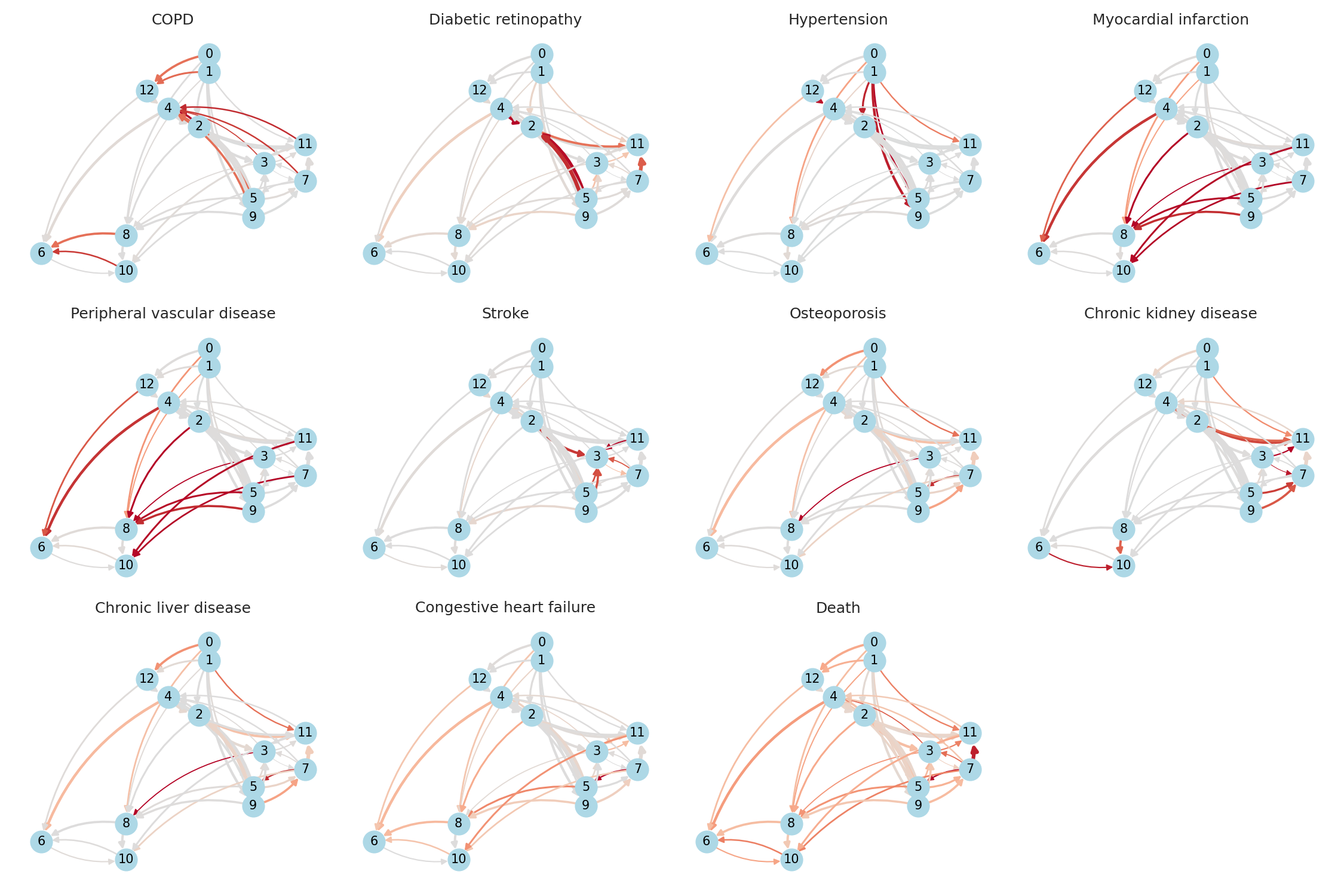}
    \caption{Transition graph. Red arrows (resp. blue) means an increase (resp. decrease) of the comorbidity risk (\textit{Men $65+$}). Arrow thickness is proportional to the number of observed transitions. Transitions with less than 1\% probability are not shown.}
    \label{fig:male65+_graph}
\end{figure}

\clearpage
\section{Female $65+$}
\begin{table}[ht!]
    \centering
    \resizebox{\textwidth}{!}{
\begin{tabular}{l|ccccccccccccc}
    \toprule
Cluster & 0 & 1 & 2 & 3 & 4 & 5 & 6 & 7 & 8 & 9 & 10 & 11 & 12\\
    \midrule
\# of patients & 2604 & 857 & 1629 & 742 & 1626 & 1276 & 787 & 1258 & 681 & 3702 & 716 & 644 & 452\\
Avg. age at NIAD (y) $\pm$ SD & 72.7 $\pm$ 5.8 & 72.98 $\pm$ 5.79 & 74.91 $\pm$ 6.32 & 74.08 $\pm$ 6.18 & 76.72 $\pm$ 6.69 & 79.47 $\pm$ 7.44 & 76.98 $\pm$ 7.06 & 73.32 $\pm$ 5.86 & 78.33 $\pm$ 7.13 & 73.24 $\pm$ 5.72 & 73.69 $\pm$ 6.0 & 77.85 $\pm$ 7.15 & 74.04 $\pm$ 6.35\\
Avg. follow-up post-NIAD (y) $\pm$ SD & 4.7 $\pm$ 3.12 & 5.11 $\pm$ 2.68 & 5.07 $\pm$ 2.71 & 4.35 $\pm$ 2.77 & 5.52 $\pm$ 3.12 & 3.5 $\pm$ 2.66 & 3.79 $\pm$ 2.64 & 6.0 $\pm$ 3.23 & 4.04 $\pm$ 2.31 & 4.82 $\pm$ 3.24 & 5.7 $\pm$ 2.68 & 4.24 $\pm$ 2.78 & 4.91 $\pm$ 2.73\\
Avg. BMI at NIAD (kg/m\textsuperscript{2}) $\pm$ SD & 30.24 $\pm$ 6.15 & 29.99 $\pm$ 5.92 & 29.84 $\pm$ 4.83 & 32.63 $\pm$ 6.74 & 28.96 $\pm$ 5.31 & 27.99 $\pm$ 4.72 & 29.53 $\pm$ 6.46 & 36.76 $\pm$ 7.37 & 30.7 $\pm$ 6.34 & 30.74 $\pm$ 4.8 & 36.67 $\pm$ 7.09 & 34.36 $\pm$ 7.83 & 31.65 $\pm$ 6.33\\
Avg. HbA1c at NIAD (\%) $\pm$ SD & 6.96 $\pm$ 1.27 & 7.24 $\pm$ 1.32 & 7.09 $\pm$ 1.29 & 7.08 $\pm$ 1.39 & 6.83 $\pm$ 1.3 & 7.08 $\pm$ 1.35 & 7.15 $\pm$ 1.42 & 6.98 $\pm$ 1.26 & 7.37 $\pm$ 1.49 & 6.9 $\pm$ 1.25 & 7.05 $\pm$ 1.18 & 7.09 $\pm$ 1.41 & 7.32 $\pm$ 1.37\\
\% Ever smoked & 30.72 & 32.79 & 24.86 & 64.56 & 18.02 & 31.03 & 43.96 & 24.24 & 37.0 & 21.47 & 25.56 & 33.7 & 23.67\\
\% Ever alcohol & 79.26 & 86.23 & 86.68 & 83.42 & 83.95 & 77.59 & 74.84 & 80.21 & 85.32 & 81.5 & 84.36 & 77.02 & 84.07\\
    \midrule
\multicolumn{14}{l}{\textit{Comorbidities at NIAD [N,(\%)]}}\\
Stroke & 137 (5.26) & 63 (7.35) & 141 (8.66) & 80 (10.78) & 158 (9.72) & 263 (20.61) & 109 (13.85) & 105 (8.35) & 138 (20.26) & 197 (5.32) & 73 (10.20) & 123 (19.10) & 30 (6.64)\\
Myocardial infarction & $<1$ \% & $<1$ \% & $<1$ \% & $<1$ \% & $<1$ \% & 322 (25.24) & 259 (32.91) & $<1$ \% & 189 (27.75) & $<1$ \% & $<1$ \% & 162 (25.16) & $<1$ \%\\
Congestive heart failure & $<1$ \% & $<1$ \% & $<1$ \% & $<1$ \% & $<1$ \% & 299 (23.43) & 179 (22.74) & $<1$ \% & 214 (31.42) & $<1$ \% & $<1$ \% & 230 (35.71) & $<1$ \%\\
Peripheral vascular disease & $<1$ \% & $<1$ \% & $<1$ \% & $<1$ \% & $<1$ \% & 203 (15.91) & 120 (15.25) & $<1$ \% & 161 (23.64) & $<1$ \% & $<1$ \% & 132 (20.50) & $<1$ \%\\
Diabetic retinopathy & 61 (2.34) & 485 (56.59) & 1014 (62.25) & 33 (4.45) & 24 (1.48) & 41 (3.21) & 96 (12.20) & $<1$ \% & 436 (64.02) & 73 (1.97) & 465 (64.94) & 23 (3.57) & 252 (55.75)\\
Hypertension & 110 (4.22) & 9 (1.05) & 1599 (98.16) & 704 (94.88) & 1582 (97.29) & 1226 (96.08) & 31 (3.94) & 1233 (98.01) & 663 (97.36) & 3587 (96.89) & 708 (98.88) & 626 (97.20) & 439 (97.12)\\
Chronic kidney disease & 113 (4.34) & 44 (5.13) & $<1$ \% & 42 (5.66) & 26 (1.60) & $<1$ \% & 55 (6.99) & 326 (25.91) & 56 (8.22) & $<1$ \% & 173 (24.16) & 204 (31.68) & 18 (3.98)\\
Osteoporosis & 177 (6.80) & 60 (7.00) & 65 (3.99) & 78 (10.51) & 483 (29.70) & 80 (6.27) & 101 (12.83) & $<1$ \% & 88 (12.92) & $<1$ \% & 53 (7.40) & 76 (11.80) & 22 (4.87)\\
COPD & 246 (9.45) & 95 (11.09) & 91 (5.59) & 647 (87.20) & $<1$ \% & 132 (10.34) & 195 (24.78) & $<1$ \% & 124 (18.21) & $<1$ \% & 53 (7.40) & 118 (18.32) & 31 (6.86)\\
Chronic liver disease & 57 (2.19) & 32 (3.73) & $<1$ \% & 18 (2.43) & 22 (1.35) & 16 (1.25) & 15 (1.91) & 19 (1.51) & 12 (1.76) & 46 (1.24) & $<1$ \% & 13 (2.02) & 157 (34.73)\\
    \midrule
\multicolumn{14}{l}{\textit{Comorbidities during follow-up [N,(\%)]}}\\
Stroke & 98 (3.76) & 22 (2.57) & 75 (4.60) & 27 (3.64) & 77 (4.74) & 68 (5.33) & 50 (6.35) & 57 (4.53) & 37 (5.43) & 134 (3.62) & 44 (6.15) & 45 (6.99) & 25 (5.53)\\
Myocardial infarction & $<1$ \% & $<1$ \% & 19 (1.17) & 9 (1.21) & $<1$ \% & 43 (3.37) & 14 (1.78) & 26 (2.07) & 36 (5.29) & 39 (1.05) & 8 (1.12) & 30 (4.66) & 5 (1.11)\\
Congestive heart failure & 45 (1.73) & 22 (2.57) & 33 (2.03) & 33 (4.45) & 40 (2.46) & 86 (6.74) & 58 (7.37) & 66 (5.25) & 58 (8.52) & 73 (1.97) & 38 (5.31) & 76 (11.80) & 13 (2.88)\\
Peripheral vascular disease & $<1$ \% & $<1$ \% & $<1$ \% & 13 (1.75) & 21 (1.29) & 42 (3.29) & 29 (3.68) & 16 (1.27) & 34 (4.99) & $<1$ \% & $<1$ \% & 28 (4.35) & 6 (1.33)\\
Diabetic retinopathy & 296 (11.37) & 372 (43.41) & 615 (37.75) & 109 (14.69) & 215 (13.22) & 82 (6.43) & 107 (13.60) & 169 (13.43) & 245 (35.98) & 415 (11.21) & 251 (35.06) & 57 (8.85) & 153 (33.85)\\
Hypertension & 324 (12.44) & 86 (10.04) & 27 (1.66) & 21 (2.83) & 32 (1.97) & 13 (1.02) & 56 (7.12) & 22 (1.75) & 9 (1.32) & 76 (2.05) & $<1$ \% & $<1$ \% & 12 (2.65)\\
Chronic kidney disease & 64 (2.46) & 34 (3.97) & 30 (1.84) & 26 (3.50) & 41 (2.52) & 17 (1.33) & 33 (4.19) & 116 (9.22) & 25 (3.67) & 61 (1.65) & 65 (9.08) & 89 (13.82) & 9 (1.99)\\
Osteoporosis & 78 (3.00) & 26 (3.03) & 56 (3.44) & 22 (2.96) & 147 (9.04) & 24 (1.88) & 22 (2.80) & 30 (2.38) & 24 (3.52) & 43 (1.16) & 23 (3.21) & 21 (3.26) & 12 (2.65)\\
COPD & 78 (3.00) & 22 (2.57) & 38 (2.33) & 93 (12.53) & 34 (2.09) & 37 (2.90) & 36 (4.57) & 33 (2.62) & 31 (4.55) & 54 (1.46) & 20 (2.79) & 40 (6.21) & 11 (2.43)\\
Chronic liver disease & 40 (1.54) & 9 (1.05) & $<1$ \% & 10 (1.35) & 23 (1.41) & $<1$ \% & $<1$ \% & 14 (1.11) & 10 (1.47) & 37 (1.00) & $<1$ \% & $<1$ \% & 43 (9.51)\\
Deaths before 2020 & 191 (7.33) & 86 (10.04) & 142 (8.72) & 165 (22.24) & 203 (12.48) & 603 (47.26) & 352 (44.73) & 134 (10.65) & 268 (39.35) & 211 (5.70) & 87 (12.15) & 279 (43.32) & 34 (7.52)\\
    \bottomrule
\end{tabular}
}
    \caption{Patient characteristics in each cluster for the \textit{Women $65+$} cohort with average values and standard deviation where applicable. Comorbidities are presented as absolute counts at NIAD initiation, along with the number of new comorbidities developed post-NIAD. Proportions, expressed as percentages of the cohort population, are reported in parentheses.}
    \label{tab:cluster_descr_female_65+}
\end{table}

\begin{figure}[ht]
    \centering
    \includegraphics[width=0.8\linewidth]{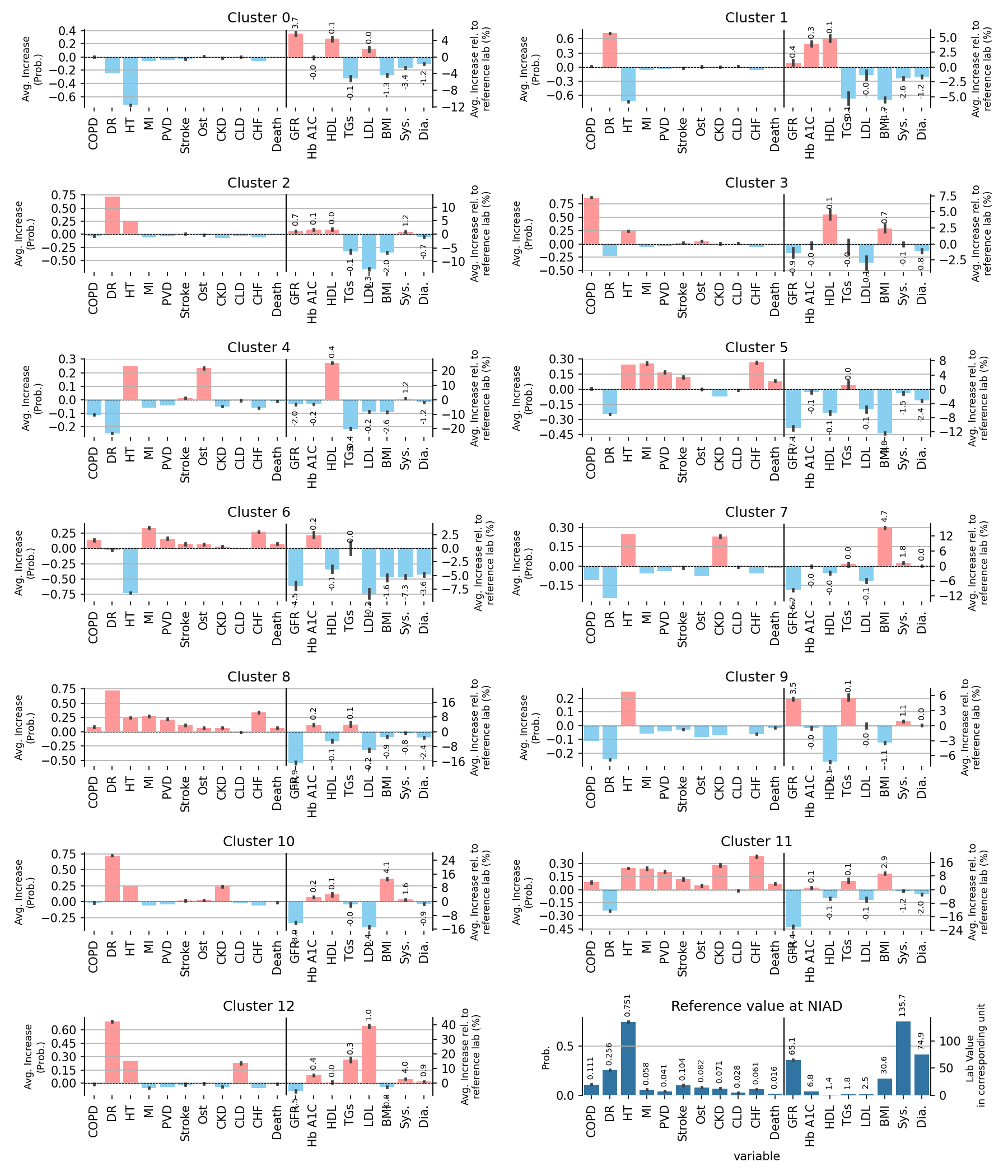}
    \caption{Average cluster-specific probabilities and lab values for \textit{Women $65+$}. Values have been centered by subtracting the average at the time of NIAD initiation to improve readability. Red bars indicate increases, and blue bars indicate decreases relative to the average patient at NIAD. Comorbidity risks are shown as changes in probability (left axis), and lab values as percentage changes relative to the reference value (right axis). Annotated numbers indicate absolute changes in the original units (see Table \ref{tab:list_of_var}). Error bars represent the standard error of the mean.}
    \label{fig:female65+_proba_avg}
\end{figure}

\begin{figure}[ht]
    \centering
   \begin{subfigure}[b]{0.31\textwidth}
        \centering
         \includegraphics[width=1\textwidth]{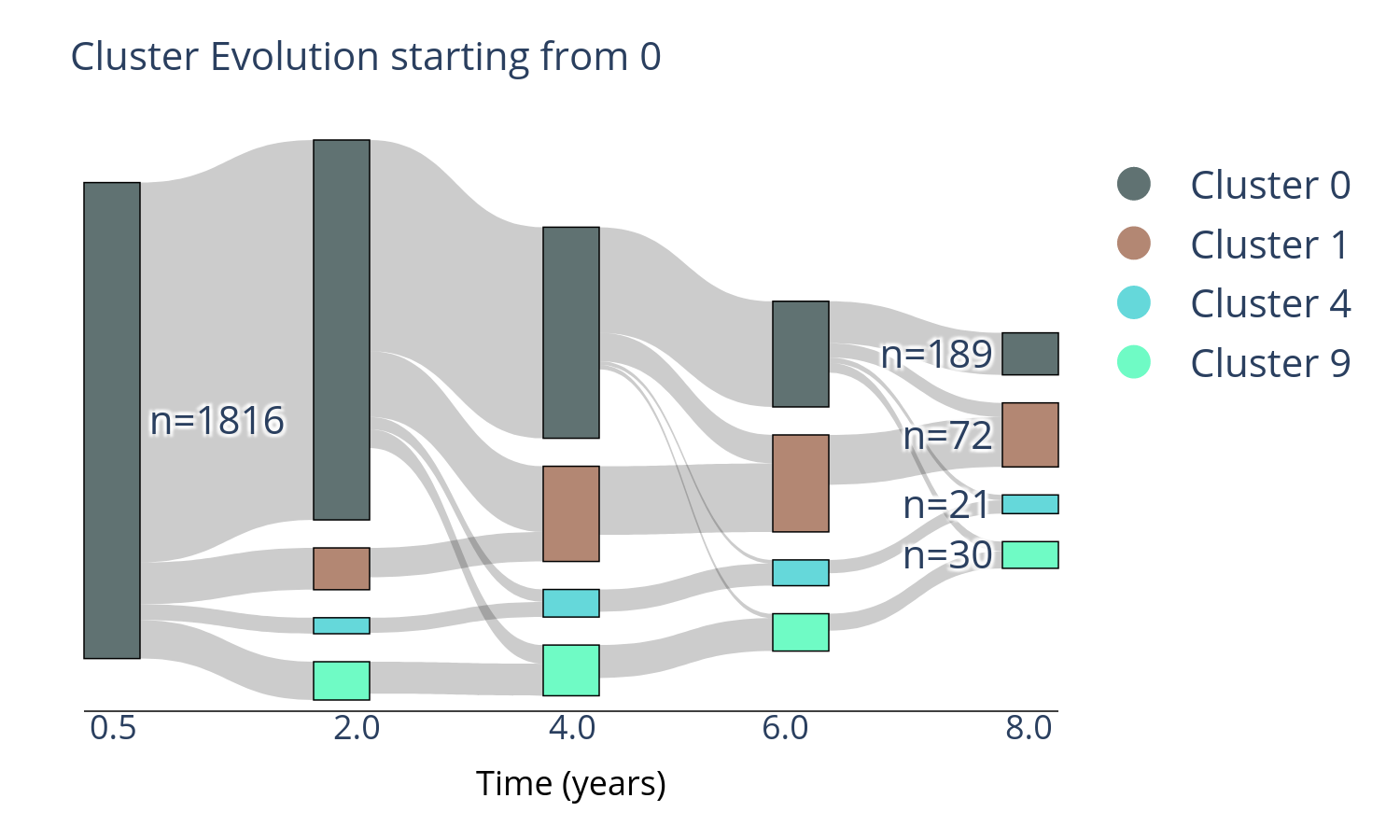}
    \end{subfigure}
    \begin{subfigure}[b]{0.31\textwidth}
        \centering
         \includegraphics[width=1\textwidth]{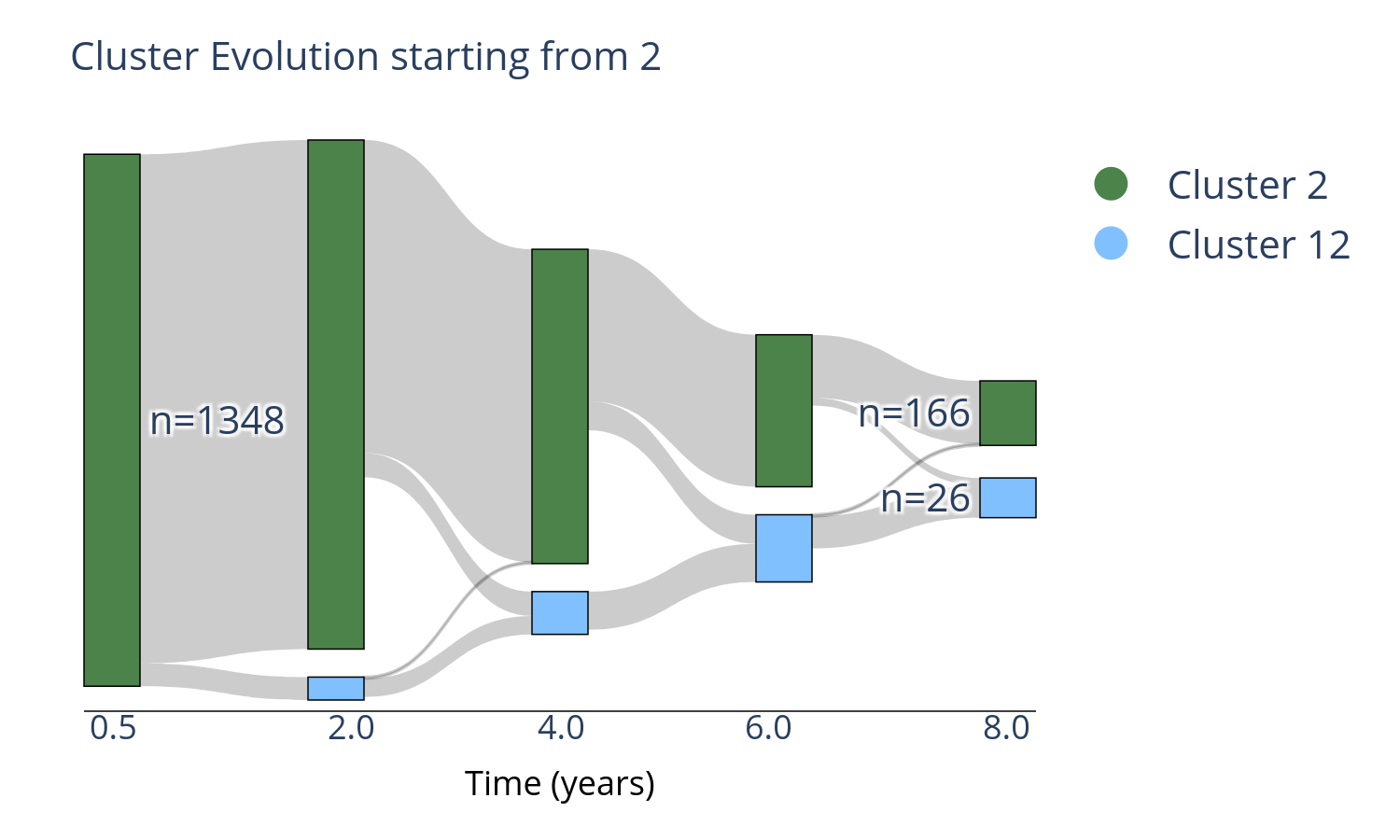}
    \end{subfigure}
    \begin{subfigure}[b]{0.31\textwidth}
        \centering
         \includegraphics[width=1\textwidth]{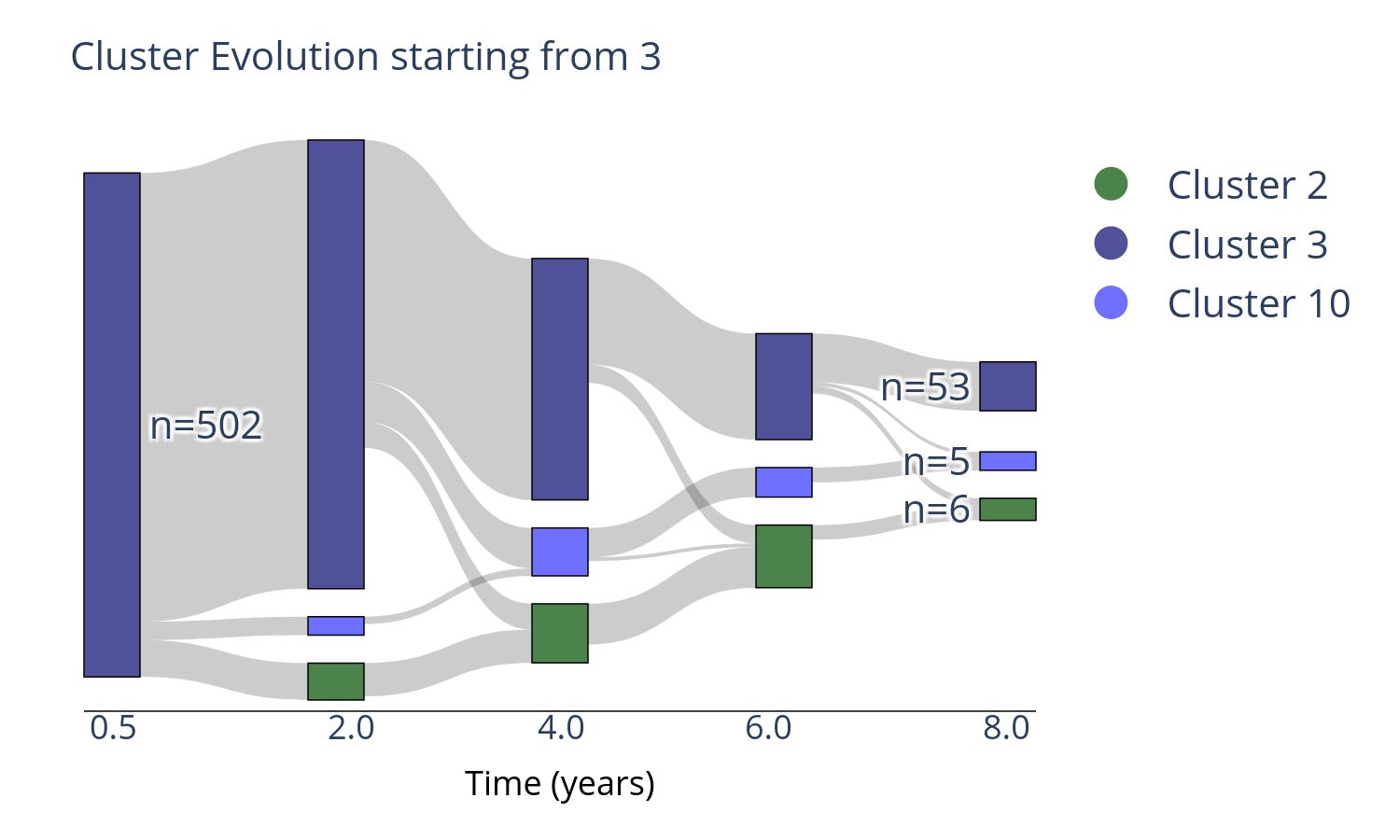}
    \end{subfigure}
     \begin{subfigure}[b]{0.31\textwidth}
        \centering
         \includegraphics[width=1\textwidth]{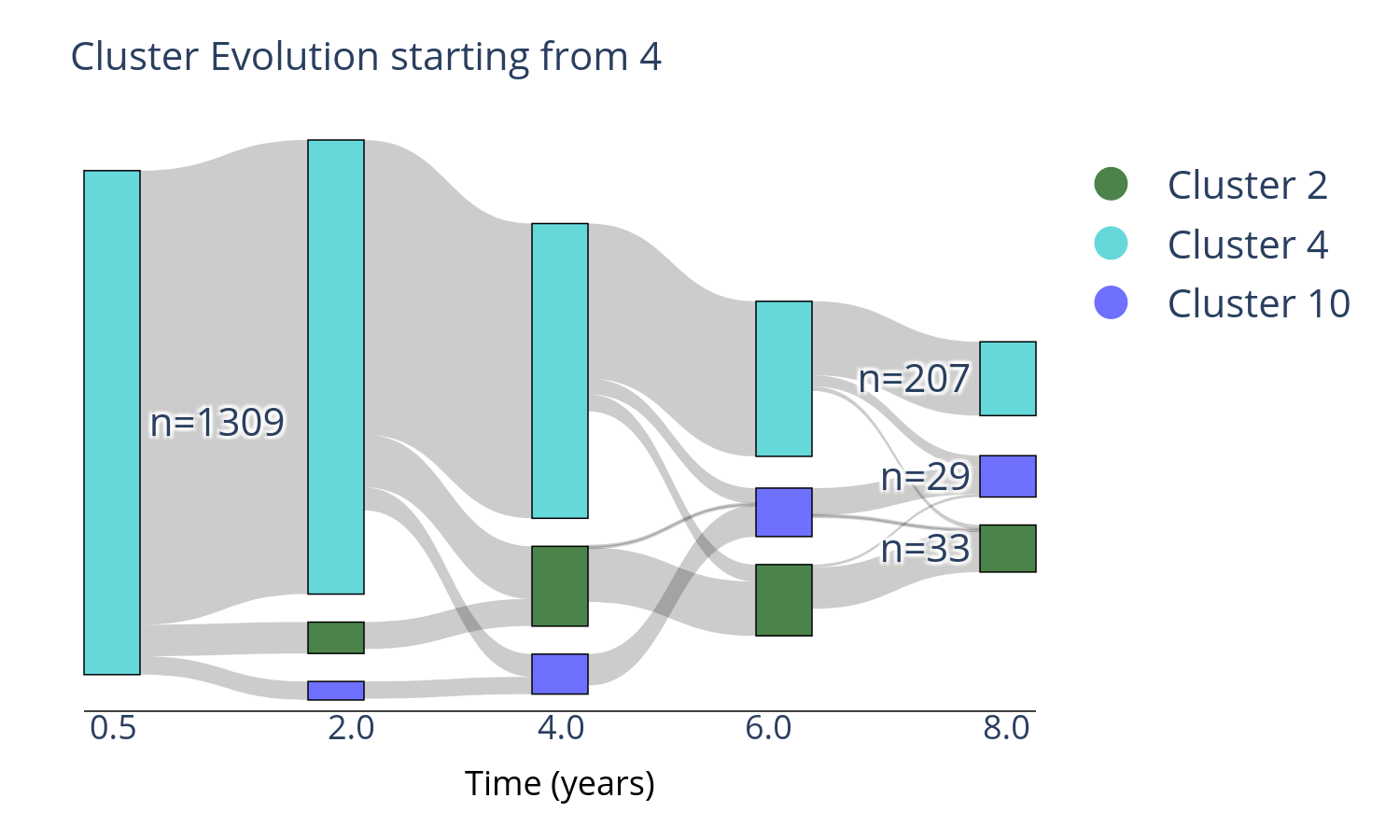}
    \end{subfigure}
     \begin{subfigure}[b]{0.31\textwidth}
        \centering
         \includegraphics[width=1\textwidth]{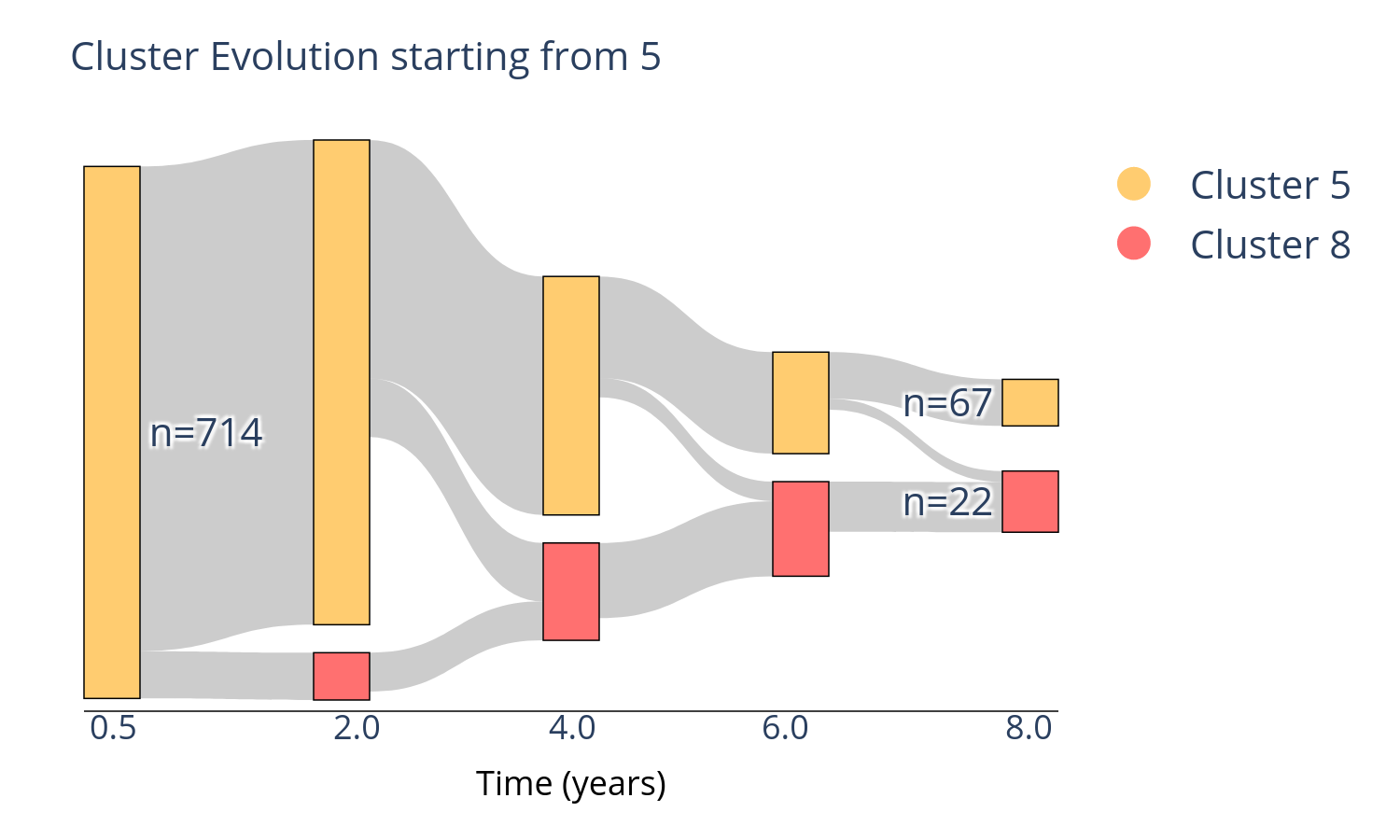}
    \end{subfigure}
     \begin{subfigure}[b]{0.31\textwidth}
        \centering
         \includegraphics[width=1\textwidth]{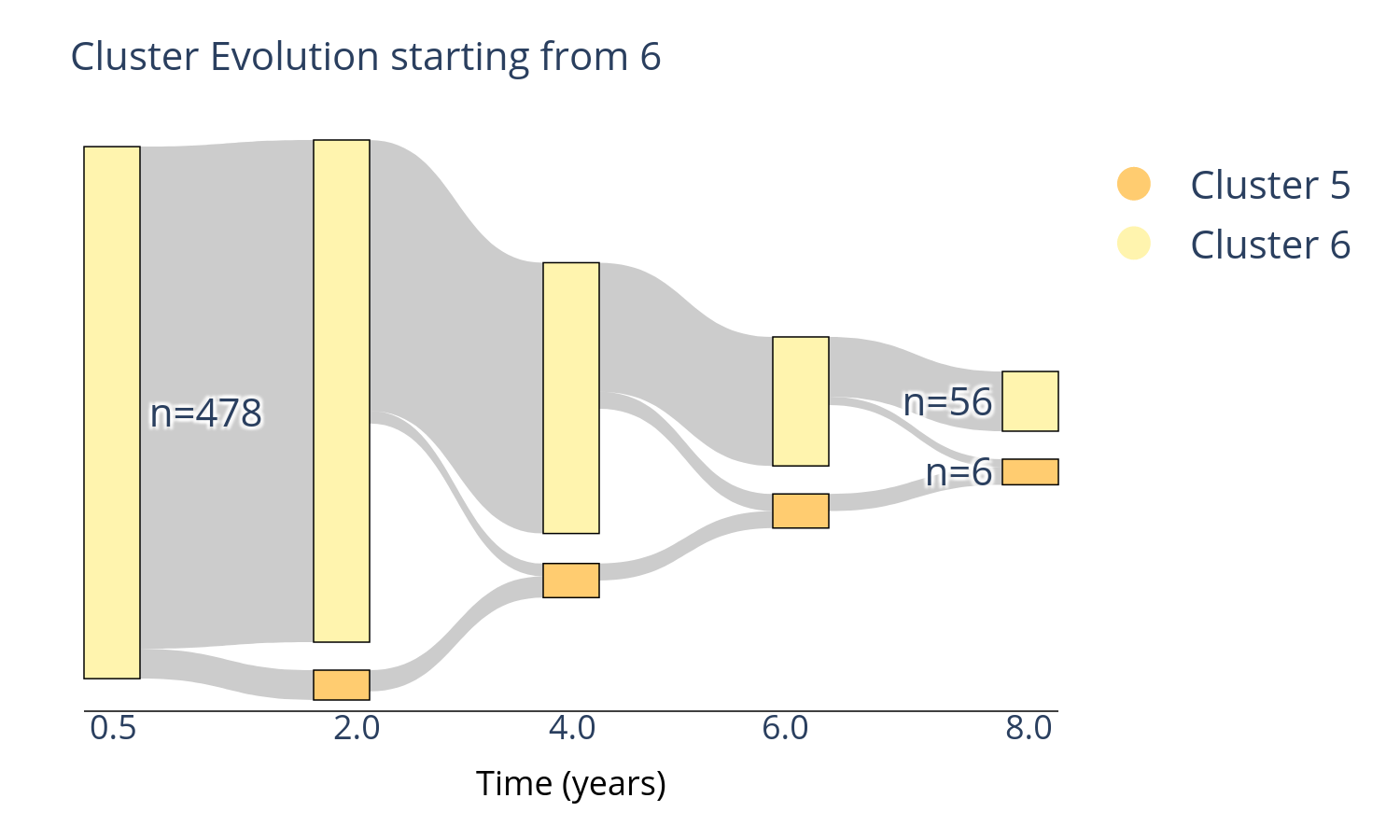}
    \end{subfigure}
     \begin{subfigure}[b]{0.31\textwidth}
        \centering
         \includegraphics[width=1\textwidth]{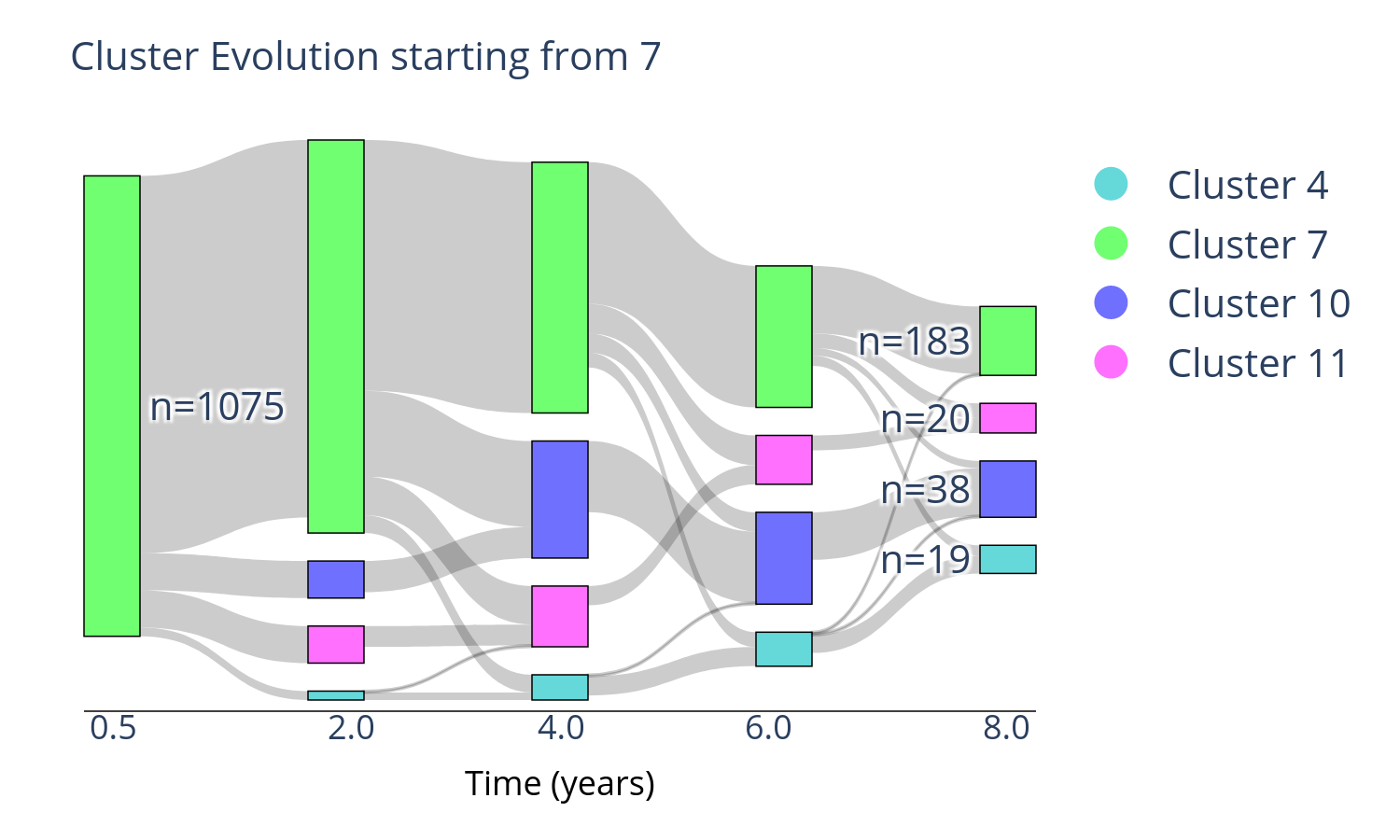}
    \end{subfigure}
    \begin{subfigure}[b]{0.31\textwidth}
        \centering
         \includegraphics[width=1\textwidth]{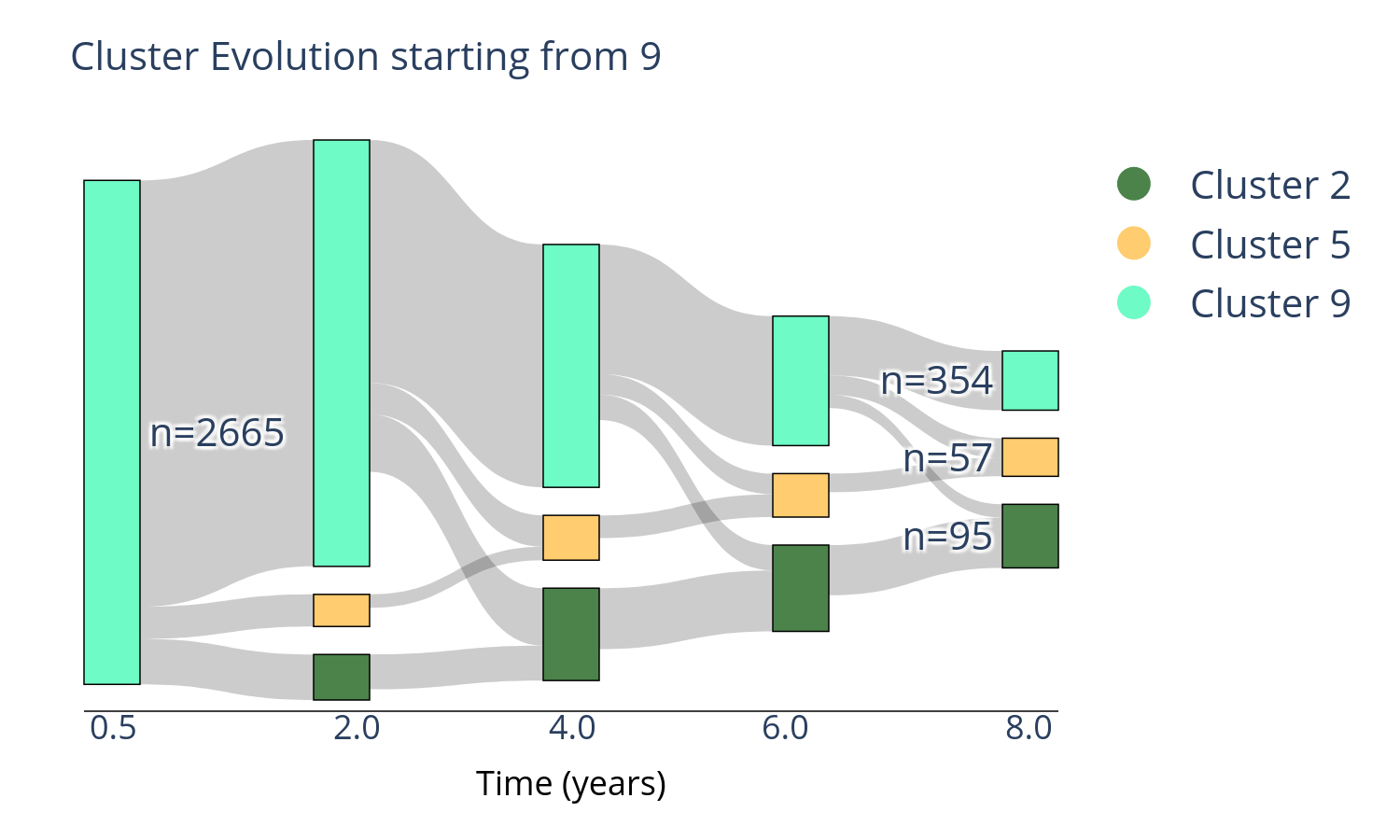}
    \end{subfigure}
    \begin{subfigure}[b]{0.31\textwidth}
        \centering
         \includegraphics[width=1\textwidth]{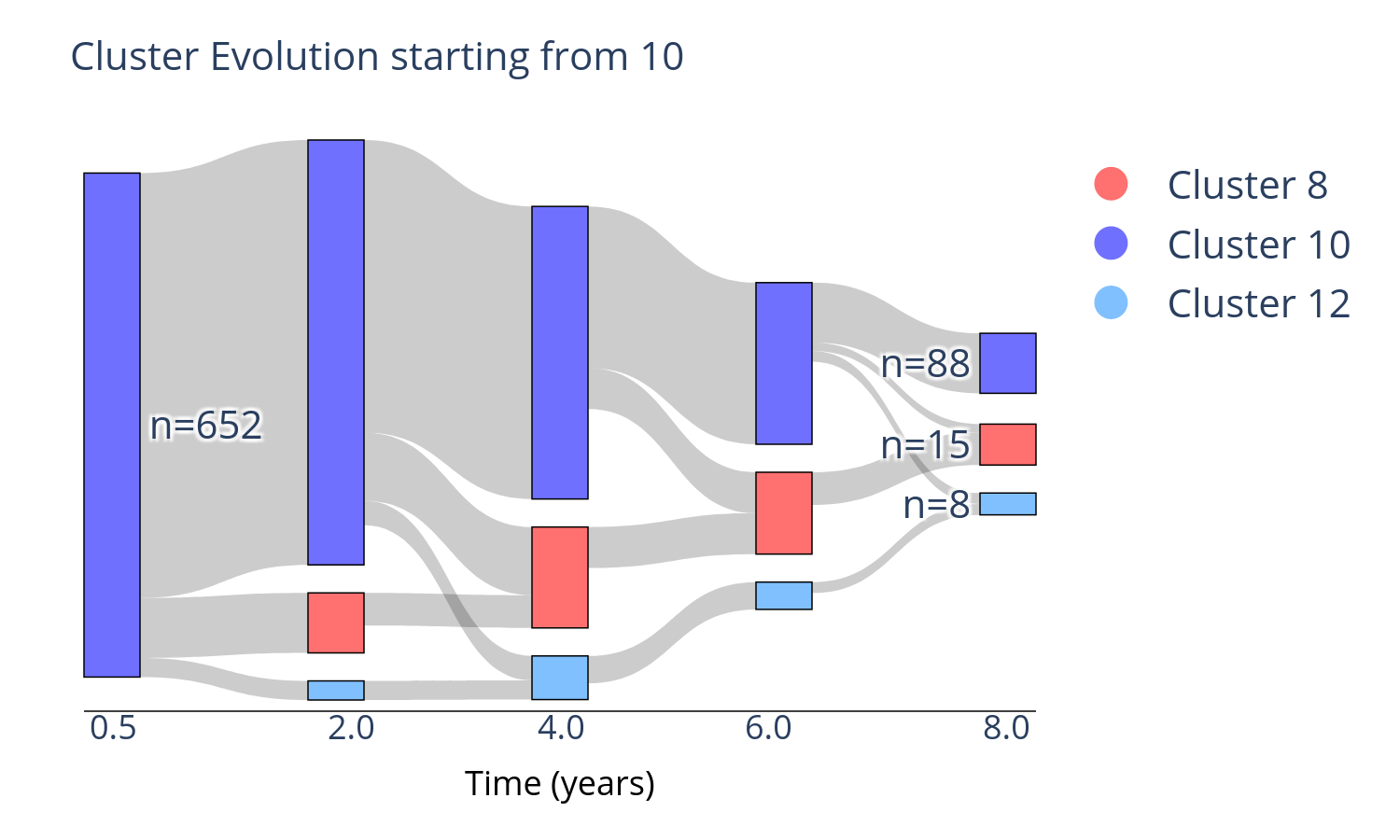}
    \end{subfigure}
    \begin{subfigure}[b]{0.31\textwidth}
        \centering
         \includegraphics[width=1\textwidth]{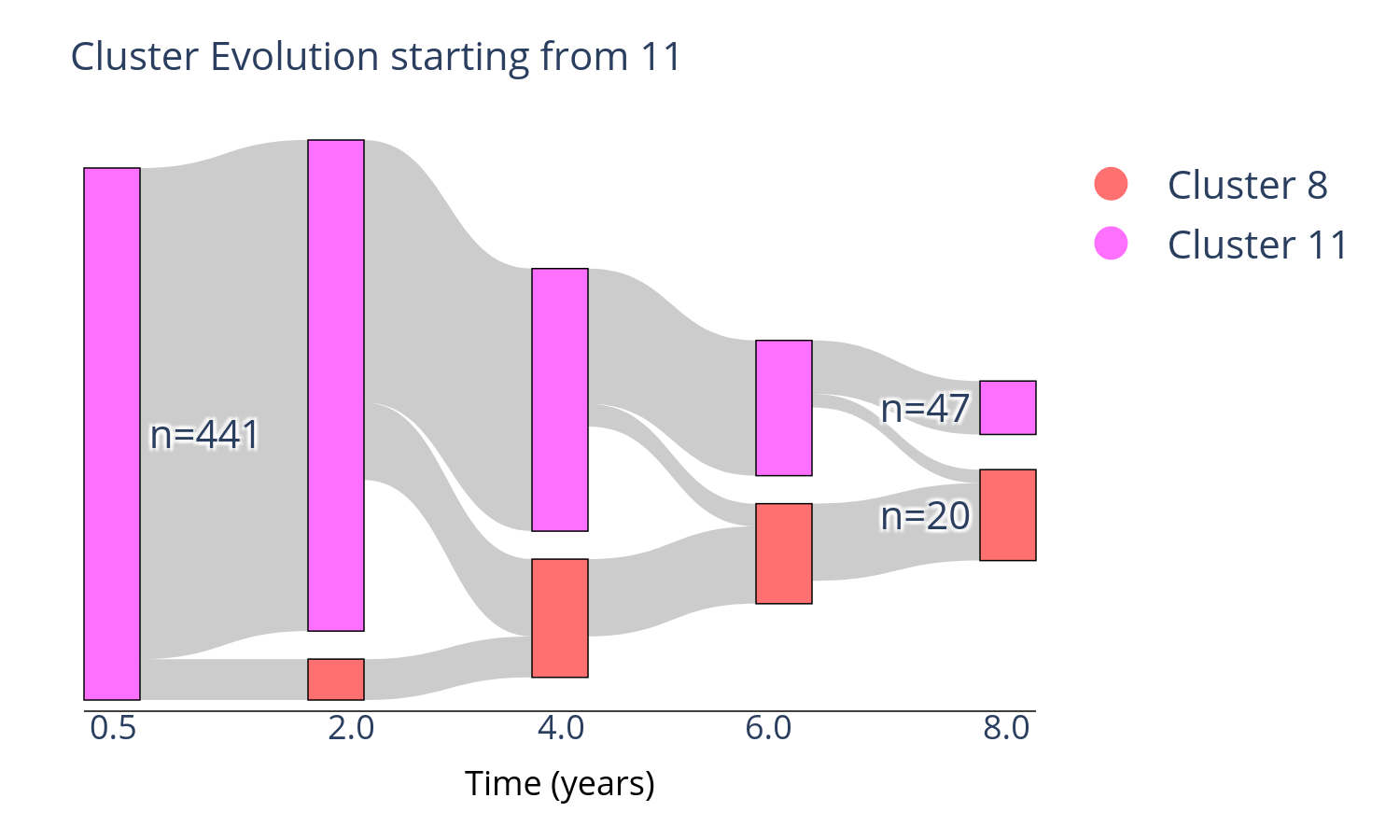}
    \end{subfigure}
    \begin{subfigure}[b]{0.31\textwidth}
        \centering
         \includegraphics[width=1\textwidth]{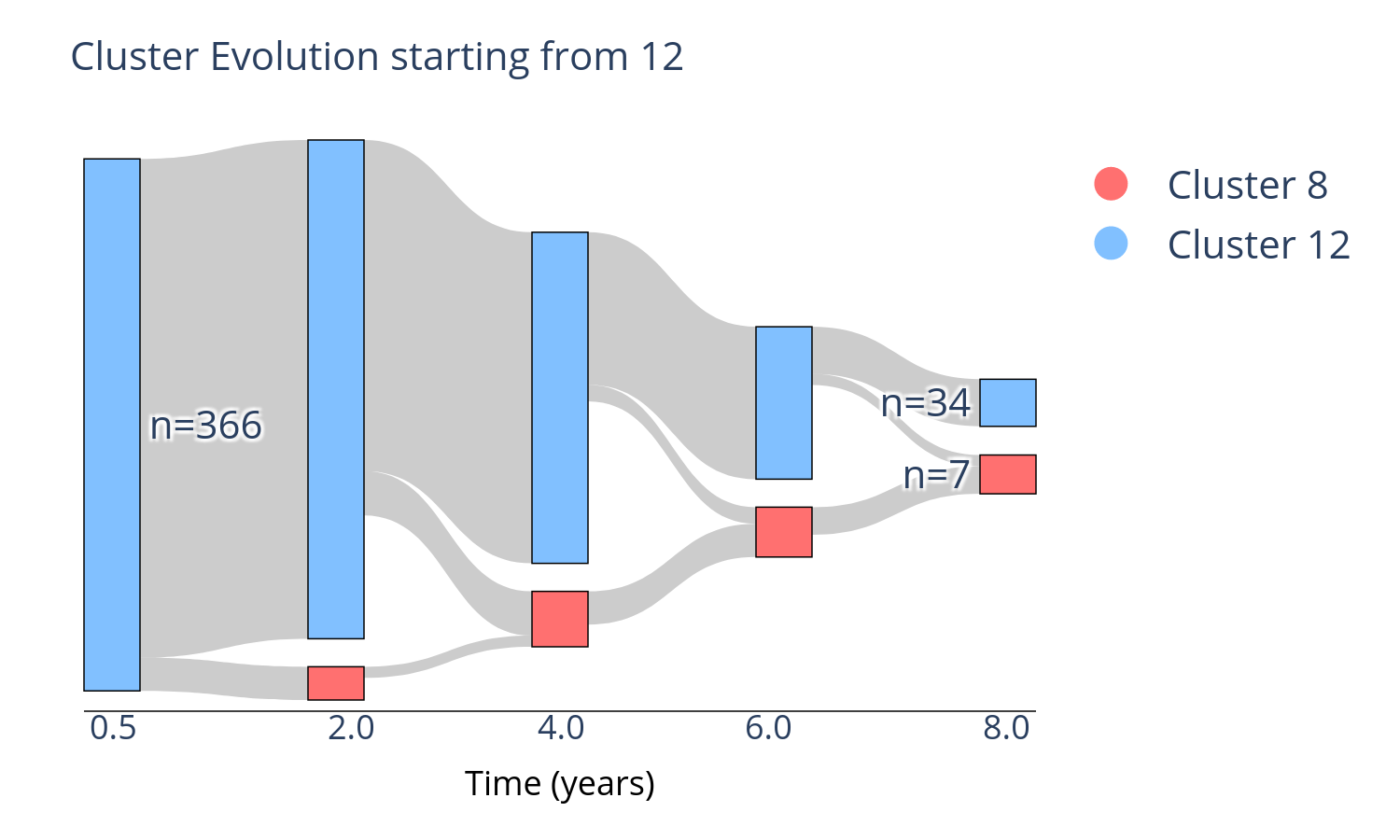}
    \end{subfigure}
    \caption{Evolution of patient clusters over time (\textit{Women $65+$}), conditional on the initial cluster assignment (6 months post-NIAD). The number of patients in each cluster is indicated alongside the corresponding bars. For clarity, only the most frequent transitions are shown (see Figure \ref{fig:female65+_transition_prob} for more details). Clusters 1,8 are excluded due to the low number of transitions observed from these clusters.}
    \label{fig:female65+_evolution}
\end{figure}

\begin{figure}[ht]
    \centering
       \includegraphics[width=0.8\textwidth]{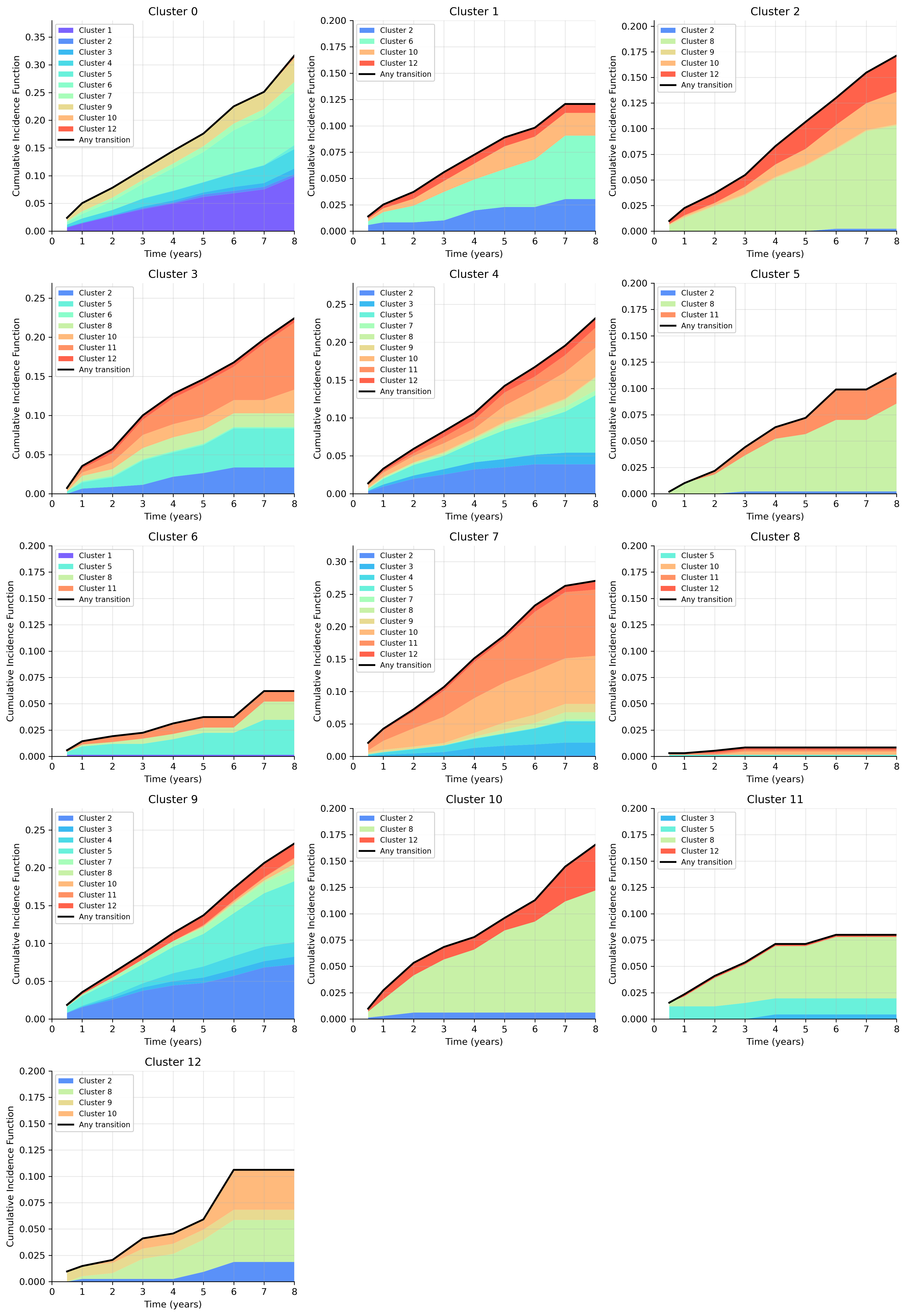}
        \caption{
Cumulative Incidence Function (CIF) for the time to first cluster transition in \textit{Women $65+$}, stratified by initial cluster assignment at 6 months post-NIAD.
The estimates account for censoring due to loss to follow-up and study termination, computed using the Aalen nonparametric estimator \cite{aalen1978}.
}
    \label{fig:female65+_transition_prob}
\end{figure}

\begin{figure}[ht]
    \centering
    \includegraphics[width=1\linewidth]{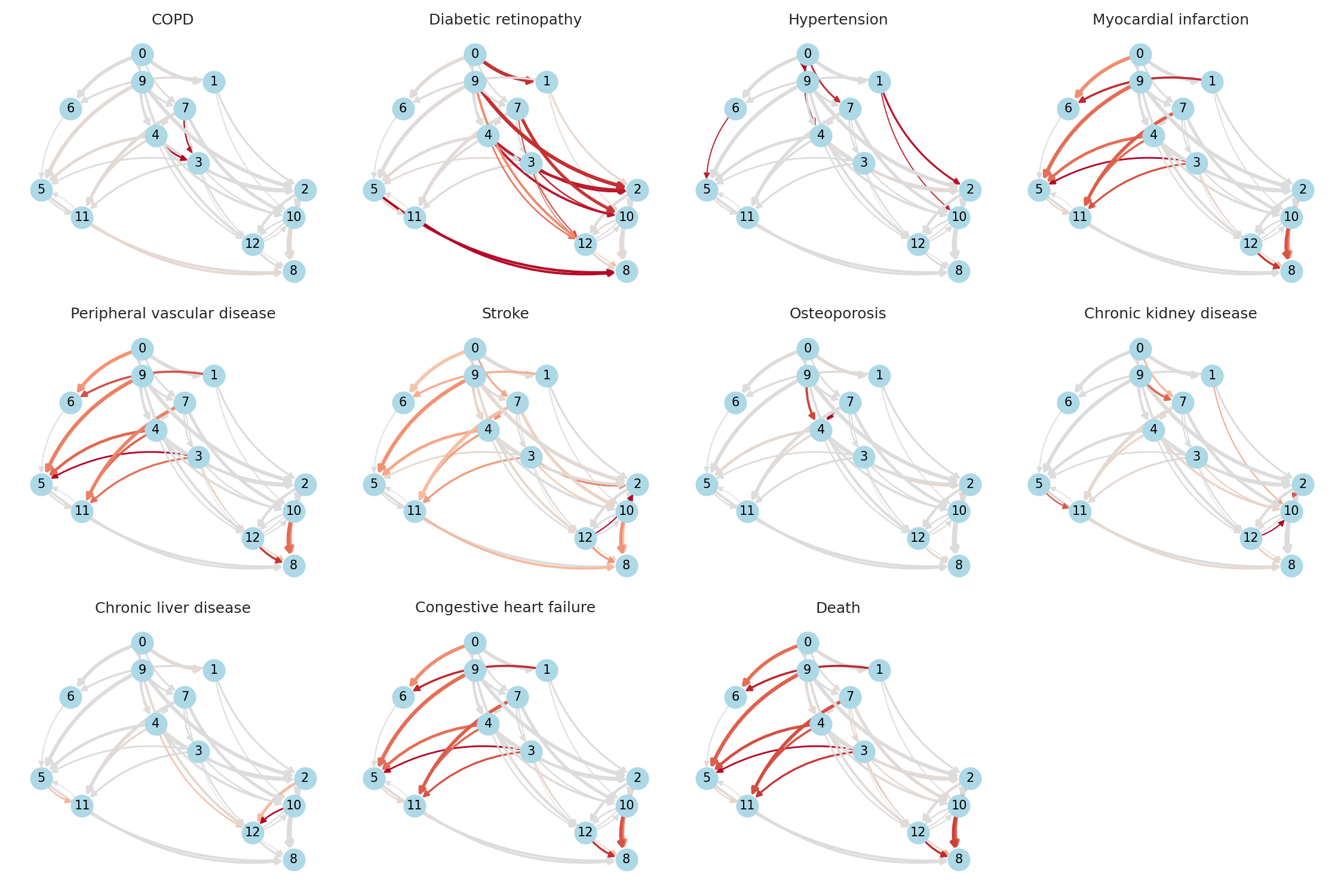}
    \caption{Transition graph. Red arrows (resp. blue) means an increase (resp. decrease) of the comorbidity risk (\textit{Women $65+$}). Arrow thickness is proportional to the number of observed transitions. Transitions with less than 1\% probability are not shown.}
    \label{fig:female65+_graph}
\end{figure}

\end{document}